\documentclass[journal]{IEEEtran}
\usepackage{times}
\usepackage{helvet}
\usepackage{courier}
\usepackage{times}
\usepackage[dvips]{graphicx}
\usepackage{multirow}
\usepackage{epstopdf}
\usepackage{graphicx}
\usepackage{subfigure}

\usepackage{times}
\usepackage{helvet}
\usepackage{courier}

\usepackage{url}
\usepackage{array}
\usepackage{ifthen}
\usepackage{balance}
\usepackage{bm}
\usepackage{amsmath}
\usepackage{amssymb}
\usepackage{mathrsfs}
\usepackage{color}
\usepackage{graphicx}
\usepackage{algorithm,algorithmic}
\usepackage{sidecap}
\usepackage{float}
\usepackage{graphicx} 
\usepackage{epstopdf}
\usepackage{threeparttable}
\usepackage{color, colortbl}
\definecolor{Gray}{gray}{0.9}

\def\length{{0.1}}
\def\lengthw{{0.1}}

\def\X{\mathbf{X}}
\def\LL{\mathbf{L}}
\def\E{\mathbf{E}}
\def\J{\mathbf{J}}
\def\Y{\mathbf{Y}}
\def\W{\mathbf{W}}
\def\Z{\mathbf{Z}}
\def\D{\mathbf{D}}
\def\U{\mathbf{U}}
\def\V{\mathbf{V}}
\def\A{\mathbf{A}}

\definecolor{blue}{RGB}{0,0,255}

\ifCLASSINFOpdf
\else
\fi
\begin{document}

\title{Superpixel-guided Discriminative Low-rank Representation of
Hyperspectral Images for Classification}

\author{Shujun~Yang,~\IEEEmembership{Member,~IEEE,}
	Junhui~Hou,~\IEEEmembership{Senior Member,~IEEE,}
    Yuheng~Jia,~\IEEEmembership{Member,~IEEE,}\\
     Shaohui Mei,~\IEEEmembership{Senior Member,~IEEE,}
    and Qian Du,~\IEEEmembership{Fellow,~IEEE}
 \thanks{This work was supported in part by the Hong Kong RGC under Grants 9048123 (CityU 21211518), 9042820 (CityU 11219019), and 9042955 (CityU 11202320), and in part by the Natural Science Foundation of Jiangsu Province (Grants No BK20210221). (\textit{Corresponding Author: Junhui Hou and Yuheng Jia})}
\thanks{S. Yang  and J. Hou are with the Department of Computer Science, City University of Hong Kong, Kowloon, Hong Kong; (e-mail: sjyang8-c@my.cityu.edu.hk; jh.hou@cityu.edu.hk).}
\thanks{Y. Jia is with the School of Computer Science and Engineering, Southeast University, Nanjing 210096, China, and also with Key Laboratory of Computer Network and Information Integration (Southeast University), Ministry of Education, China; (e-mail: yhjia@seu.edu.cn).}
\thanks{S. Mei is with the School of Electronics and Information, Northwestern
        Polytechnical University, Xian 710129, China; (e-mail: meish@nwpu.edu.cn).}
\thanks{Q. Du is with the Department of Electrical and Computer Engineering,
        Mississippi State University, Starkville, MS 39762 USA; (e-mail: du@ece.msstate.edu).}
}

\maketitle

\begin{abstract}
In this paper, we propose a novel classification scheme for the remotely sensed hyperspectral image (HSI), namely SP-DLRR, by comprehensively exploring its unique characteristics, including the local spatial information and low-rankness. SP-DLRR is mainly composed of two modules, i.e., the classification-guided superpixel segmentation and the discriminative low-rank representation, which are iteratively conducted. Specifically, by utilizing the local spatial information and incorporating the predictions from a typical classifier, the first module segments pixels of an input HSI (or its restoration generated by the second module) into superpixels. According to the resulting superpixels, the pixels of the input HSI are then grouped into clusters and fed into our novel discriminative low-rank representation model with an effective numerical solution. Such a model is capable of increasing the intra-class similarity by suppressing the spectral variations locally while promoting the inter-class discriminability globally, leading to a restored HSI with more discriminative pixels. Experimental results on three benchmark datasets demonstrate the significant superiority of SP-DLRR over state-of-the-art methods, especially for the case with an extremely limited number of training pixels.
\end{abstract}

\IEEEpeerreviewmaketitle
\section{Introduction}
A hyperspectral image (HSI) contains hundreds of narrow spectral bands ranging from the visible to the infrared spectrum~\cite{plaza2009recent}.
Such rich spectral information of HSIs gives rise to breakthroughs in various applications, such as military~\cite{manolakis2003detection,zhang2018target}, agriculture~\cite{datt2003preprocessing,patel2001study}, and mineralogy~\cite{horig2001hymap}.
Due to the complex imaging conditions (e.g. illumination, environmental, atmospheric, and temperature conditions and sensor interference) \cite{zare2013endmember}, the acquired HSIs in remote sensing usually suffer from spectral variations, i.e., pixels (or spectral signatures) corresponding to the same material may be different, which will degrade the downstream processing, such as classification \cite{mei2017hyperspectral,mei2018simultaneous,zhang2013nonlocal}, segmentation \cite{li2010semisupervised}, and unmixing \cite{bioucas2012hyperspectral,iordache2011sparse}. We aim to recover the discriminative representation of a degenerated HSI to improve the performance of the subsequent applications. Particularly , this paper is focused on the classification task.

Recent studies show that the exploration of the low-rank property of HSIs can alleviate the spectral variations and thus improve the classification performance \cite{mei2017hyperspectral,mei2018simultaneous,mei2016spectral}.
 Numerous low-rank-based models have been proposed for HSI imagery~\cite{mei2016spectral,de2016kernel,he2015hyperspectral,sumarsono2016low,wang2018self,zhang2013hyperspectral,zhang2015ensemble}. For example, some methods assume that the spectral signatures satisfy the low-rank property from a global perspective \cite{mei2016spectral,sumarsono2016low,sun2018graph,sun2019lateral}.
 To better take advantage of the local spatial structure, i.e., the pixels within a small neighborhood often belong to the same class, local low-rank-based methods were also proposed 
 \cite{he2015hyperspectral,zhang2013hyperspectral,zhu2014spectral},
 in which an HSI is divided into patches, and then each patch is processed sequentially by a robust low-rank matrix representation model.
As the regular patch-based segmentation cannot sufficiently capture the complex local spatial structure accurately, some superpixel-based models \cite{wang2018self,xu2015spectral,mei2019psasl,liu2019kernel,li2020superpixel,xu2020hypergraph} were proposed, where an HSI is segmented into many shape-adaptive local regions called superpixels. 
Most recently, spatial and spectral joint low-rank representation models were proposed, in which the low-rank property is explored in both spatial and spectral domains \cite{mei2017hyperspectral,mei2018simultaneous}.
These methods are similar to the tensor-based low-rank modeling \cite{wang2021error} for HSIs to some extent \cite{an2018tensor,deng2018tensor}.
In addition, owing to the powerful ability of deep neural networks in learning representations 
~\cite{zhang2019hybrid,jiang2018hyperspectral,wang2018low},
deep learning-based low-rank modeling has been proposed, which combines the strong deep learning-based feature extraction and the low-rank representation-based robust classifier to improve classification accuracy~\cite{wang2018low}.

 However, the previous low-rank-based methods either compromise the discriminability between classes or fail to capture the local spatial structure sufficiently.
To this end, we propose a discriminative low-rank representation model, which is capable of increasing the intra-class similarity by suppressing the spectral variations  locally while promoting the inter-class discriminability in a global manner, leading to a more discriminative representation. Technically, our model is elegantly formulated as a constrained optimization problem, which is then numerically solved with an effective iterative algorithm. Based on such a discriminative low-rank model, we propose a superpixel induced classification framework, which alternatively implements two modules in an iterative manner, i.e., the classification-guided superpixel segmentation and the discriminative low-rank representation. Specifically, the first module takes advantage of both the spatial information and predictions of a typical classifier for segmenting an input HSI or its discriminative representation by the second module into superpixels. Then, the pixels of the input HSI are grouped into clusters according to the resulting superpixels, which are further discriminatively represented with the second module.
The experiments validate that our method outperforms state-of-the-art methods to a significant extent. Especially such an advantage is more obvious when the training pixels are extremely limited. Besides, our method can handle the imbalance behavior better, i.e., it improves the classification accuracy of the classes with only a few pixels significantly.

The rest of this paper is organized as follows. In Section \uppercase\expandafter{\romannumeral2}, the related work about low-rank-based methods for modeling HSIs are reviewed. The proposed classification framework is then presented in Section
\uppercase\expandafter{\romannumeral3}, followed by the extensive experimental results and discussion in Section \uppercase\expandafter{\romannumeral4}. Finally, Section \uppercase\expandafter{\romannumeral5} concludes this paper.

\section{Related Work}\label{sec:related}
In this section, we mainly review the low-rank-based methods for processing HSIs. Some researchers restored the discriminative representation of a degenerated HSI by exploring the low-rank property globally. Specifically, Mei \textit{et al.} \cite{mei2016spectral} utilized an $\ell_1$-based low-rank matrix approximation for hyperspectral classification.
Sumarsono and Du proposed a low-rank subspace representation model \cite{sumarsono2016low} to preprocess the spectral feature, that followed by the classification on both supervised and unsupervised learning.

However, the above methods did not utilize the local structure well, i.e., the pixels within a small region often belong to the same class, which limits their performance. Thus, local low-rank-based methods \cite{wang2016robust} have been proposed \cite{wang2018self,zhang2013hyperspectral,zhu2014spectral,xu2015spectral,mei2019psasl,li2020superpixel,xu2020hypergraph}. Specifically,
Zhang \textit{et al.} \cite{zhang2013hyperspectral} divided an HSI into patches, and then each patch was processed sequentially by a robust low-rank matrix representation model.
Zhu \textit{et al.} \cite{zhu2014spectral} utilized the low-rank property to obtain pre-cleaned patches and applied the spectral non-local method to restore the image.
Furthermore, as the regular patch-based segmentation methods cannot accurately characterize the complex local structure of an HSI, some superpixel-based  methods~\cite{wang2018self,xu2015spectral,mei2019psasl,li2020superpixel,xu2020hypergraph} were proposed, which 
segment an HSI into many shape-adaptive regions called superpixels.
Specifically,
Xu \textit{et al.} \cite{xu2015spectral} explored the low-rank property for each homogeneous region followed by the probabilistic support vector
machine (SVM) classifier, and the markov random field model then captured the local correlation.
Wang \textit{et al.} \cite{wang2018self} proposed a self-supervised low-rank representation model which takes advantage of the spectral-spatial graph regularization on pixel-level to suppress the redundancy of an HSI, and the superpixel-level constraint is also employed to incorporate the semantic information around pixels.
Mei \textit{et al.} \cite{mei2019psasl} also proposed a pixel-level and superpixel-level aware subspace learning method based on the reconstruction independent component analysis algorithm. Li \textit{et al.}~\cite{li2020superpixel} proposed a sparse unmixing algorithm, which applies a low-rank constraint to  the pixels inside each superpixel, which are assumed to have the same endmembers and similar abundance values. Furthermore, the TV regularization is also added to boost the smoothness of the abundance maps. Xu \textit{et al.}~\cite{xu2020hypergraph} proposed a hyper-graph-regularized low-rank subspace clustering, which constructs a hyper-graph from the superpixels, i.e., building the hyper-edge weights from these superpixels to include both spectral and spatial similarities between pixels.

 All the above methods only exploit the low-rank property in the spectral domain. Recently, methods for exploring the low-rank property in both spectral and spatial domains were proposed.
 Pan \textit{et al.} \cite{pan2018hyperspectral} proposed a latent low-rank representation model that explores the low-rank property from two directions, i.e., row and column directions, to construct more discriminative representations. Mei \textit{et al.} \cite{mei2017hyperspectral} proposed to exploit the spectral and spatial low-rank property in two separate processing steps. Mei \textit{et al.} \cite{mei2018simultaneous} also proposed to unify spatial and spectral low-rank property using a single optimization model, where the low-rank property is considered from two directions: locality in the spectral domain (patch-wise-based image segmentation) and globality in the spatial domain (the low-rank prior to each spectral band).
 The nonlinear low-rank was also considered for modeling the classification of HSIs \cite{de2016kernel,liu2019kernel}. Morsier \textit{et al.} \cite{de2016kernel} proposed a nonlinear low-rank and sparse representation model to optimize the sample relationships. Liu \textit{et al.} \cite{liu2019kernel} proposed a kernel low-rank model based on the local similarity, which exploits the low-rank property for each superpixel segmentation region in a kernel space.

 Low-rank representation was also used together with dictionary-based sparse representation for HSI classification  \cite{wang2017weighted,wang2018locality,chen2016fast,pan2017hyperspectral}. Specifically, Wang \textit{et al.} \cite{wang2017weighted}
 proposed locally weighted regularization to model the correlation between samples to preserve the local structure.
 Wang \textit{et al.} \cite{wang2018locality} proposed a local and structure-regularized low-rank representation method, which integrates a distance metric that characterizes the local structure. Chen \textit{et al.} \cite{chen2016fast} proposed a group-wise-based model that employs low-rank decomposition to partition the HSIs into many groups of similar pixels, where the pixels in each group can be classified jointly and assigned to the same label. Pan \textit{et al.} \cite{pan2017hyperspectral} proposed a low-rank and sparse representation classifier with a spectral consistency constraint, which integrates the adaptive spectral constraint into low-rank and sparse terms.

As HSIs are naturally represented in the form of a 3D tensor, tensor low-rank was recently investigated for modeling HSIs \cite{an2018tensor,deng2018tensor}. An \textit{et al.} \cite{an2018tensor} addressed the dimensionality reduction problem by a tensor-based low-rank graph, which characterizes the within-class and between-class relationship by multi-manifold regularization. Deng \textit{et al.} \cite{deng2018tensor} proposed the tensor low-rank discriminative embedding to preserve the intrinsic geometric structure, in which the label information is utilized to strengthen the discriminative ability of the representation.

\section{Proposed Method}\label{sec:scHSI}
\subsection{Overview of the Proposed Classification Scheme}
\label{sec:scheme}

\begin{figure}[t]
\centering
\includegraphics[width=0.35\textwidth]{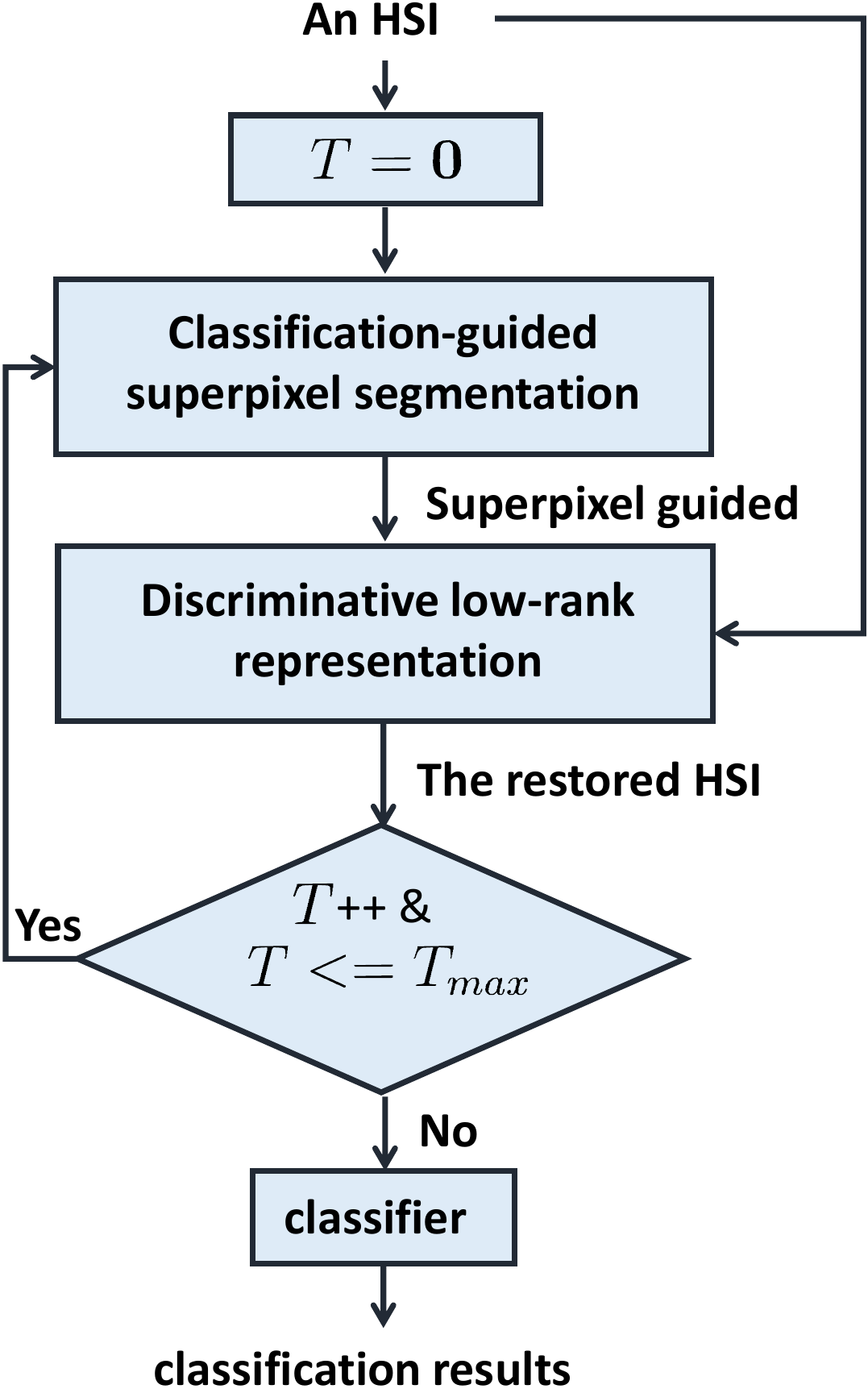}
\caption{The flowchart of the proposed SP-DLRR for HSI classification. The iteration index $T$ starts from zero, and $T_{max}$ is the max number of iterations.}
\label{fig:flowchart}
\end{figure}
Let $\mathbf{X}\in\mathbb{R}^{b\times n}$ be an HSI with $n$ pixels and $b$ spectral bands, i.e., each column of $\mathbf{X}$ corresponds to a $b$-dimensional pixel. As shown in Fig. \ref{fig:flowchart}, the proposed method named SP-DLRR 
for HSI classification consists of the following two modules.

1) \textit{Classification-guided superpixel segmentation}:
we apply a typical superpixel segmentation algorithm on the input of this module, which is $\X$ for the first iteration and its restoration for the subsequent iterations. Meanwhile, the predictions of a typical classifier trained on the input are employed to refine the segmentation process.
 More details is presented in the succeeding Section \ref{sec:SR}.

2) \textit{Discriminative low-rank representation}: according to the induces of the resulting superpixels, the pixels of $\X$ are then grouped into multiple clusters, which are further processed by the proposed discriminative low-rank representation model
to produce a restored HSI with more discriminative pixels, denoted as $\LL\in\mathbb{R}^{b\times n}$. See more details in Section \ref{sec:DLR}.

We alternatively conduct the two modules  in an iterative manner until the pre-defined number of iterations $T_{max}$ is achieved. The resulting $\LL$ is finally used for classification with a typical classifier, e.g., SVM.

\subsection{Classification-guided Superpixel Segmentation}\label{sec:SR}
 Considering the fact that a set of neighboring pixels mostly correspond to the same terrestrial object, the input of this module (i.e., $\X$ for the first iteration and $\LL$ for the subsequent ones) is segmented.
 Specifically,
 we first generate the superpixel segmentation on the current input of this module by using a typical superpixel segmentation algorithm. In this paper, we adopt the entropy rate superpixel method \cite{liu2011entropy}. Note that other advanced superpixel segmentation algorithms can be used as well.
For each superpixel, based on the predictions of a typical classifier trained on the current input, we count the number of pixels for each class denoted as  $\{h_1,\ldots,h_c,\ldots,h_C\}$ with $h_c$ being the number of pixels belonging to the $c$-th class and $C$ being the total number of classes.
 Then the following ratio is defined for each class:
 \begin{equation}
 r_c=\frac{h_c}{\sum_{i=1}^{C}h_i} \in [0,1].
 \end{equation}
The maximum ratio (MR) is defined as ${MR}=\max(r_1,r_2,\ldots,r_C)$, which indicates the percentage of the pixels of the dominating class in the superpixel.
 If $MR$ is less than a predefined threshold denoted as $\delta$, the superpixel can be regarded as a noisy one, and it is further segmented into more sub-superpixels. Specifically, for each noisy shape-adaptive superpixel, we apply the entropy rate superpixel method \cite{liu2011entropy} on the corresponding smallest enclosing square region, leading to $M$ more sub-superpixels for each square.
That is, given a superpixel, we first divide its smallest enclosing square area into multiple sub-regions, and then for each sub-region, we only extract pixels belonging to the input superpixel to generate  sub-superpixels.
\begin{figure*}[t]
     \centering
     \subfigure[]{
\begin{minipage}[t]{0.3\linewidth}
\centering
\includegraphics[width=1.0\textwidth , height=0.14\textheight]{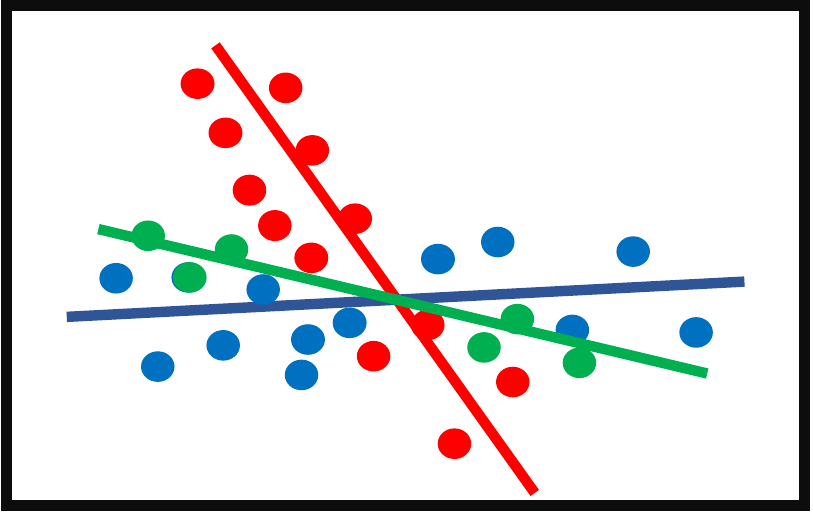}
\label{Fig:8}
\end{minipage}}
    \subfigure[]{
\begin{minipage}[t]{0.3\linewidth}
\centering
\includegraphics[width=1.0\textwidth , height=0.14\textheight]{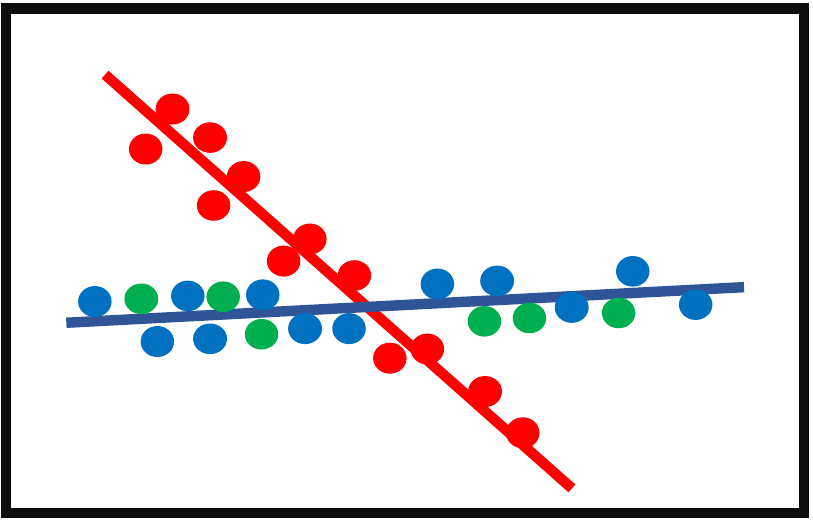}
\label{Fig:8}
\end{minipage}}
    \subfigure[]{
\begin{minipage}[t]{0.3\linewidth}
\centering
\includegraphics[width=1.0\textwidth , height=0.14\textheight]{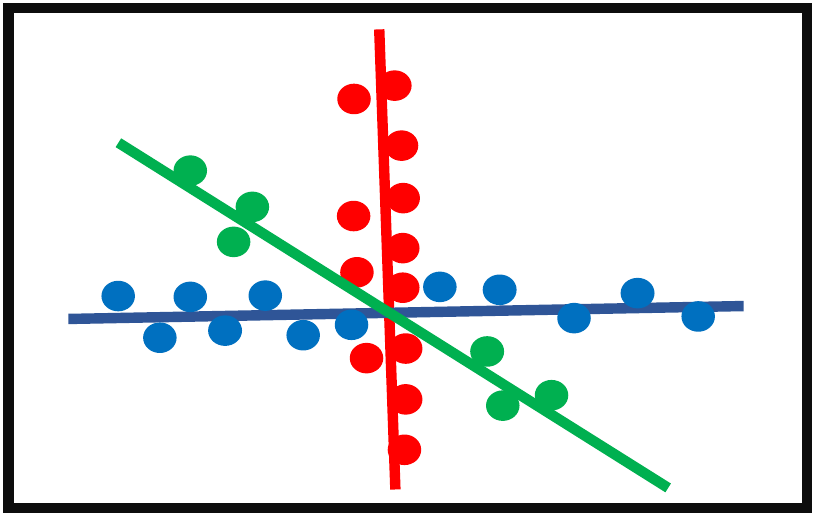}
\label{Fig:8}
\end{minipage}}
     \caption{Visual illustration of our motivation. Here we take three classes represented with red, green, and blue dots as an example. Moreover, the green and blue samples are grouped into an identical superpixel, and the green samples are much fewer than the blue samples. Each of three lines stands for the subspaces of the corresponding class. (a) Input data. (b) Results of applying robust low-rank approximation superpixel by superpixel (i.e., our DLRR with $\beta=0$).  (c) Results of our DLRR.}
     \label{fig:motivation}
 \end{figure*}

\subsection{Discriminative Low-rank Representation}\label{sec:DLR}

\subsubsection{Problem Formulation}

To suppress the spectral variations, one may intuitively explore its low-rank property by applying robust low-rank approximation on an HSI globally, i.e.,
\begin{eqnarray}\label{eq:rpca}
\begin{aligned}
\min_{[\LL,\E]} \|\LL\|_{*}+\gamma \|\E\|_1, \  s.t., \X=\LL+\E,
\end{aligned}
\end{eqnarray}
where $\|\cdot\|_*$ and $\|\cdot\|_1$ are the nuclear norm and $\ell_1$ norm of a matrix, respectively, $\gamma$ is a positive regularization parameter, $\LL\in\mathbb{R}^{b\times n}$ is the desired discriminative representation of $\X$, and $\mathbf{E}\in\mathbb{R}^{b\times n}$ corresponds to the variations.
However, such an intuitive and global manner fails to capture the local spatial structure sufficiently \cite{zhang2013hyperspectral,xu2015spectral}, i.e., a typical pixel and its neighbours often come from the same class.
 Moreover, the pixels of an HSI should be located in multiple subspaces rather than a single subspace, as it usually contains multiple classes. But Eq. (\ref{eq:rpca}) assumes that all pixels belong to a single subspace.

Being aware of these issues, we propose to explore the low-rank prior locally, i.e., the robust low-rank approximation is separately performed on each superpixel.
Although performing the robust low-rank approximation superpixel by superpixel can suppress the variations for increasing the intra-class similarity, the inter-class discriminability is not considered and it may be compromised. In addition, as a superpixel may contain more than one class, such a manner will also weaken the inter-class discriminability. Fig. \ref{fig:motivation}(b) visually illustrates such an issue.
To this end, we further propose to maximize the nuclear norm of the matrix containing all superpixels to promote the inter-class discriminability. In other words, we expect the distance between any two subspaces (one subspace for one class) is as far as possible such that the class with only a few pixels will not be absorbed by the class with much more pixels.
The proposed discriminative low-rank representation (DLRR) is finally formulated as
\begin{eqnarray}\label{eq:global}
\begin{aligned}
\min_{[\LL,\E]}& \sum_{i=1}^{S}\|\LL_i\|_{*}+\lambda \|\E\|_{1}-\beta \|\LL\|_{*}\\
s.t.\ & \X=\LL+\E,\  \X=[\X_1,\X_2,\ldots,\X_S]\\
&\LL=[\LL_1,\LL_2,\ldots,\LL_S],\  \E=[\E_1,\E_2,\ldots,\E_S],
\end{aligned}
\end{eqnarray}
where $\lambda$ and $\beta$ are two positive regularization parameters, $S$ is the total number of superpixels, $\X_i$ is the $i$-th superpixel, and $\mathbf{L}_i$ and $\mathbf{E}_i$ are the decomposed components corresponding to $\mathbf{X}_i$. Fig. \ref{fig:motivation}(c) visually illustrates the behavior of our DLRR.


\subsubsection{Numerical Solution}
By introducing an auxiliary variable $\J\in\mathbb{R}^{b\times n}$, the non-convex problem in Eq. (\ref{eq:global}) can be equivalently reformulated as
\begin{eqnarray}\label{eq:LEJ}
\begin{aligned}
\min_{[\LL,\E,\J]} &\sum_{i=1}^{S}\|\LL_i\|_{*}+\lambda \|\E\|_{1}-\beta \|\J\|_{*}\\
s.t.\ & \X=\LL+\E,\ \J=\LL\\
&\X=[\X_1,\X_2,\ldots,\X_S],\ \LL=[\LL_1,\LL_2,\ldots,\LL_S]\\
&\E=[\E_1,\E_2,\ldots,\E_S].
\end{aligned}
\end{eqnarray}
We adopt the inexact augmented Lagrangian multiplier
(IALM) \cite{lin2010augmented} to solve Eq. (\ref{eq:LEJ}) in an iterative manner, in which the unknown variables $\LL_i$, $\E$, and $\J$ are alternatively updated by solving the three subproblems in each iteration. The augmented Lagrange form of Eq. (\ref{eq:LEJ}) can be written as
\begin{eqnarray}\label{eq:obj_fun}
\begin{aligned}
\min_{[\LL_i,\E,\J]}&\sum_{i=1}^{S}\|\LL_i\|_{*}+\lambda \|\E\|_{1}-\beta \|\J\|_{*}\\
&+\frac{\mu}{2}\|\X-\LL-\E+\frac{\Y_1}{\mu}\|_F^2+\frac{\mu}{2}\|\J-\LL+\frac{\Y_2}{\mu}\|_{F}^2,
\end{aligned}
\end{eqnarray}
where $\Y_1 \in \mathbb{R}^{b\times n}$ and $\Y_2 \in \mathbb{R}^{b\times n}$ are the Lagrange multipliers, $\|\cdot\|_F$ is the Frobenius norm of a matrix, and $\mu$ is the penalty parameter. The three subproblems are given as follows. See Algorithm \ref{algorithm:lrgr} for the pseudocode.

\paragraph{}
With fixed $\mathbf{E}$ and $\mathbf{J}$, the $\{\mathbf{L}_i\}_{i=1}^S$ subproblem is written as
\begin{eqnarray}\label{eq:Lis}
\begin{aligned}
\min_{[\LL_1,\LL_2,\ldots,\LL_S]}&\sum_{i=1}^S\|\LL_i\|_{*}+\frac{\mu}{2}\|\X-\LL-\E+\frac{\Y_1}{\mu}\|_F^2\\
&+\frac{\mu}{2}\|\J-\LL+\frac{\Y_2}{\mu}\|_{F}^2.
\end{aligned}
\end{eqnarray}
The $S$ variables  $\{\mathbf{L}_i\}_{i=1}^S$ can be independently optimized by solving
\begin{eqnarray}\label{eq:Li}
\begin{aligned}
&\min_{[\LL_i]}\|\LL_i\|_{*}+\mu\|\LL_i-\W_i\|_{F}^2,
\end{aligned}
\end{eqnarray}
where $\W_{i}=\frac{1}{2}{\{(\X_i-\E_i+\frac{\Y_{1,i}}{\mu})+(\J_i+\frac{\Y_{2,i}}{\mu})\}}$, with $\mathbf{J}_i$, $\Y_{1,i}$ and $\Y_{2,i}$ being the $i$-th block of $\J$, $\Y_1$, and $\Y_2$, respectively. Let the singular value decomposition of $\W_i$ be $\U_i\mathbf{\Sigma}_{i}\V_i^\mathsf{T}$, where $\mathbf{\Sigma}_{i}$ is a diagonal matrix with the singular values on its diagonal, and the columns of $\U_i$ and $\mathbf{V}_i$ are the left and right singular vectors, respectively. Eq. (\ref{eq:Li}) has a closed-form solution \cite{cai2010singular} as \begin{equation}\LL_i=\U_i\mathcal{S}_{{(2\mu)}^{-1}}(\mathbf{\Sigma}_{i})\V_i^\mathsf{T},\end{equation} where
\begin{eqnarray}
\begin{aligned}
\mathcal{S}_{\varepsilon}(x):={\rm sgn}(x)\max(|x|-\varepsilon,0)\\\nonumber
=\left\{\begin{matrix}
\begin{aligned}
&x-\varepsilon, & \text{if}\par\ & x\par>\par\varepsilon,\\\nonumber
&x+\varepsilon, & \text{if}\par\ & x\par<\par-\varepsilon,\\\nonumber
&0, & \text{if}\ &\par-\varepsilon\par\leq\par x\par\leq\par\varepsilon,\nonumber
\end{aligned}
\end{matrix}\right.
\end{aligned}
\end{eqnarray}
is the soft thresholding operator in element-wise, and ${\rm sgn}(\cdot)$ takes the sign of a real number.

\paragraph{}
Fixing $\LL$ and $\J$, the $\E$ subproblem can be rewritten as

\begin{eqnarray}\label{eq:E}
\begin{aligned}
\min_{\E} \lambda \|\E\|_{1}+\frac{\mu}{2}\|\D-\E\|_F^2,
\end{aligned}
\end{eqnarray}
 where $\D=\X-\LL+\frac{\Y_{1}}{\mu}$. Eq. (\ref{eq:E}) has a closed-form solution \cite{lin2017low,lin2011linearized}: \begin{equation}\E=\mathcal{S}_{\lambda({\mu})^{-1}}(\D).\end{equation}.

\paragraph{}
Fixing $\LL$ and $\E$, the $\J$ subproblem becomes

\begin{eqnarray}\label{eq:J}
\begin{aligned}
\min_{\J}-\beta \|\J\|_*+\frac{\mu}{2}\|\J-\LL+\frac{\Y_2}{\mu}\|_F^2.
\end{aligned}
\end{eqnarray}
Note that Eq. (\ref{eq:J}), which is sum of a concave term and a convex term,  is a non-convex problem. To solve it, we approximate the concave term, i.e.,$-\beta\|\J\|_*$, using its first order Tayor expansion at the value of $\J$ obtained in the previous iteration, denoted as $\widehat{\J}$.
According to the subgradient of the nuclear norm of a matrix given in Theorem 1 \cite{watson1992characterization}, Eq. (\ref{eq:J}) can be approximately reformulated as
\begin{eqnarray}\label{eq:Jcon}
\begin{aligned}
\min_{\J}-\beta \textrm{Tr}(\partial\|\widehat{\J}\|_*\J^\mathsf{T})+\frac{\mu}{2}\|\J-\LL+\frac{\Y_2}{\mu}\|_F^2,
\end{aligned}
\end{eqnarray}
where $\J^\mathsf{T}$ represents the transpose of $\J$.
By setting the first order derivative of the objective function with respect to $\mathbf{J}$ to be zero, the optimal solution of Eq. (\ref{eq:Jcon}) is achieved, leading to \begin{equation}\J=\frac{\beta}{\mu}\partial\|\widehat{\J}\|_{*}-\frac{\Y_2}{\mu}+\LL.\end{equation}

{\textit{Theorem 1} \cite{watson1992characterization}:
Let $\A \in \mathbb{R}^{m_1 \times m_2}$ be an arbitrary matrix and $\U\mathbf{\Sigma} \V^\mathsf{T}$ be its SVD.
It is known \cite{watson1992characterization} that
$\partial\|\A\|_*=\{\U\V^\mathsf{T}+\Z : \Z \in \mathbb{R}^{m_1 \times m_2}, \U^\mathsf{T}\Z=0, \Z\V=0, \|\Z\|_2 \leq 1\}$, where $\|\cdot\|_2$ is the $\ell_2$ norm of a matrix.
}

\paragraph{}
Updating the Lagrangian Parameters $\Y_1$, $\Y_2$ and penalty parameter $\mu$ as

\begin{align}
&\Y_1^{\rm{iter+1}}=\Y_1^{\rm{iter}}+\mu^{\rm{iter}}(\X-\LL-\E),\\
&\Y_2^{\rm{iter+1}}=\Y_2^{\rm{iter}}+\mu^{\rm{iter}}(\J-\LL),\\
&\mu^{\rm{iter+1}}=\min (\mu_{\max},\rho\mu^{\rm{iter}}),
\end{align}
where iter is the iteration index, and the parameter $\rho > 1$ improves the convergence rate.

The global convergence of IALM when applied for optimizing a convex problem with at most two blocks has been theoretically proven~\cite{lin2010augmented,lin2015global,lin2018global}. However,
there are three blocks in Algorithm \ref{algorithm:lrgr}, and to the best of our knowledge, the theoretical global convergence of the IALM solver with three or more blocks is still unsolved~\cite{lin2015global}. Fortunately,
thanks to the closed-form solution for each subproblem,
we observe that Algorithm \ref{algorithm:lrgr} empirically converges well over all the employed datasets (see Fig. \ref{fig:convergence}). 

\begin{algorithm}[t]
\caption{The IALM-based optimization algorithm for solving DLRR in Eq. (\ref{eq:global})}\label{algorithm:lrgr}
\begin{algorithmic}[1]
\renewcommand{\algorithmicrequire}{\textbf{Input:}}
		\renewcommand{\algorithmicensure}{\textbf{Output:}}
		\REQUIRE $\X$, $\X_i$ and parameters $\lambda$, $\beta$, $S$.
        \ENSURE  $\LL$, variations $\E$
\STATE \textbf{Initialize:} $\mu \leftarrow 10^{-4}$, $\epsilon \leftarrow 10^{-6}$, $\rho=1.1$, $\mu_{\max}=10^{12}$, $\LL_i \leftarrow \bf{0}$, $\E \leftarrow \bf{0}$, $\J \leftarrow \bf{0}$, $\Y_1 \leftarrow \bf{0}$ and $\Y_2 \leftarrow \bf{0}$.
\WHILE{not converged}
\FOR{$i$=1:$S$}
 \STATE {Update $\LL_i$} using Eq. (\ref{eq:Li})
\ENDFOR
\STATE Update $\E$ using Eq. (\ref{eq:E})
\STATE Update $\J$ using Eq. (\ref{eq:J})
\STATE Update $\Y_1$, $\Y_2$ using Eq. (14) and Eq. (15), respectively
\STATE Update $\mu$ using Eq. (16)
\STATE Check the convergence condition: $\|\X-\LL-\E\|_{\infty} \le \epsilon$ and $\|\LL-\J\|_{\infty} \le \epsilon$, where $\|\cdot\|_{\infty}$ is the max norm of a matrix.
\ENDWHILE
\end{algorithmic}
\end{algorithm}
\section{Experiments}
\begin{figure*}[!tb]
\centering
\subfigure[]{
\begin{minipage}[t]{0.25\linewidth}
\centering
\includegraphics[width=1.0\textwidth , height=0.14\textheight]{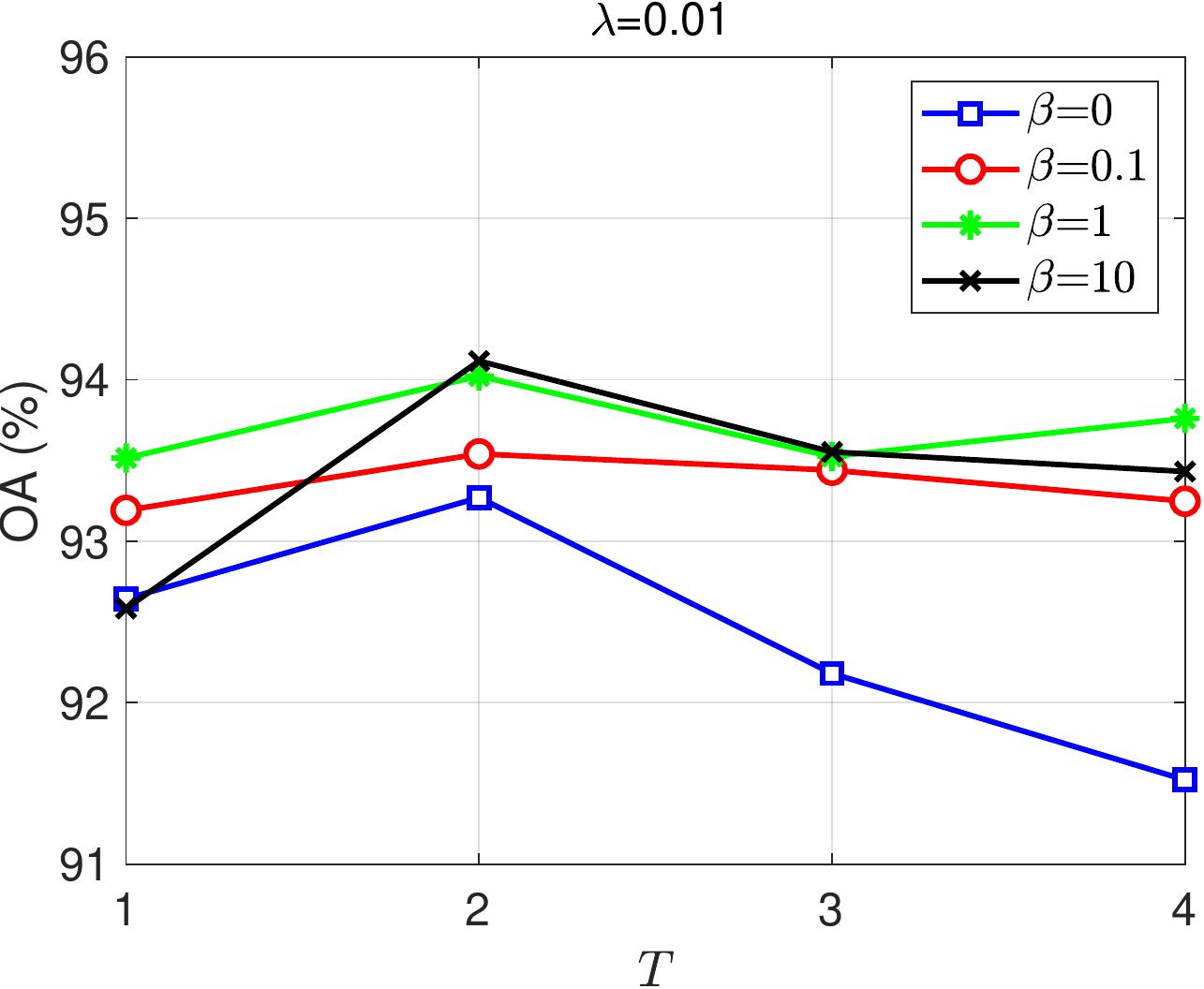}
\label{Fig:8}
\end{minipage}}
\hspace{10pt}
\subfigure[]{
\begin{minipage}[t]{0.25\linewidth}
\centering
\includegraphics[width=1.0\textwidth, height=0.14\textheight]{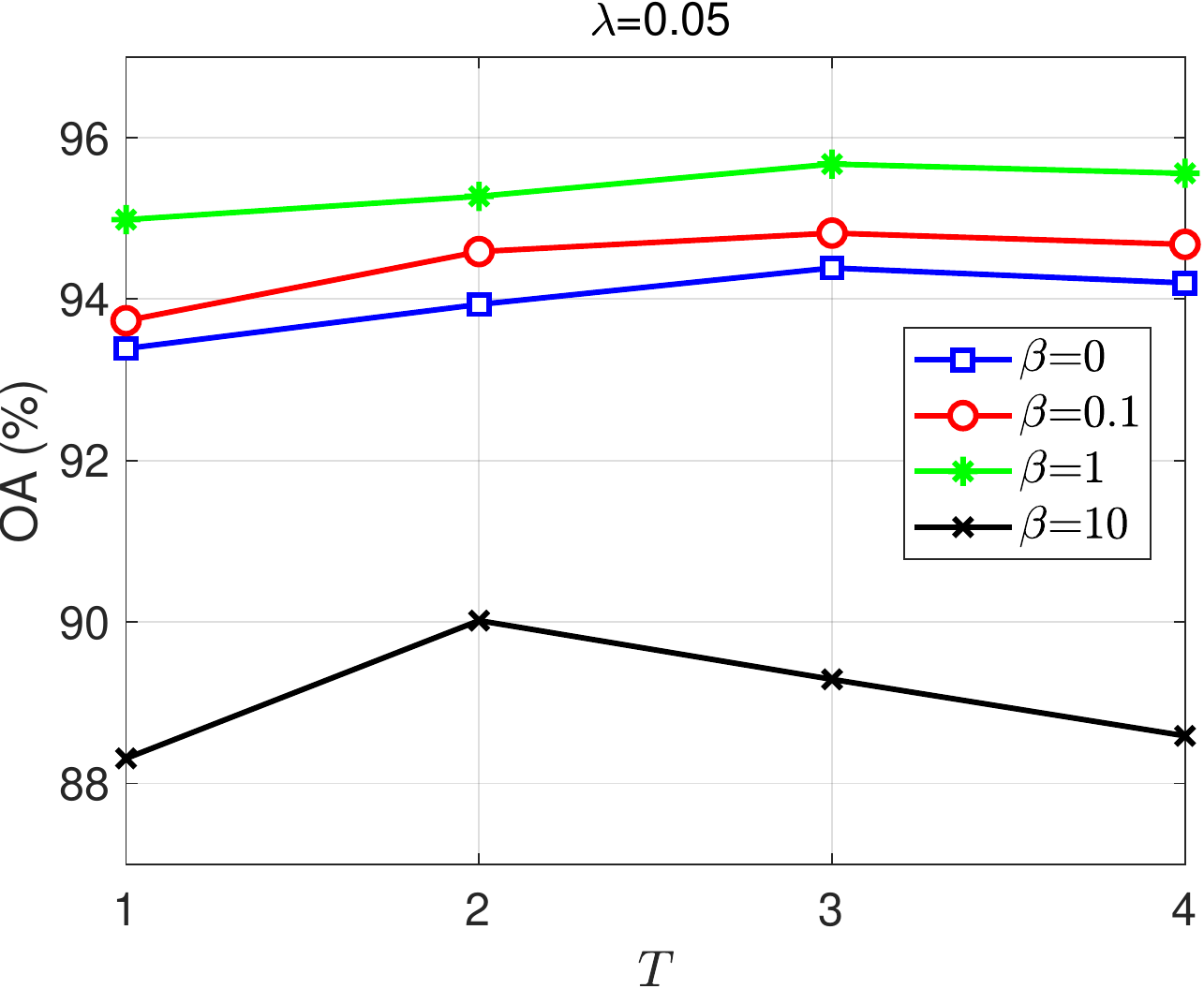}
\label{Fig:8}
\end{minipage}}
\hspace{10pt}
\subfigure[]{
\begin{minipage}[t]{0.25\linewidth}
\centering
\includegraphics[width=1.0\textwidth, height=0.14\textheight]{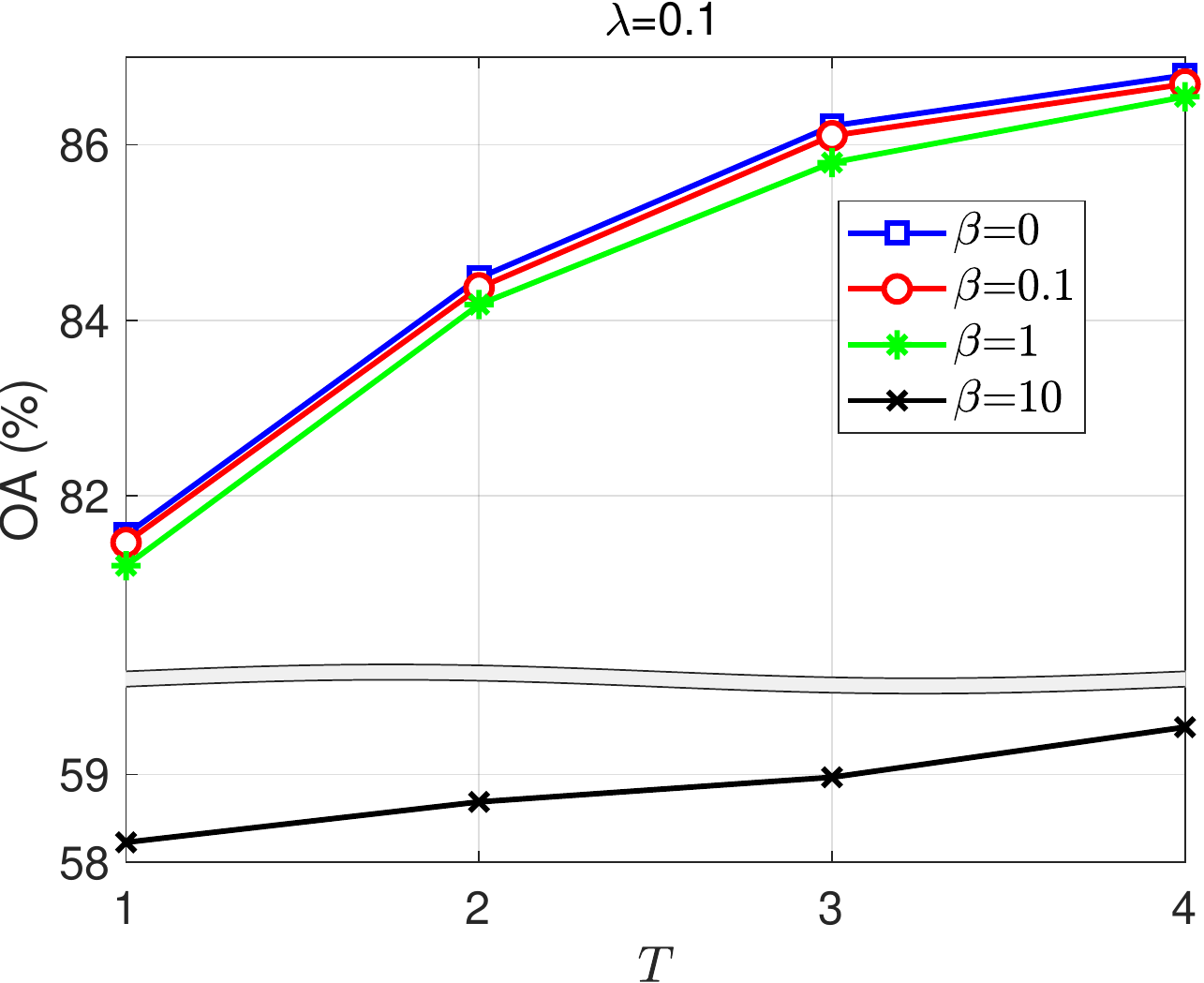}
\label{Fig:8}
\end{minipage}}

\subfigure[]{
\begin{minipage}[t]{0.25\linewidth}
\centering
\includegraphics[width=1.0\textwidth , height=0.14\textheight]{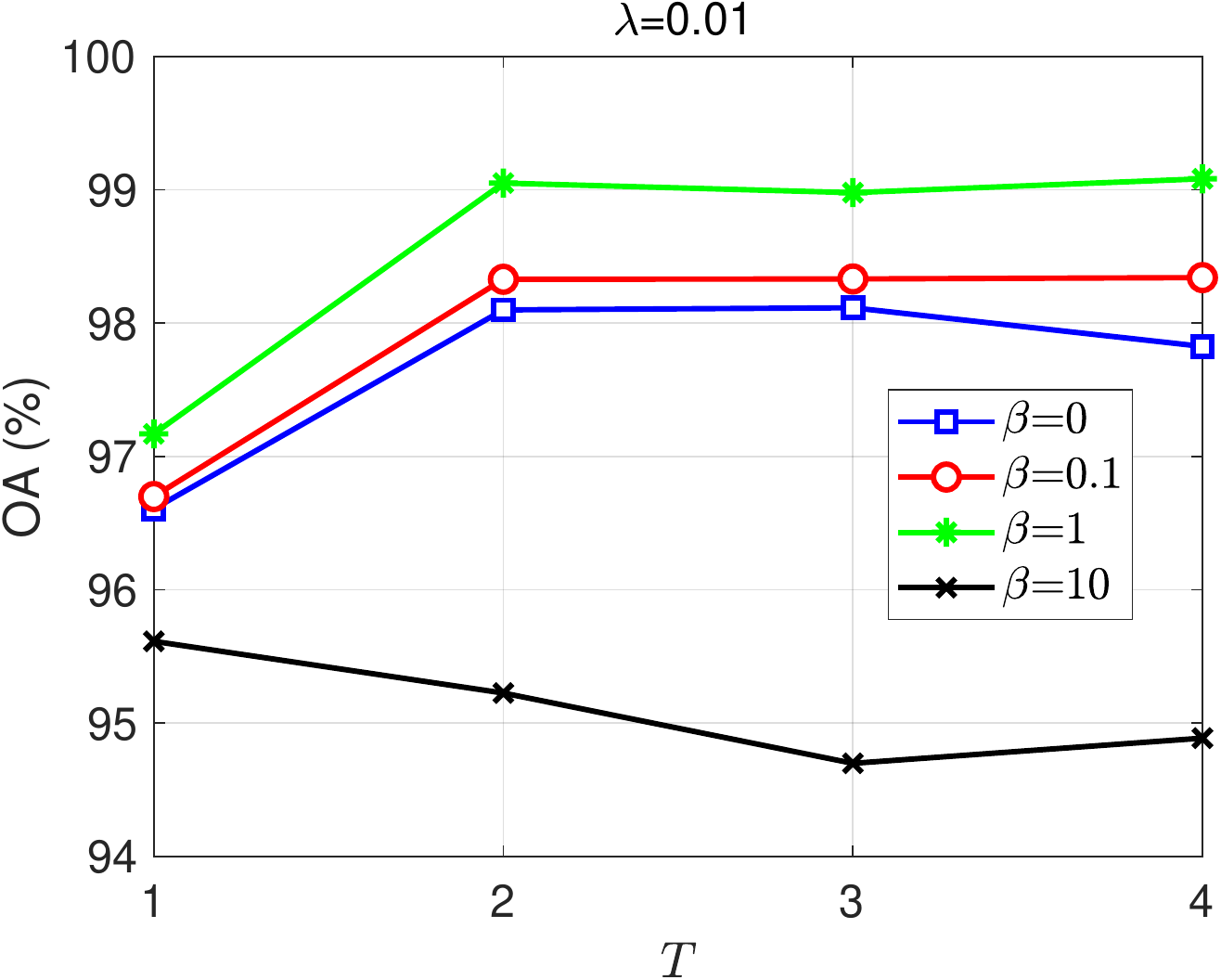}
\label{Fig:8}
\end{minipage}}
\hspace{10pt}
\subfigure[]{
\begin{minipage}[t]{0.25\linewidth}
\centering
\includegraphics[width=1.0\textwidth, height=0.14\textheight]{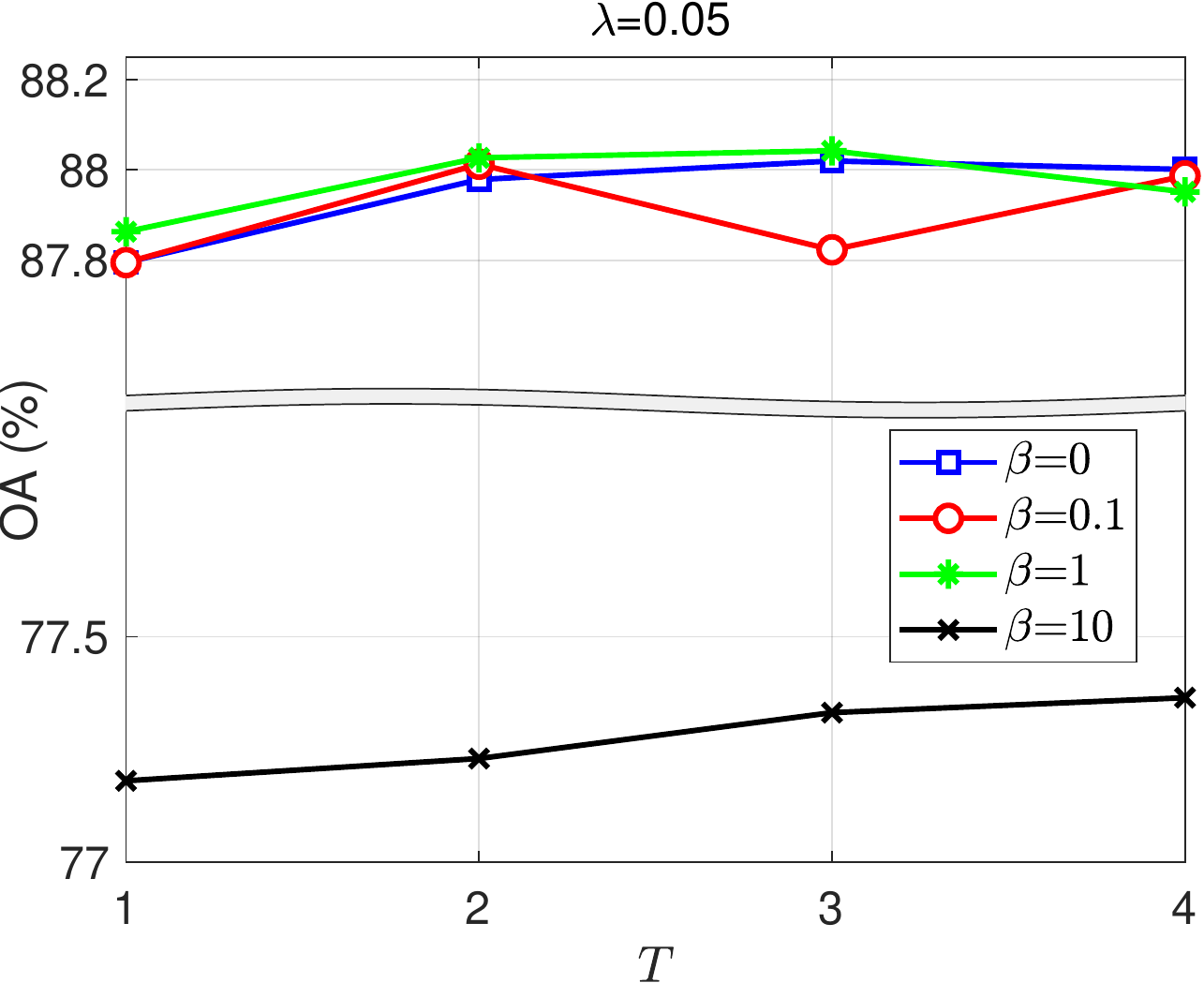}
\label{Fig:8}
\end{minipage}}
\hspace{10pt}
\subfigure[]{
\begin{minipage}[t]{0.25\linewidth}
\centering
\includegraphics[width=1.0\textwidth, height=0.14\textheight]{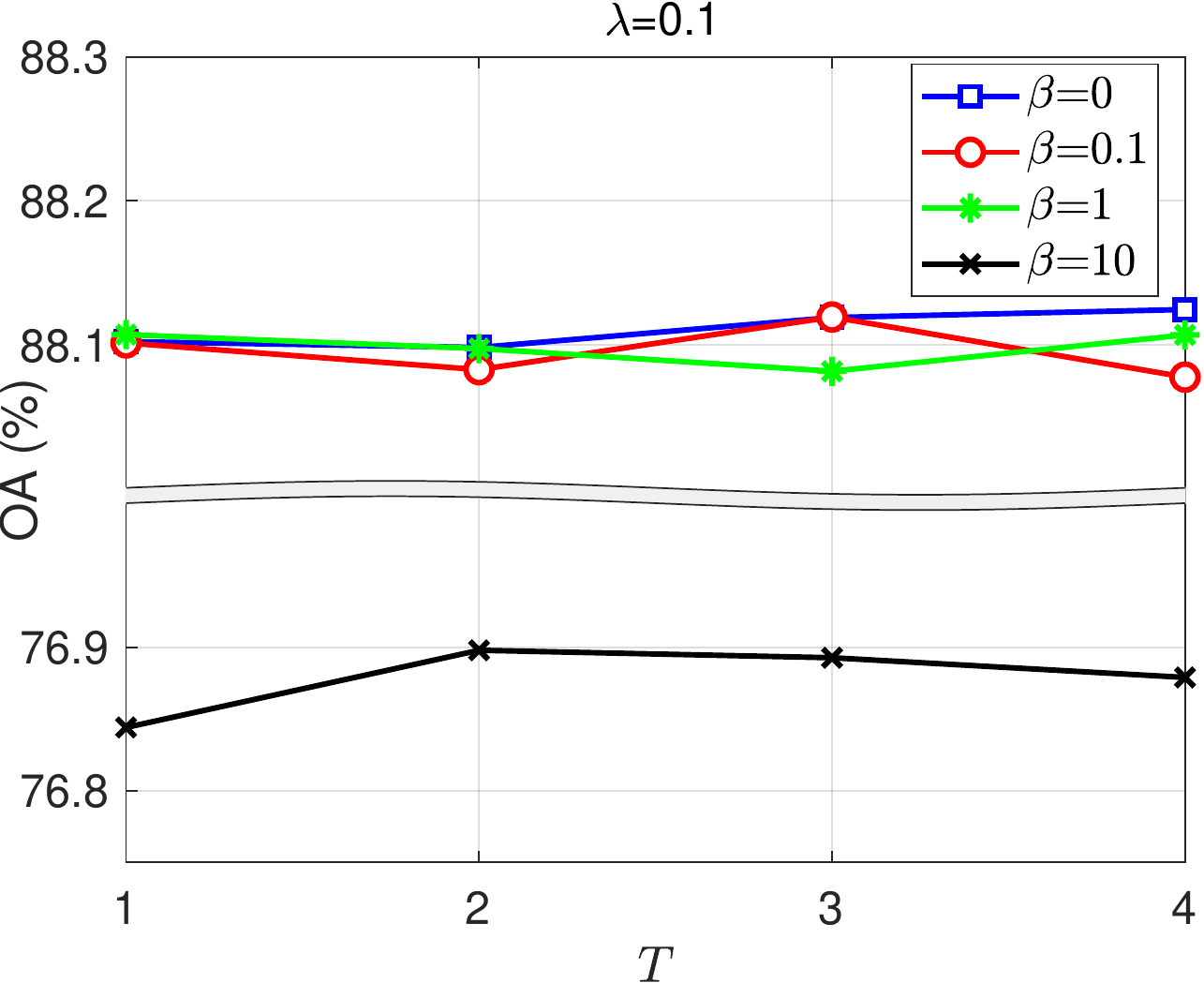}
\label{Fig:8}
\end{minipage}}

\subfigure[]{
\begin{minipage}[t]{0.25\linewidth}
\centering
\includegraphics[width=1.0\textwidth , height=0.14\textheight]{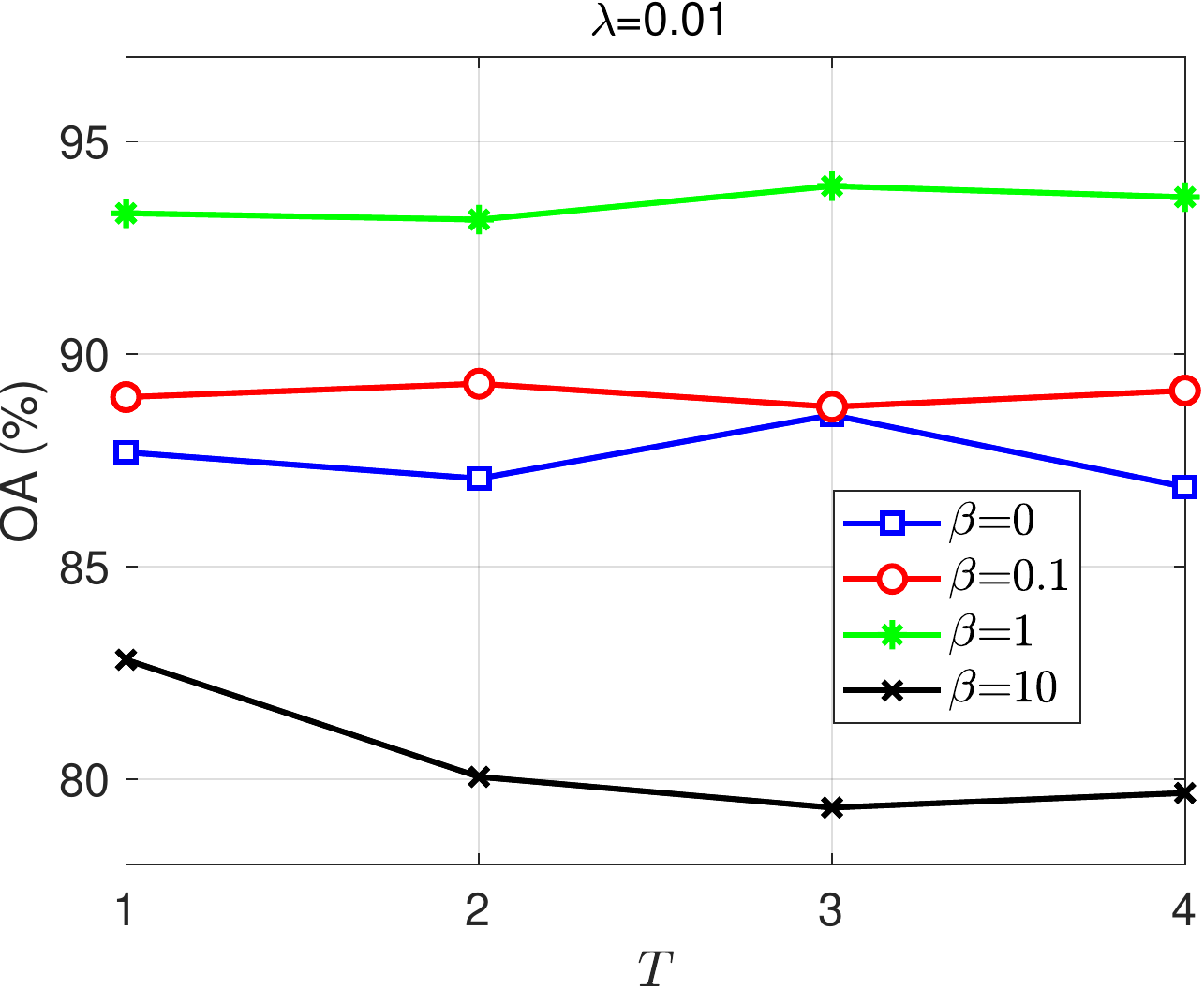}
\label{Fig:8}
\end{minipage}}
\hspace{10pt}
\subfigure[]{
\begin{minipage}[t]{0.25\linewidth}
\centering
\includegraphics[width=1.0\textwidth, height=0.14\textheight]{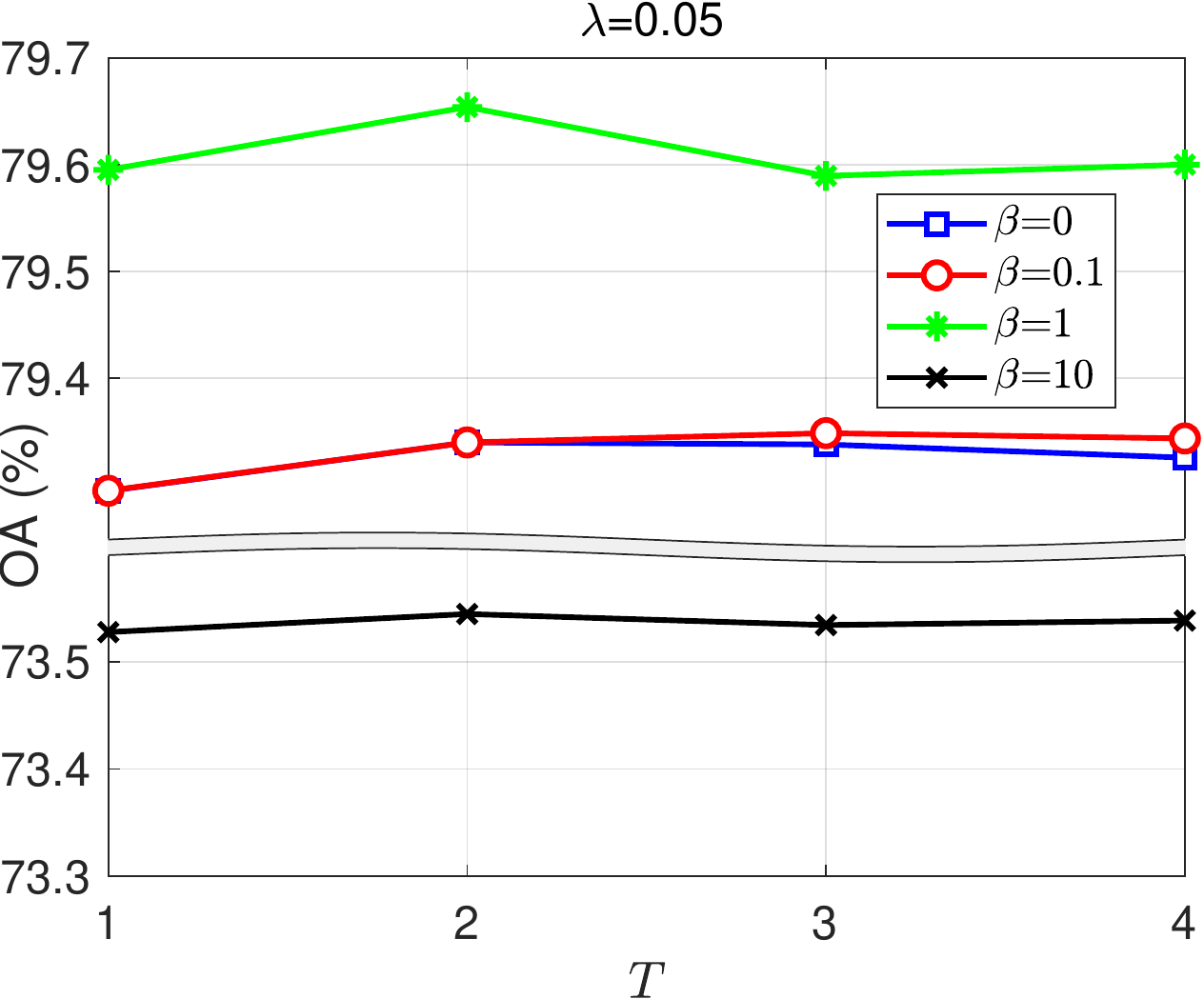}
\label{Fig:8}
\end{minipage}}
\hspace{10pt}
\subfigure[]{
\begin{minipage}[t]{0.25\linewidth}
\centering
\includegraphics[width=1.0\textwidth, height=0.14\textheight]{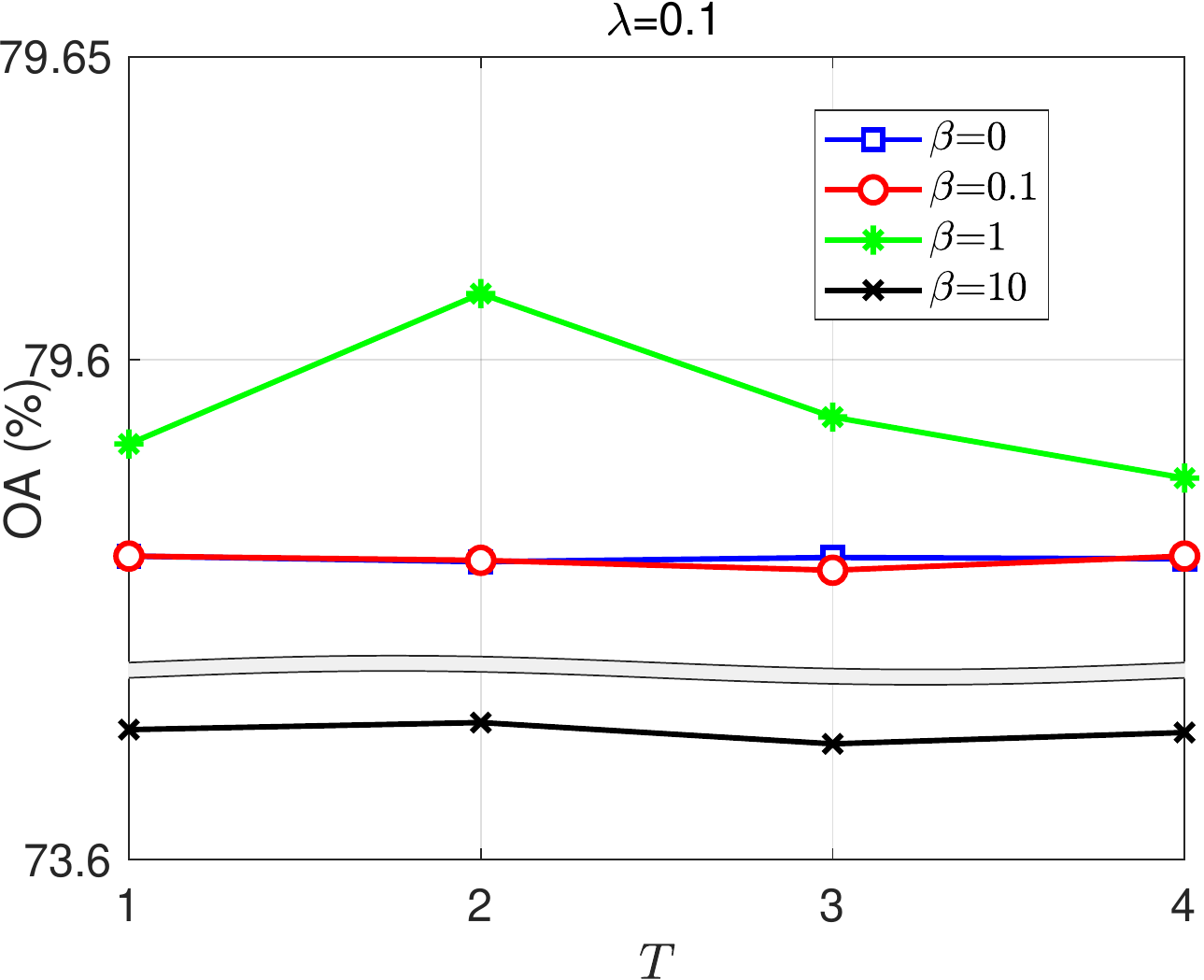}
\label{Fig:8}
\end{minipage}}
\caption{Illustration of the effect of the values of $\lambda$ and $\beta$ and the number of iterations $T$ on the performance of SP-DLRR in terms of OA.
Here the percentage of labeled pixels per class equals to $5\%$, $0.5\%$ and $0.2\%$ for \textit{Indian Pines}, \textit{Salinas Valley}, and  \textit{Pavia University} respectively. (a)-(c) for \textit{Indian Pines}, (d)-(f) for \textit{Salinas Valley}, and (g)-(i) for \textit{Pavia University}.}
\label{fig:parameteranalysisoa}
\end{figure*}

\begin{figure*}[!tb]
\centering
\subfigure[]{
\begin{minipage}[t]{0.25\linewidth}
\centering
\includegraphics[width=1.0\textwidth , height=0.14\textheight]{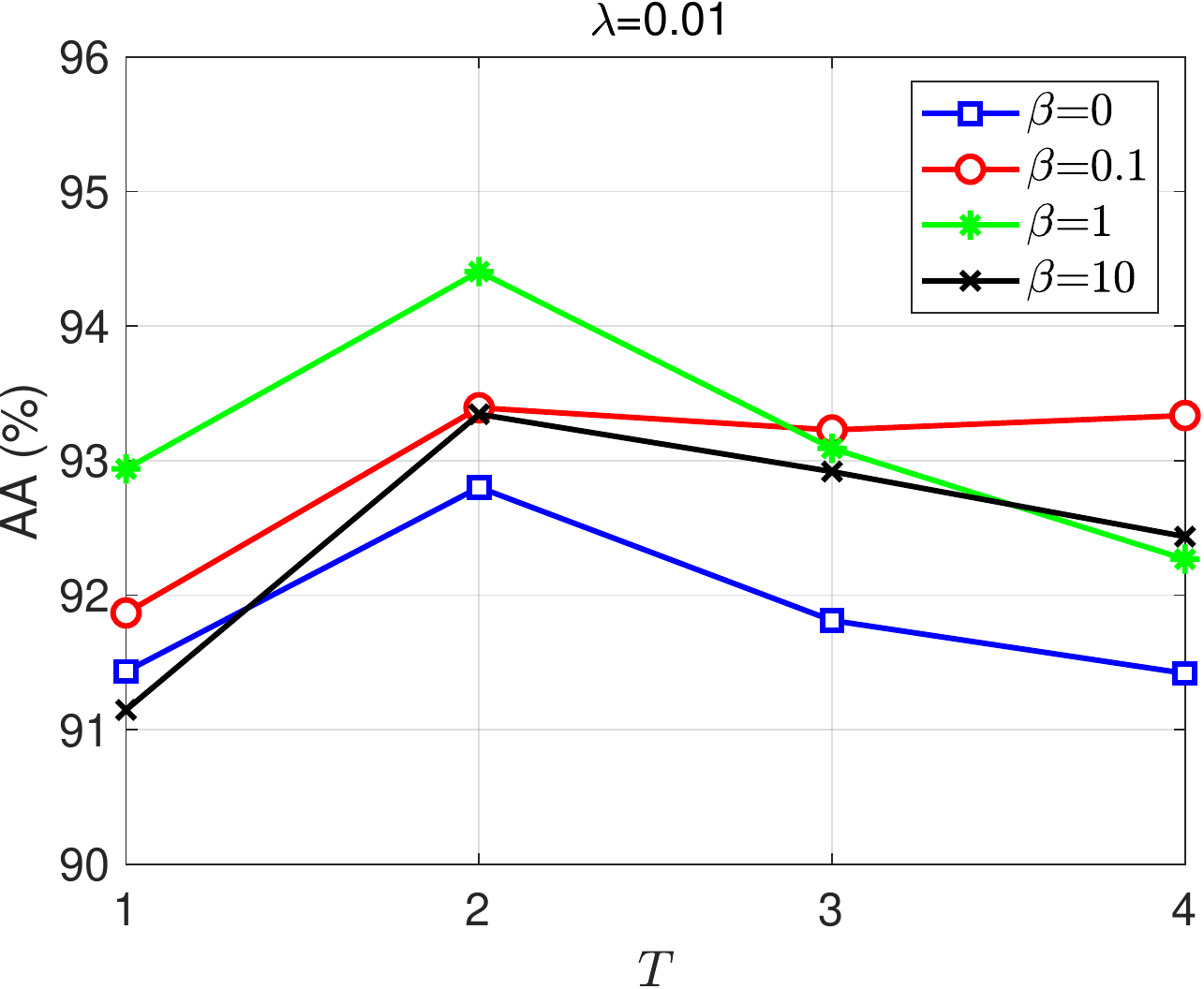}
\end{minipage}}
\hspace{10pt}
\centering
\subfigure[]{
\begin{minipage}[t]{0.25\linewidth}
\centering
\includegraphics[width=1.0\textwidth , height=0.14\textheight]{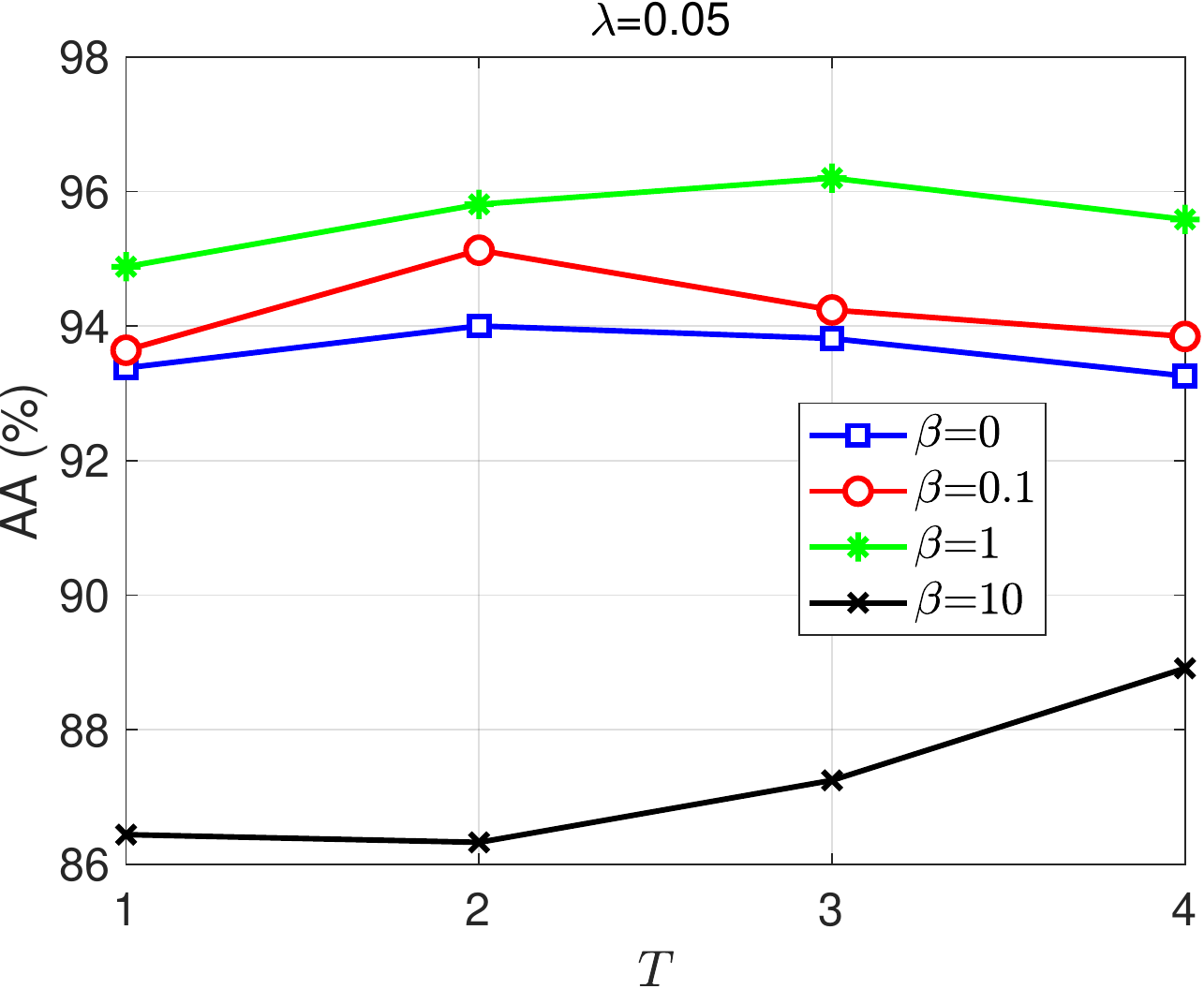}
\end{minipage}}
\hspace{10pt}
\centering
\subfigure[]{
\begin{minipage}[t]{0.25\linewidth}
\centering
\includegraphics[width=1.0\textwidth , height=0.14\textheight]{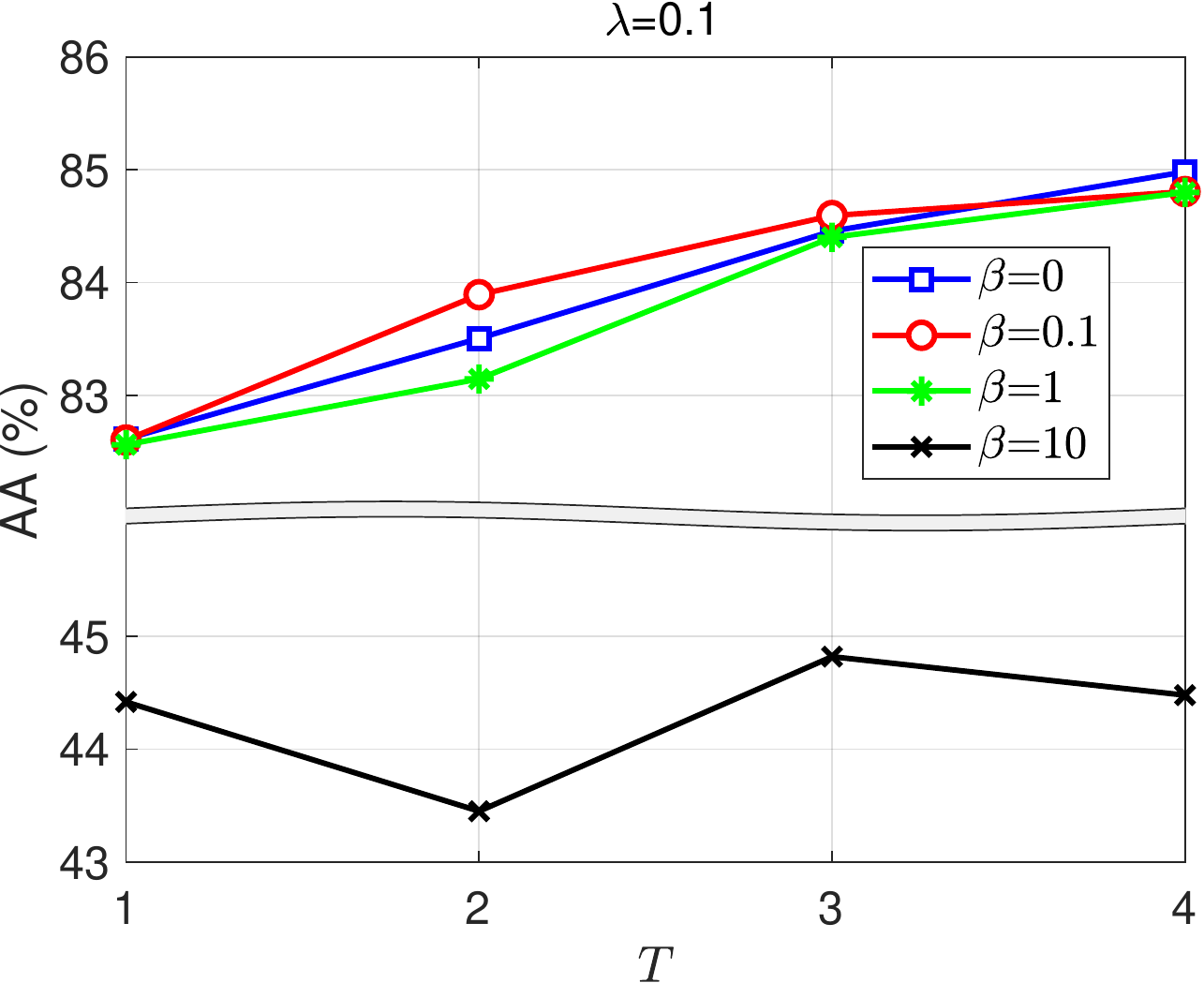}
\label{Fig:8}
\end{minipage}}
\subfigure[]{
\begin{minipage}[t]{0.25\linewidth}
\centering
\includegraphics[width=1.0\textwidth , height=0.14\textheight]{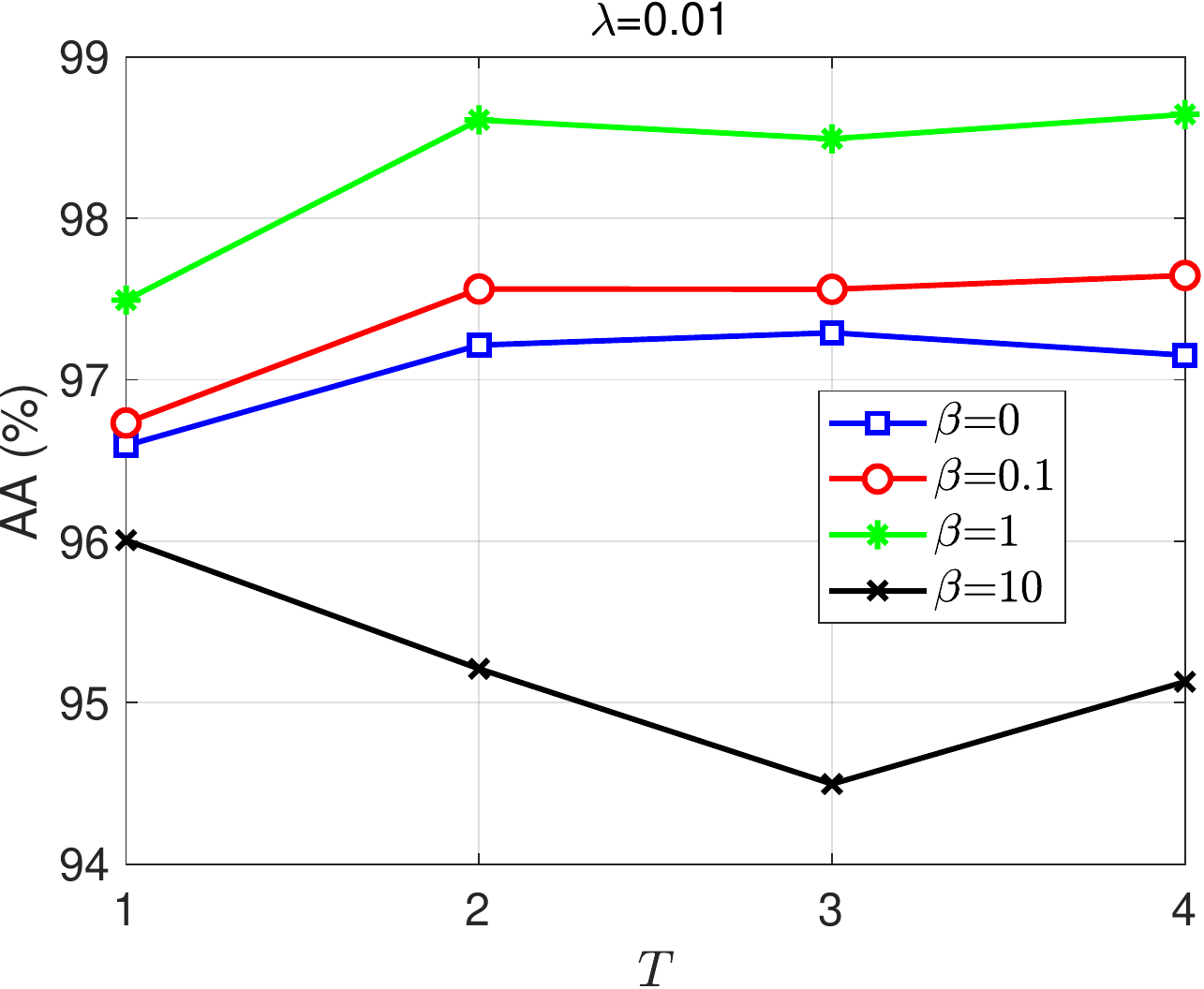}
\label{Fig:8}
\end{minipage}}
\hspace{10pt}
\subfigure[]{
\begin{minipage}[t]{0.25\linewidth}
\centering
\includegraphics[width=1.0\textwidth, height=0.14\textheight]{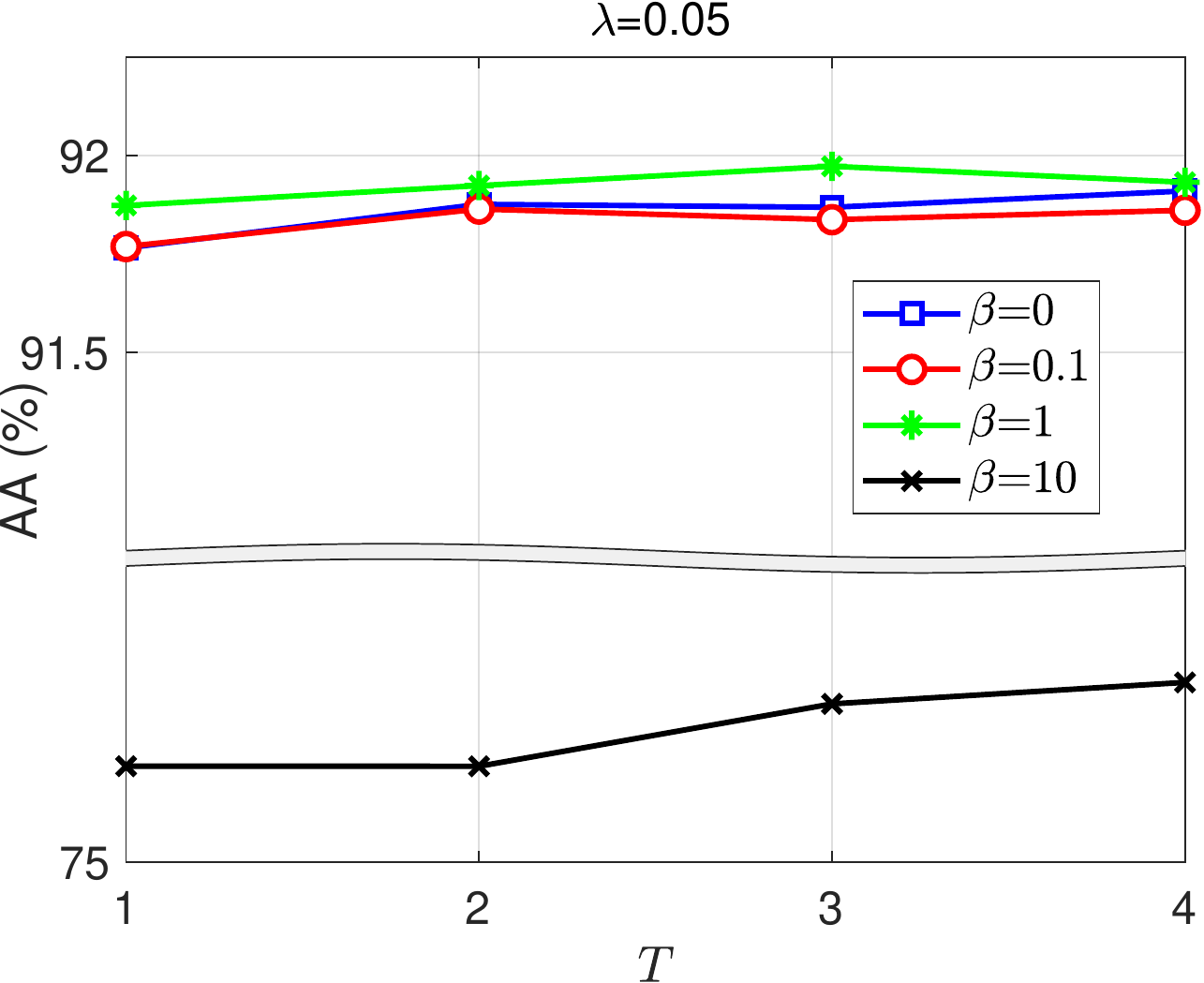}
\label{Fig:8}
\end{minipage}}
\hspace{10pt}
\subfigure[]{
\begin{minipage}[t]{0.25\linewidth}
\centering
\includegraphics[width=1.0\textwidth, height=0.14\textheight]{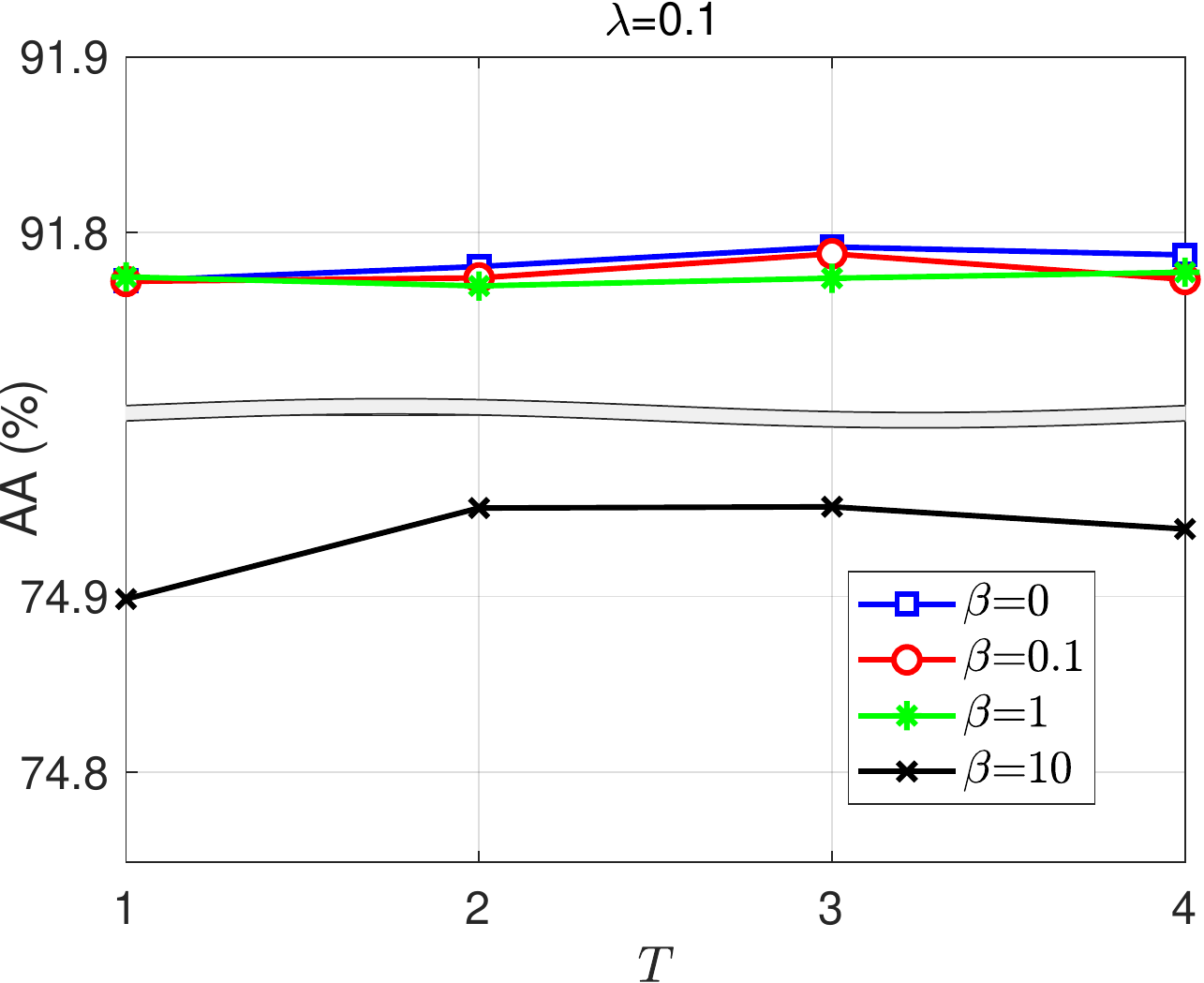}
\end{minipage}}

\subfigure[]{
\begin{minipage}[t]{0.25\linewidth}
\centering
\includegraphics[width=1.0\textwidth , height=0.14\textheight]{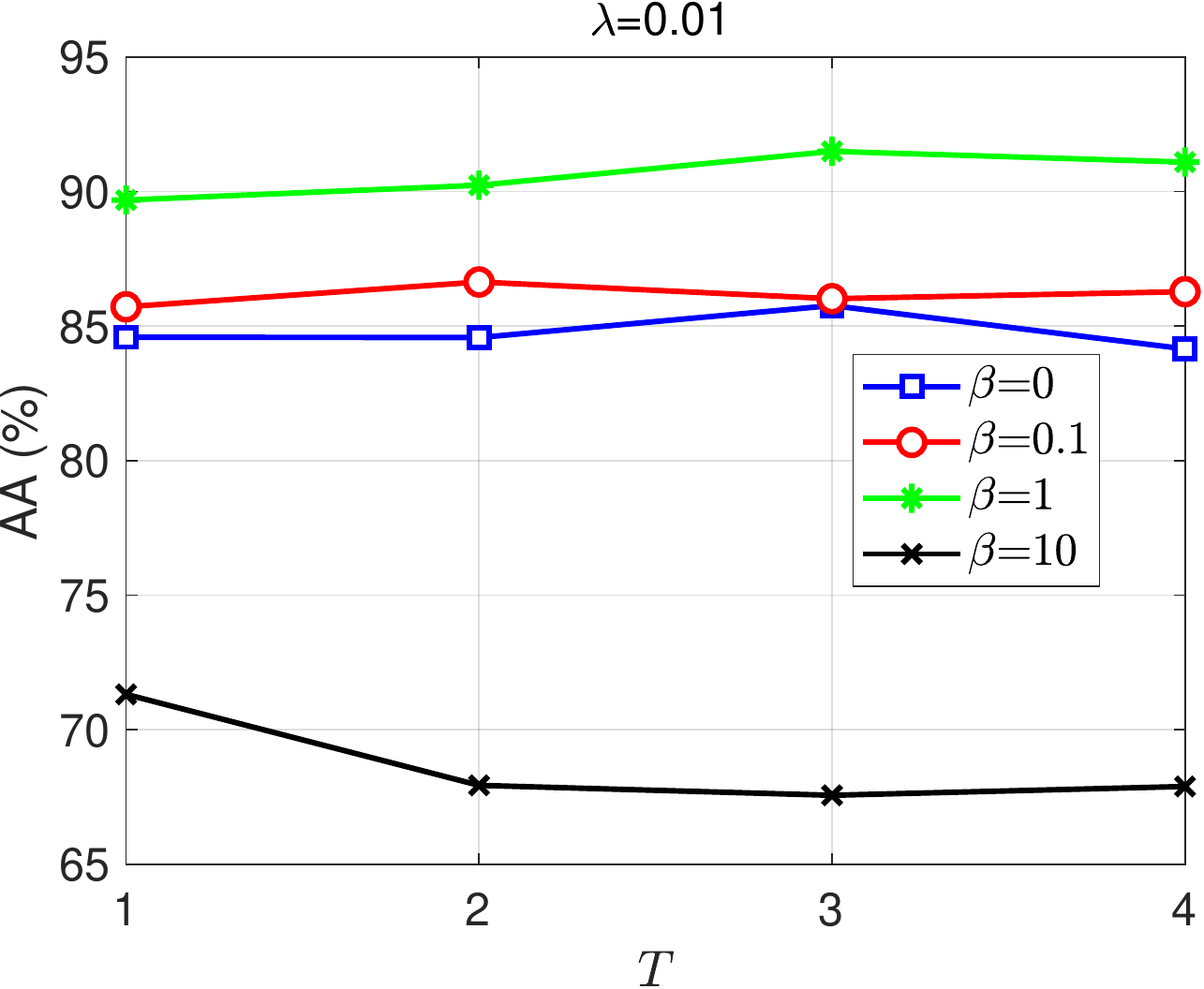}
\end{minipage}}
\hspace{10pt}
\subfigure[]{
\begin{minipage}[t]{0.25\linewidth}
\centering
\includegraphics[width=1.0\textwidth, height=0.14\textheight]{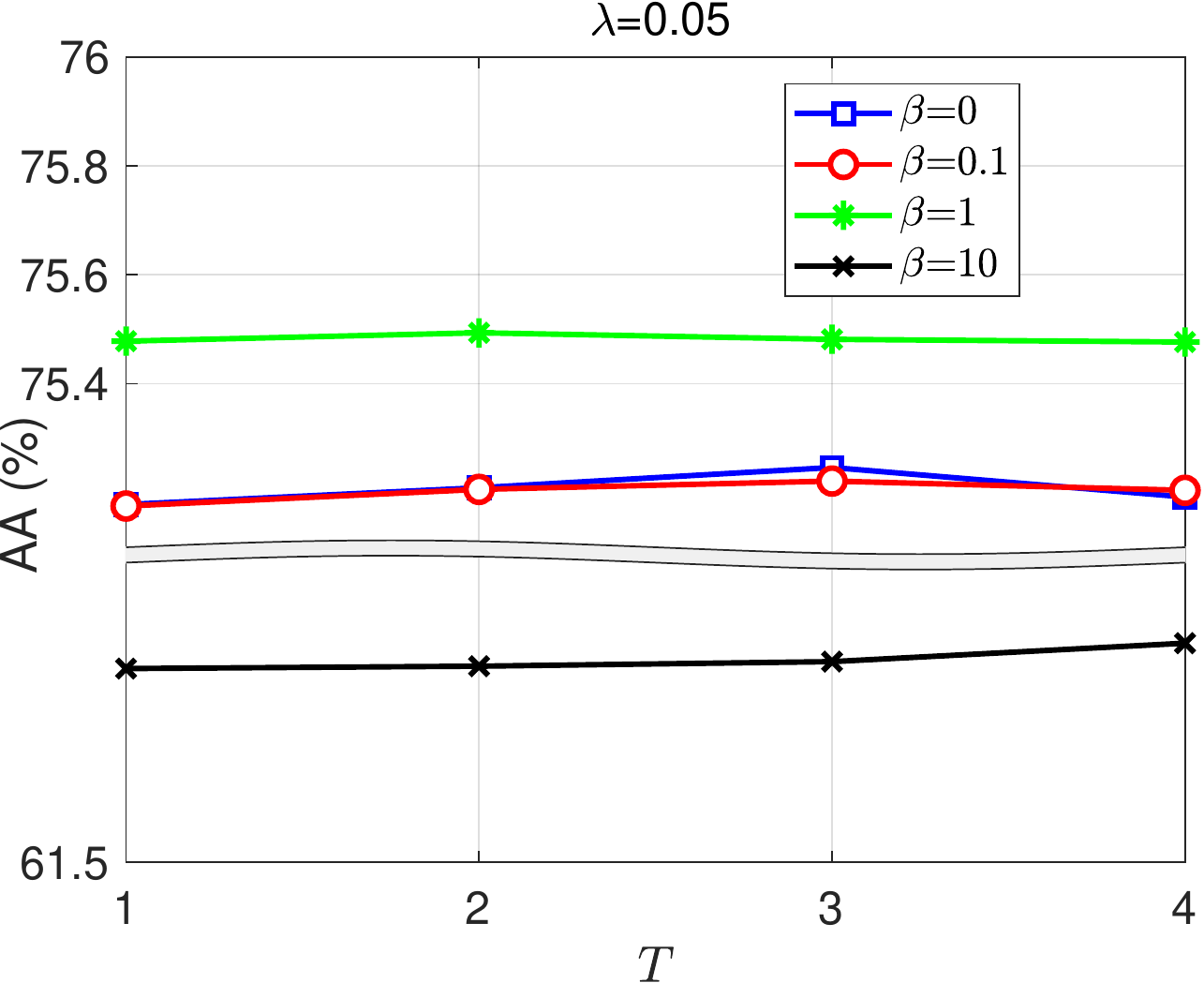}
\end{minipage}}
\hspace{10pt}
\subfigure[]{
\begin{minipage}[t]{0.25\linewidth}
\centering
\includegraphics[width=1.0\textwidth, height=0.14\textheight]{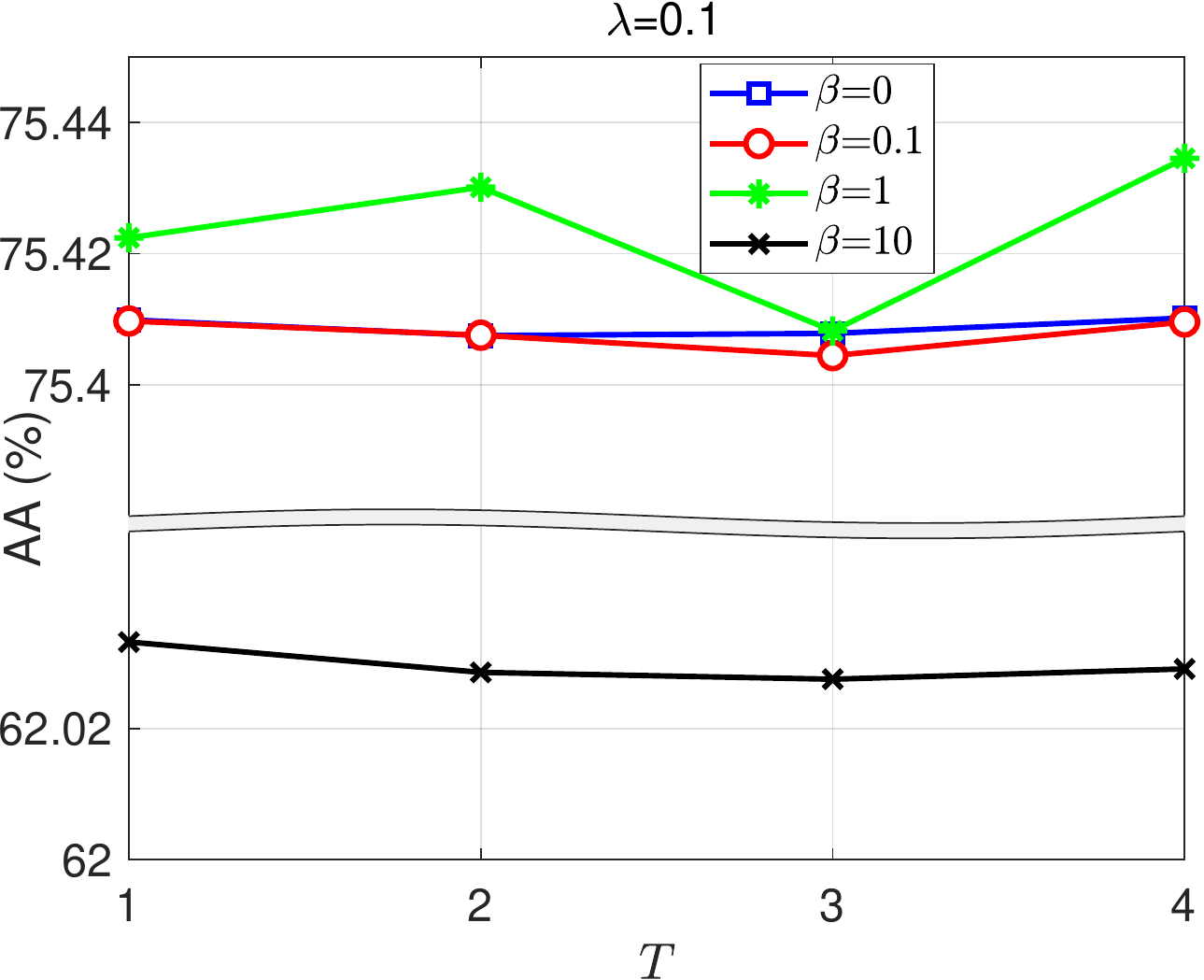}
\label{Fig:8}
\end{minipage}}
\caption{Illustration of the effect of the values of $\lambda$ and $\beta$ and the number of iterations $T$ on the performance of our SP-DLRR in terms of  AA.
Here the percentage of labeled pixels per class equals to $5\%$, $0.5\%$ and $0.2\%$ for \textit{Indian Pines}, \textit{Salinas Valley}, and \textit{Pavia University}, respectively. (a)-(c) for \textit{Indian Pines}, (d)-(f) for \textit{Salinas Valley}, and (g)-(i) for \textit{Pavia University}.}
\label{fig:parameteranalysisaa}
\end{figure*}
\subsection{Datasets and Experiment Settings}
Three commonly used benchmark datasets were adopted for evaluation\footnote{\scriptsize\url{http://www.ehu.eus/ccwintco/index.php/Hyperspectral_Remote_Sensing_Scenes}}, i.e., \textit{Indian Pines}, \textit{Salinas Valley}, and \textit{Pavia University}.
We describe more characteristics of these three datasets as follows.
\subsubsection{Indian Pines Dataset}
This image scene was collected by Airborne Visible and InfraRed Imaging Spectrometer (AVIRIS) sensor over Indian Pines test site, which consists of $145 \times 145$ pixels and 224 spectral bands with a wavelength range in the 0.4-2.5 um. Since there is no useful information in the water-absorption bands, these bands that cover the region of water absorption are discarded and only 200 bands are remained. There is a total of 10249 pixels for 16 different classes labeled in the ground truth map.
\subsubsection{Salinas Valley Dataset}
This image scene was collected by the AVIRIS sensor over Salinas Valley test site, which consists of $512 \times 217$ pixels and 224 spectral bands with a spatial resolution of 3.7-meter. Only 20 water-absorption bands are discarded, i.e., bands 108-112, 154-167, and 224. In total, 204 bands are remained after bad band removal. There are 54129 labeled pixels for 16 different classes in the ground truth map.
\subsubsection{Pavia University Dataset}
This image scene was collected by the ROSIS sensor over the urban area of the University of Pavia, which consists of $610 \times 340$ pixels and 103 spectral bands with a geometric resolution of 1.3-meter. It contains 42776 labeled pixels for 9 different classes in the ground truth map.

We selected three commonly-used evaluation criteria to evaluate the performance of all methods, i.e., overall accuracy (OA), average accuracy (AA), and Kappa coefficient ($\kappa$). The SVM classifier with the radial basis function kernel was set as the basic classifier, which is implemented by the LIBSVM package\footnote{\scriptsize\url{https://www.csie.ntu.edu.tw/~cjlin/libsvm/}}.
\begin{table*}[t]
  \centering
  \caption{Average accuracy(\%) of ten repeated experiments on \textit{Indian Pines} obtained by different methods with 5$\%$ training samples per class.  The best  and second-best results are highlighted in bold and underlined, respectively.}
  \label{tab:IPsp}
    \begin{threeparttable}
    \begin{tabular}{c|c|c||cccccc||c}
    \hline
    \hline
  Class & {Training} & {Testing} & SGR\cite{xue2017sparse}&CCJSR\cite{tu2018hyperspectral}&LLRSSTV\cite{he2018hyperspectral}& LRMR\cite{zhang2013hyperspectral}& NAILRMA\cite{he2015hyperspectral}& S$^3$LRR\cite{mei2018simultaneous}& Proposed\\
    \hline
    \hline
          {1} & 3     & 43    & 60.23       & \underline{99.31} & 78.83  & 81.16  & 92.33  & 77.91  & \textbf{100.00} \\
          {2} & 72    & 1356  & 83.50       & 88.38  & 80.47  & 84.02  & 87.79  & \textbf{94.62} & \underline{91.35} \\
          {3} & 42    & 788   & 86.00       & 89.59  & 71.05  & 80.53  & 84.07  & \underline{89.64} & \textbf{95.19} \\
          {4} & 12    & 225   & 86.80       & \textbf{87.40} & 55.68  & 64.44  & 74.84  & 85.86  & \underline{87.38} \\
          {5} & 25    & 458   & 90.50       & \underline{94.80} & 84.45  & 91.22  & 88.40  & 91.11  & \textbf{97.97} \\
          {6} & 37    & 693   & \textbf{99.76} & 93.60  & 93.10  & 97.63  & 96.29  & 97.94  & \underline{99.57} \\
          {7} & 2     & 26    & \underline{96.15} & 84.00  & 70.38  & 91.15  & 88.07  & 61.15  & \textbf{98.08} \\
          {8} & 24    & 454   & \textbf{100.00} & \underline{99.93} & 97.00  & 98.19  & 99.36  & 97.29  & 99.76  \\
          {9} & 1     & 19    & 25.26  & \underline{77.36} & 18.42  & 36.84  & 22.10  & 51.05  & \textbf{100.00} \\
          {10} & 49    & 923   & 82.68  & \underline{88.77} & 74.63  & 81.99  & 84.27  & 87.46  & \textbf{94.52} \\
          {11} & 123   & 2332  & \textbf{96.73} & 92.96  & 88.95  & 88.46  & 94.67  & \underline{96.55} & 95.46  \\
          {12} & 30    & 563   & 88.06  & \underline{88.31} & 61.49  & 76.32  & 83.07  & 83.02  & \textbf{92.24} \\
          {13} & 11    & 194   & \textbf{99.74} & 98.47  & 90.05  & 96.34  & 94.74  & 90.88  & \underline{99.38} \\
          {14} & 64    & 1201  & \textbf{99.80} & 97.74  & 98.00  & 98.31  & 99.24  & 96.97  & \underline{99.68} \\
          {15} & 20    & 366   & 89.12  & 91.27  & 81.03  & 78.25  & 81.53  & \textbf{98.39} & \underline{97.02} \\
          {16} & 5     & 88    & \textbf{100.00} & \underline{96.38} & 89.31  & 89.32  & 89.32  & 81.93  & 91.59  \\
    \hline
    \hline
           OA    &   -  &   -   & 91.91  & 92.27 & 83.68  & 87.44  & 90.63  & \underline{93.32} & \textbf{95.67} \\
           AA    &   -   &  -   & 86.52  & \underline{91.77} & 77.06  & 83.39  & 85.01  & 86.36  & \textbf{96.20} \\
          $\kappa$  &  -  &  -   & 90.76  & 91.19 & 81.40  & 85.69  & 89.32  & \underline{92.39} & \textbf{95.07} \\
    \hline
    \hline
    \end{tabular}
    \begin{tablenotes}
    \item[*]{1: Alfalfa 2: Corn-notill 3: Corn-mintill 4: Corn 5: Grass-pasture 6: Grass-trees 7: Grass-pasture-mowed 8: Hay-windrowed
    9: Oats 10: Soybean-notill 11: Soybean-mintill 12: Soybean-clean 13: Wheat 14: Woods 15: Buildings-Grass-Trees-Drives
    16: Stone-Steel-Towers}
     \end{tablenotes}
   \end{threeparttable}
\end{table*}
\begin{table*}[htbp]
 \centering
  \caption{Classification accuracy (\%) and standard deviation (mean$\pm$std) of ten repeated experiments on \textit{Indian Pines} obtained by different methods under different percentages of training pixels per class.  The best and second-best results are highlighted in bold and underlined, respectively.}
  \label{tab:IPmp}
    \begin{tabular}{c|c|c|c|c|c|c|c|c}
     \hline
    \hline
    P & metric & SGR\cite{xue2017sparse}& CCJSR\cite{tu2018hyperspectral}&LLRSSTV\cite{he2018hyperspectral}& LRMR\cite{zhang2013hyperspectral}& NAILRMA\cite{he2015hyperspectral}&S$^3$LRR\cite{mei2018simultaneous}&Proposed \\
    \hline
           1\%   & OA       &74.81$\pm$3.86	& 74.71$\pm$1.95            &63.36$\pm$2.45	&66.93$\pm$2.76	&71.43$\pm$2.67	&\underline{76.24$\pm$1.57}	&\textbf{79.43$\pm$1.82}\\
                  & AA      &67.77$\pm$6.31	& \underline{71.57$\pm$2.10}&58.65$\pm$4.10	&65.13$\pm$3.03	&67.39$\pm$2.98	&70.10$\pm$2.74  		   	&\textbf{79.04$\pm$1.67}\\
                  & $\kappa$&71.11$\pm$4.48	& 71.04$\pm$2.35            &58.24$\pm$2.77	&62.41$\pm$3.07	&67.57$\pm$2.99	&\underline{72.99$\pm$1.78}	&\textbf{76.43$\pm$2.11}\\
            \hline
           3\%   & OA       &86.73$\pm$2.76	& 87.82$\pm$0.98   		    &78.06$\pm$1.78	&81.00$\pm$1.46	&84.74$\pm$2.19	&\underline{90.18$\pm$0.74} & \textbf{92.76$\pm$1.67}\\
                  & AA      &80.94$\pm$7.46	& \underline{86.47$\pm$1.63}&68.86$\pm$3.68	&75.51$\pm$4.24	&77.83$\pm$4.20	&81.26$\pm$2.08 			& \textbf{91.97$\pm$2.86}\\
                  & $\kappa$&84.75$\pm$3.23	& 86.09$\pm$1.12 		    &74.95$\pm$2.01	&78.37$\pm$1.64	&82.65$\pm$2.46	&\underline{88.81$\pm$0.83} & \textbf{91.73$\pm$1.90}\\
            \hline
           5\%   & OA    	&91.91$\pm$2.15 & 92.27$\pm$0.71 			&83.68$\pm$0.98 &87.44$\pm$0.63 &90.63$\pm$0.83 &\underline{93.32$\pm$0.66} & \textbf{95.67$\pm$1.14}\\
                  & AA    	&86.52$\pm$3.33 & \underline{91.77$\pm$2.06}&77.06$\pm$1.91 &83.39$\pm$2.01 &85.01$\pm$1.66 &86.36$\pm$1.90 			& \textbf{96.20$\pm$0.95}\\
                  & $\kappa$&90.76$\pm$2.46 & 91.19$\pm$0.80 			&81.40$\pm$1.11 &85.69$\pm$0.72 &89.32$\pm$0.94 &\underline{92.39$\pm$0.75} & \textbf{95.07$\pm$1.30}\\
            \hline
           7\%    & OA      &93.05$\pm$2.53 & 94.49$\pm$0.54 			&86.57$\pm$0.75 &89.91$\pm$0.87 &91.93$\pm$0.71 &\underline{95.55$\pm$1.04} & \textbf{96.64$\pm$0.45}\\
                  & AA      &88.79$\pm$5.71 & \underline{94.47$\pm$0.64}&81.64$\pm$2.11 &86.98$\pm$2.12 &87.84$\pm$1.14 &90.55$\pm$1.85 		    & \textbf{96.73$\pm$0.80}\\
                  & $\kappa$&92.04$\pm$2.92 & 93.72$\pm$0.61 			&84.69$\pm$0.86 &88.50$\pm$0.99 &90.80$\pm$0.81 &\underline{94.93$\pm$1.18} & \textbf{96.17$\pm$0.51}\\
		    \hline
            \hline
      \end{tabular}
  \label{tab:addlabel}
\end{table*}
\subsection{Parameter Setting and Convergence Analysis}
Following the suggestion for the number of patches in \cite{mei2018simultaneous}, we empirically set the initial numbers of superpixels to $64$, $50$, and $50$ for \textit{Indian Pines}, \textit{Salinas Valley}, and \textit{Pavia University}, respectively.
For the threshold $\delta$ controlling whether to partition a superpixel to more sub-superpixels, we set it to $0.7$, $0.6$ and $0.2$ for \textit{Indian Pines}, \textit{Salinas Valley}, and \textit{Pavia University},  respectively. For the parameter $M$ controlling the number of sub-superpixels, we set it to $5$, $3$ and $3$ for \textit{Indian Pines}, \textit{Salinas Valley}, and \textit{Pavia University}, respectively.

\begin{figure}[H]
\centering
\subfigure[$\beta=1$]{
\begin{minipage}[t]{0.33\linewidth}
\centering
\includegraphics[width=1.0\textwidth , height=0.07\textheight]{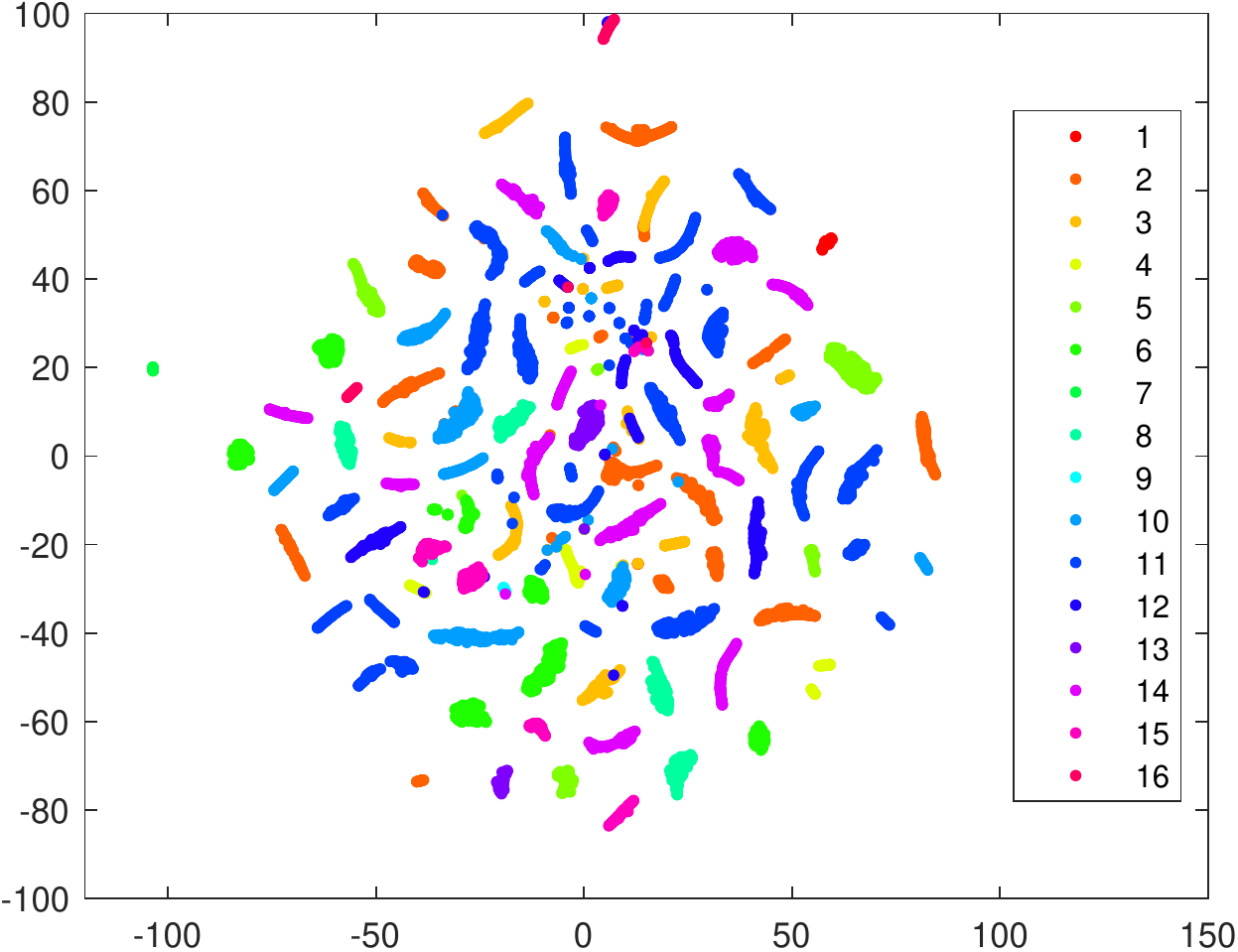}
\end{minipage}}
\subfigure[$\beta=0$]{
\begin{minipage}[t]{0.33\linewidth}
\centering
\includegraphics[width=1.0\textwidth , height=0.07\textheight]{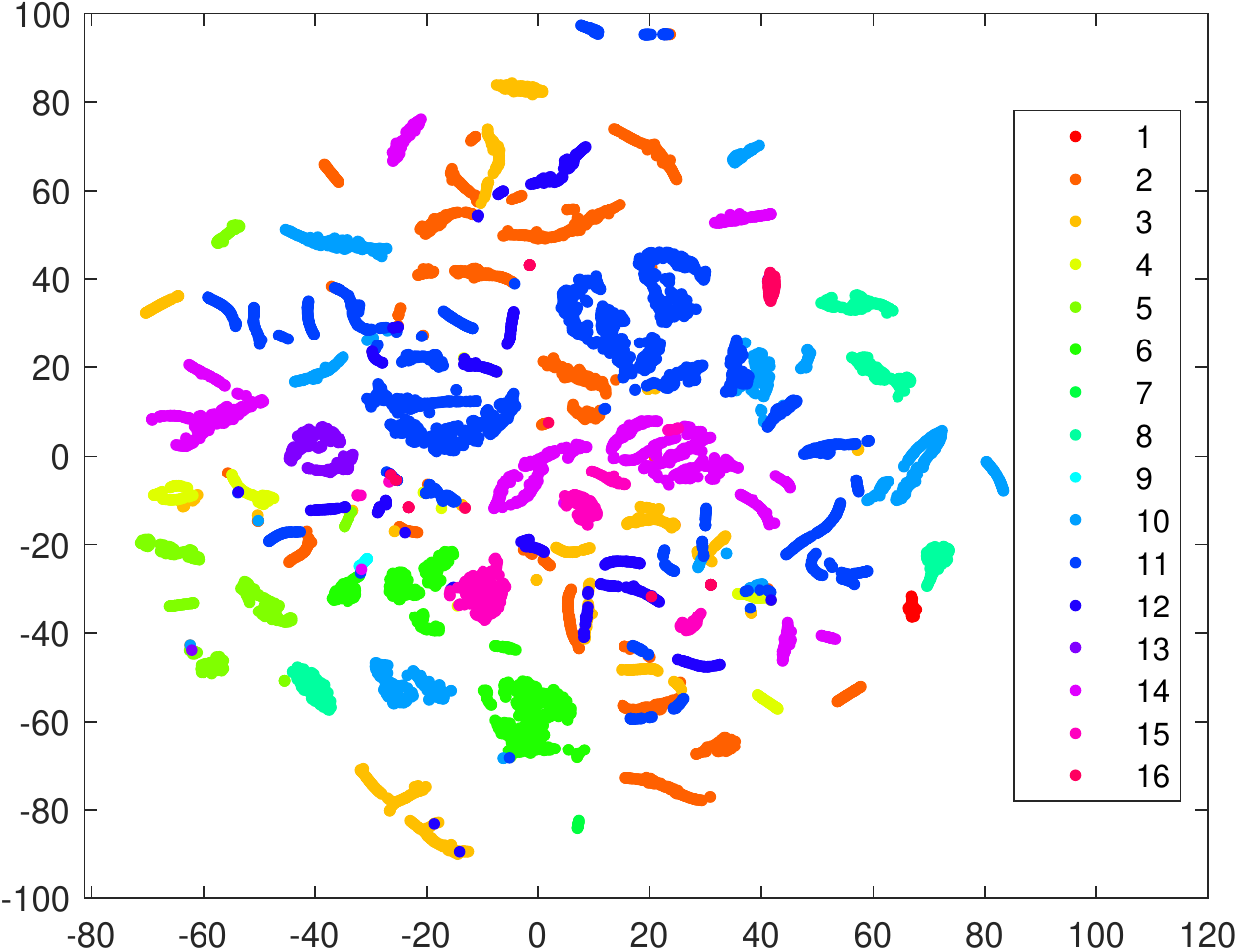}
\end{minipage}}

\subfigure[$\beta=1$]{
\begin{minipage}[t]{0.33\linewidth}
\centering
\includegraphics[width=1.0\textwidth , height=0.07\textheight]{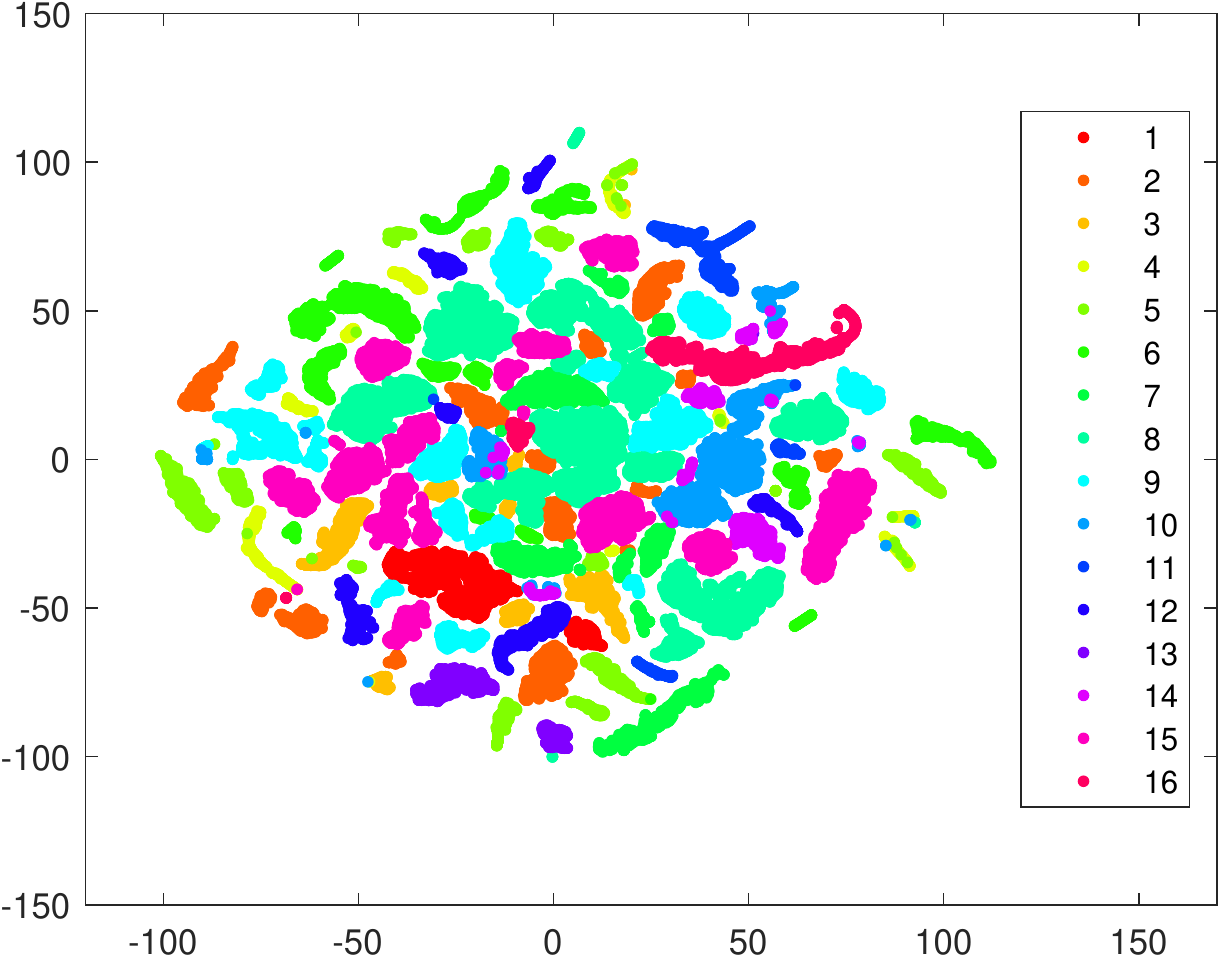}
\end{minipage}}
\subfigure[$\beta=0$]{
\begin{minipage}[t]{0.33\linewidth}
\centering
\includegraphics[width=1.0\textwidth , height=0.07\textheight]{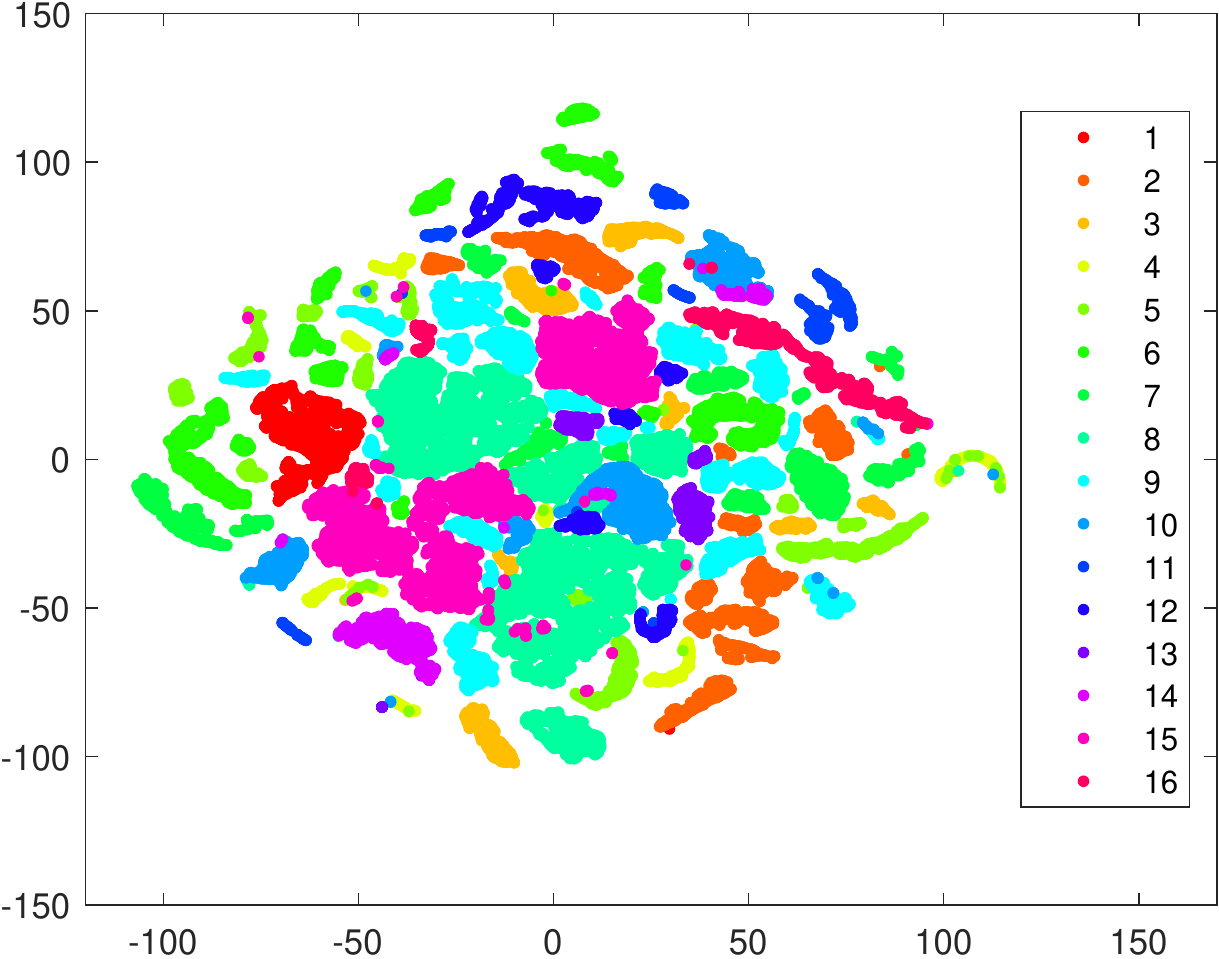}
\end{minipage}}

\subfigure[$\beta=1$]{
\begin{minipage}[t]{0.33\linewidth}
\centering
\includegraphics[width=1.0\textwidth , height=0.07\textheight]{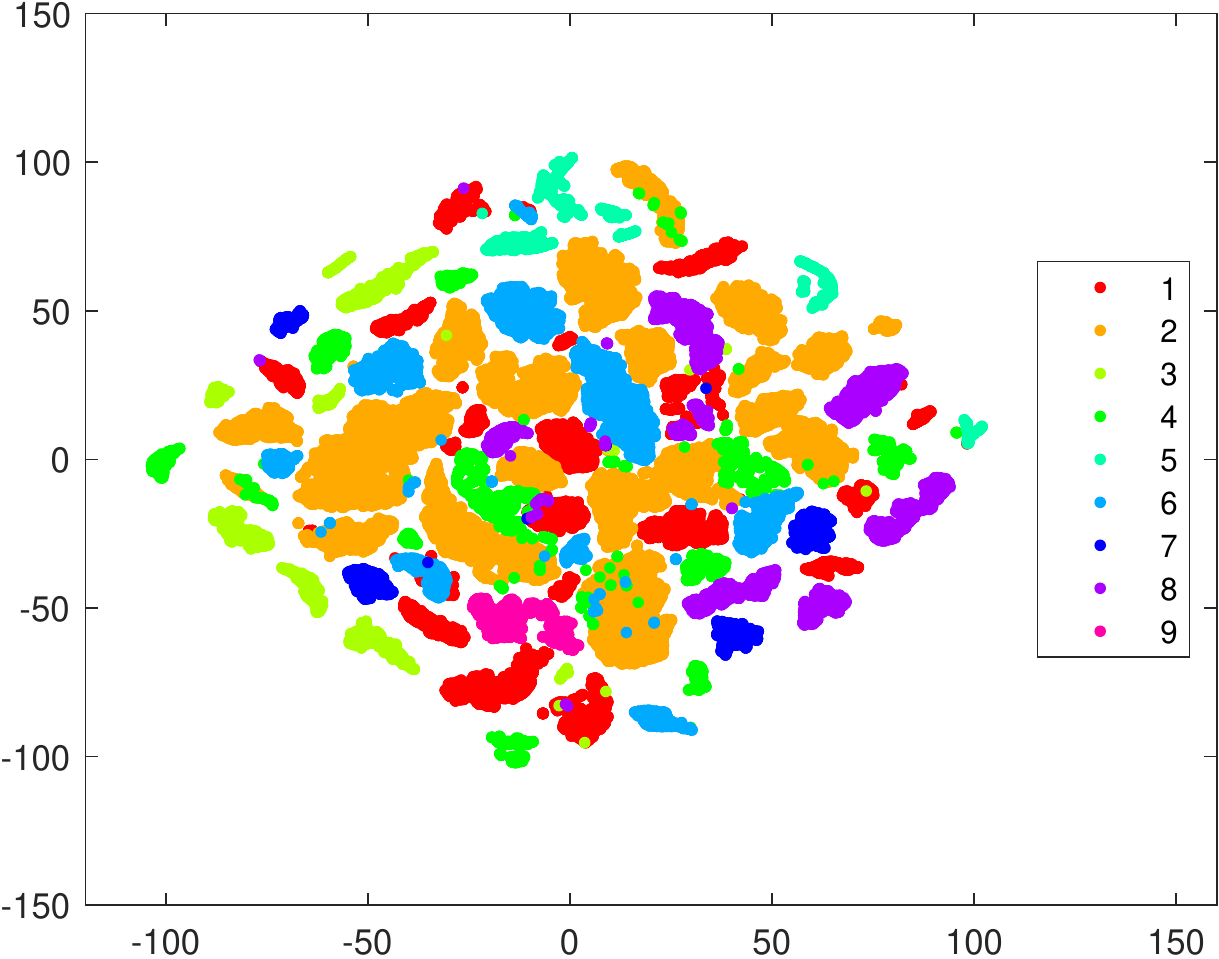}
\end{minipage}}
\subfigure[$\beta=0$]{
\begin{minipage}[t]{0.33\linewidth}
\centering
\includegraphics[width=1.0\textwidth , height=0.07\textheight]{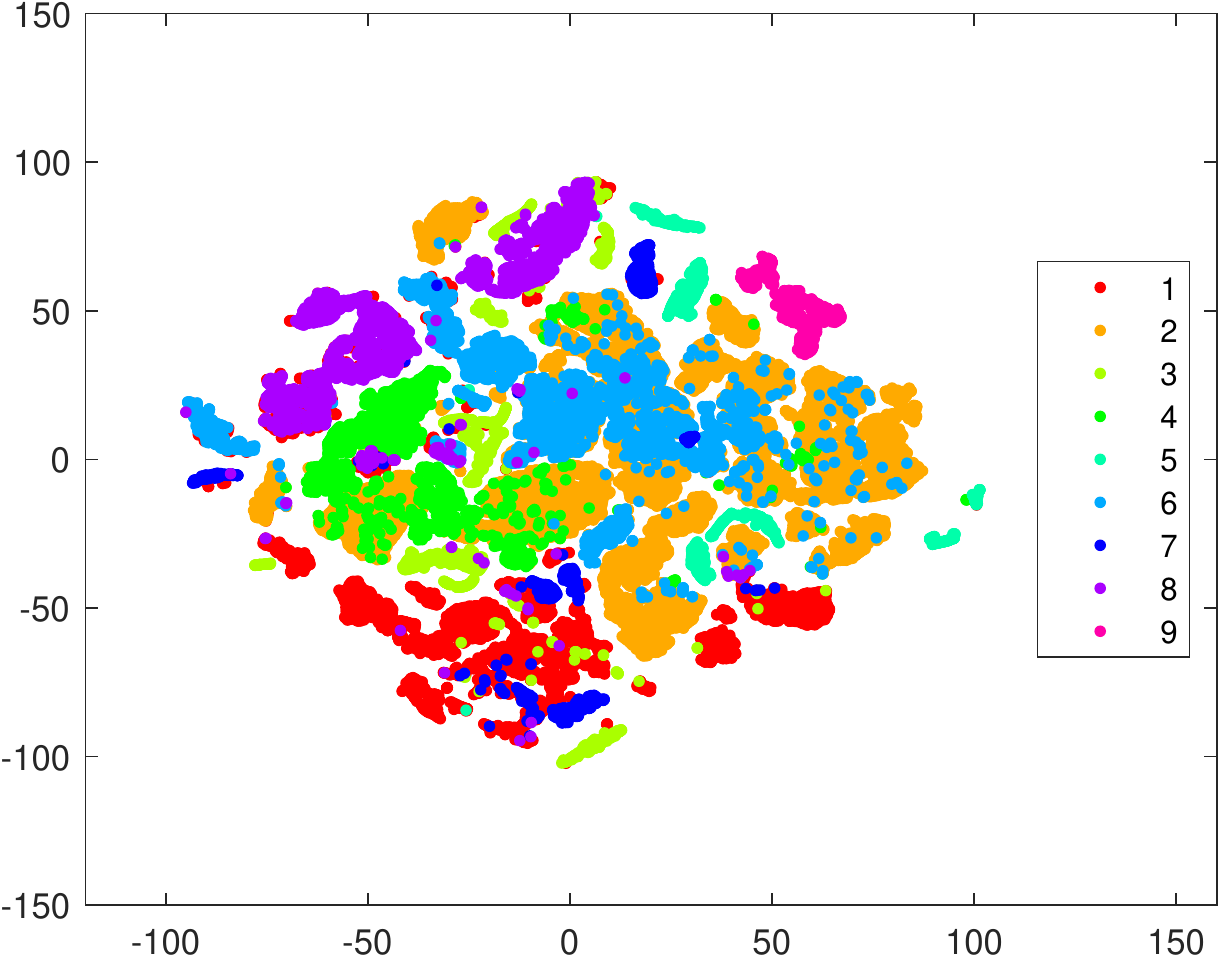}
\end{minipage}}
\caption{
Visual illustration of the effectiveness of the negative low-rank regularization term in Eq. (\ref{eq:global}), i.e., $-\beta\|\mathbf{L}\|_*$, by showing the t-SNE maps of the results by SP-DLRR with $\beta=1$ and $\beta=0$. when the percentage of labeled pixels per class equals to $5\%$, $0.5\%$ and $0.5\%$ for \textit{Indian Pines}, \textit{Salinas Valley}, and {\textit{Pavia University}}, respectively. (a)-(b) for \textit{Indian Pines}, (c)-(d) for \textit{Salinas Valley}, and (e)-(f) for {\textit{Pavia University}}.}
\label{fig:t-SNE}
\end{figure}

In the following, we investigated how the parameters $\lambda$ and $\beta$ and the number of iterations $T$ affect the performance of SP-DLRR. Note that $\lambda$ depends on the noise level of the input HSI, and $\beta$ is used to balance the intra-class similarity and inter-class dissimilarity.
The values of $\lambda$ and $\beta$ vary in the range of
$\{0.01, 0.05, 0.1\}$ and
$\{0, 0.1, 1, 10\}$
respectively. Figs. \ref{fig:parameteranalysisoa} and \ref{fig:parameteranalysisaa} illustrate the values of OA and AA for SP-DLRR with different $\lambda$ and $\beta$ on the three benchmark datasets, respectively, from which We can draw the following conclusions:
\begin{enumerate}
    \item
when $\lambda=0.05$, the OA achieves much better performance than other settings for \textit{Indian Pines}.  The highest OA is achieved for \textit{Salinas Valley} and \textit{Pavia University} when $\lambda=0.01$.  Besides, $\beta=1$ always achieves the best performance under different $\lambda$ for all the three datasets. Similar observations of AA can be obtained from Fig. \ref{fig:parameteranalysisaa}.
\item
with the optimal settings of $\lambda$ and $\beta$,
 the accuracy has a significant improvement at the beginning of the iteration, and when $T$ increases to 3, the best performance is achieved  for \textit{Indian Pines} and \textit{Pavia University}, and almost the best performance for \textit{Salinas Valley}.

\item
with the same values of $\lambda$, $\beta$ and $T$, the performance of SP-DLRR with $\beta=1$ is always better than that of $\beta=0$ overall all the three datasets, which demonstrates the effectiveness of the negative low-rank regularization term in Eq. (\ref{eq:global}) (i.e., $-\beta\|\mathbf{L}\|_*$) on promoting the inter-class discriminability. Moreover, to show the effectiveness of such a term more intuitively, we employed the t-SNE method to visualize the generated $\mathbf{L}$ by SP-DLLR with $\beta=0$ and $\beta=1$, respectively. 
As shown in the Fig. \ref{fig:t-SNE}, in the t-SNE maps of the results by SP-DLRR with $\beta=1$, the pixels belonging an identical class are more compact, and the gaps between different classes are larger, compared with those by SP-DLRR with $\beta=0$.
\end{enumerate}

Therefore, in the following experiments, we set $\lambda$ as 0.05, 0.01 and 0.01 for \textit{Indian Pines}, \textit{Salinas Valley}, and \textit{Pavia University}, respectively, and $\beta$  and $T_{max}$ as 1  and 3 for all the datasets.   

\begin{figure}[t]
\subfigure[]{
\begin{minipage}[t]{0.3\linewidth}
\centering
\includegraphics[width=1.0\textwidth , height=0.09\textheight]{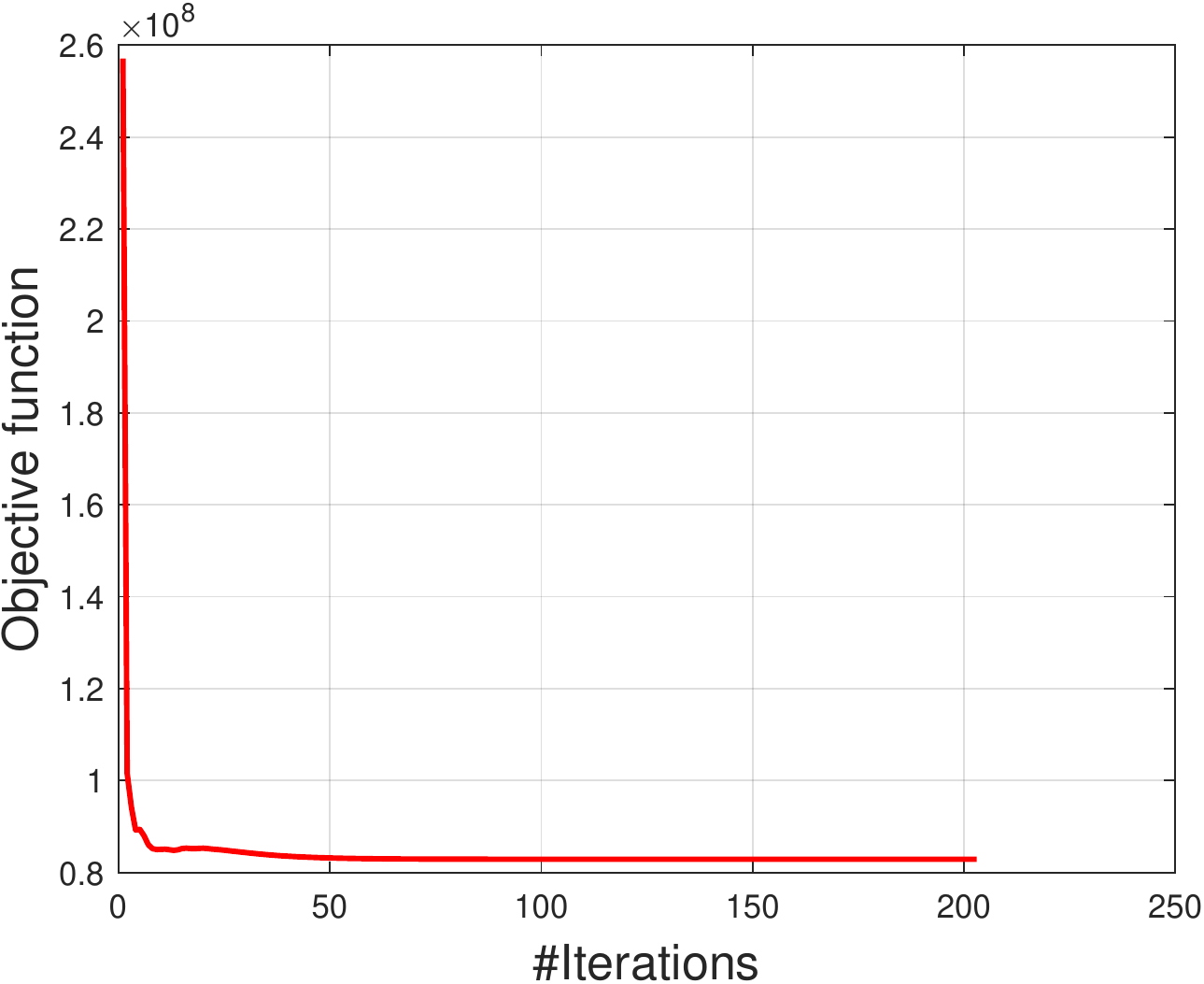}
\end{minipage}}
\centering
\subfigure[]{
\begin{minipage}[t]{0.3\linewidth}
\centering
\includegraphics[width=1.0\textwidth , height=0.09\textheight]{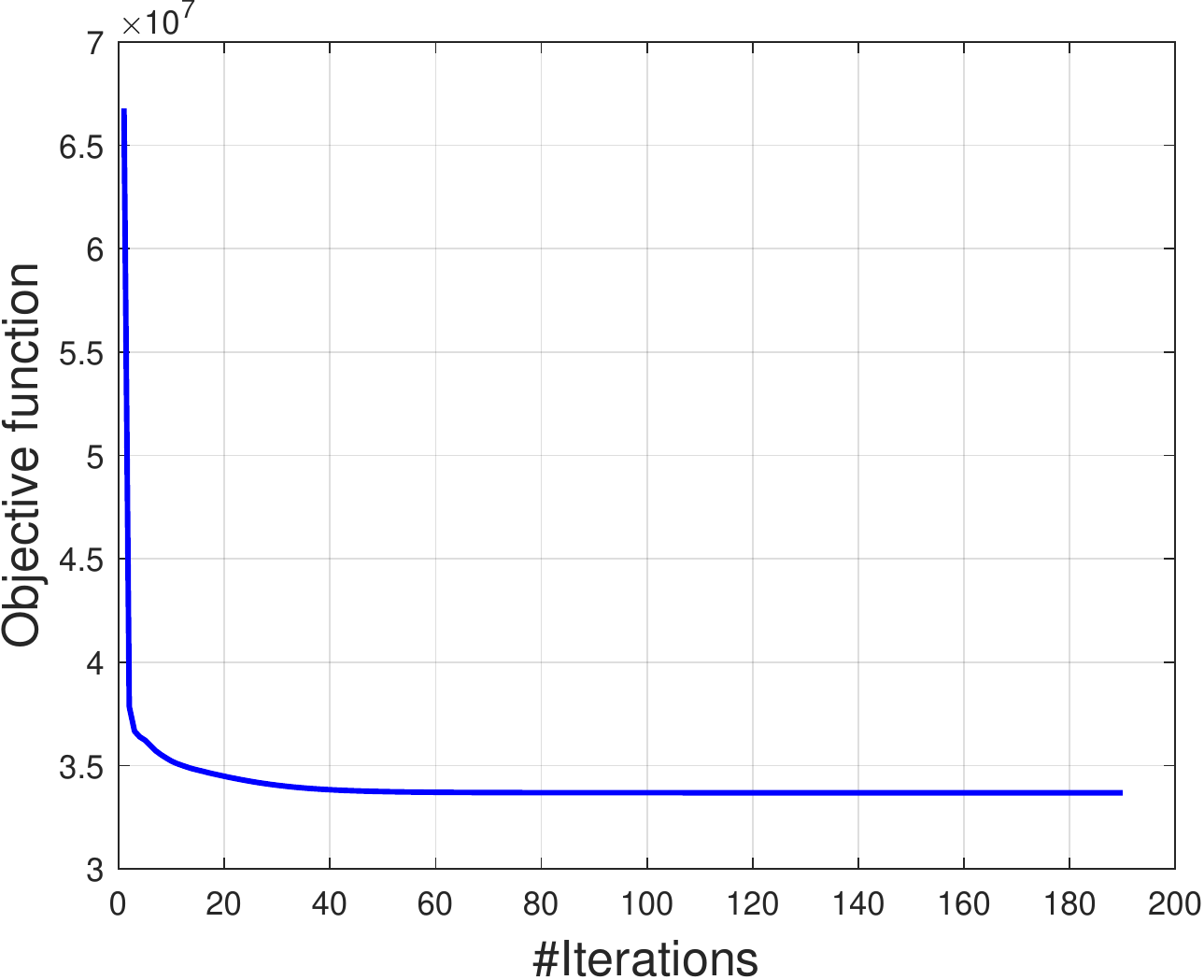}
\end{minipage}}
\subfigure[]{
\begin{minipage}[t]{0.3\linewidth}
\centering
\includegraphics[width=1.0\textwidth , height=0.09\textheight]{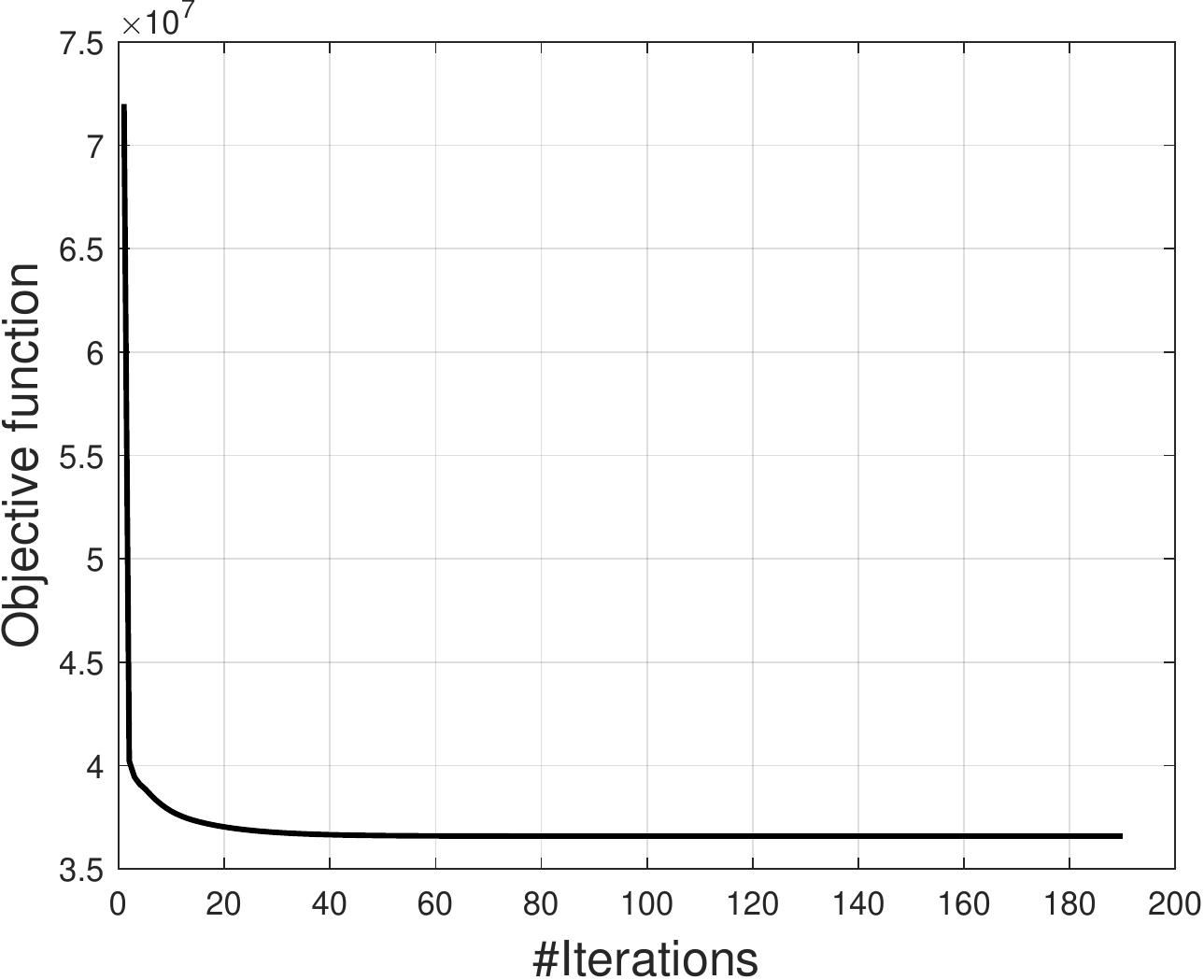}
\end{minipage}}
\caption{Empirical convergence verification of Algorithm \ref{algorithm:lrgr} (a) \textit{Indian Pines}, (b) \textit{Salinas Valley} and (c) \textit{Pavia University}. }
\label{fig:convergence}
\end{figure}

As aforementioned, although the global convergence of Algorithm \ref{algorithm:lrgr} is not theoretically guaranteed, as shown in Fig. \ref{fig:convergence}, we experimentally find that it converges well on all the three datasets, 
i.e., the objective function in Eq. (\ref{eq:obj_fun}) converges to a stable value after about 50 iterations.

\begin{table*}[htbp]
  \centering
    \caption{Average accuracy(\%) of ten repeated experiments on \textit{Salinas Valley} obtained by different methods with 0.5$\%$ training samples per class.  The best  and second-best results are highlighted in bold and underlined, respectively.}
    \label{tab:SVsp}
    \begin{threeparttable}
    \begin{tabular}{c|c|c||cccccc||c}
    \hline
    \hline
  Class & {Training} & {Testing} & SGR\cite{xue2017sparse}&CCJSR\cite{tu2018hyperspectral}&LLRSSTV\cite{he2018hyperspectral}& LRMR\cite{zhang2013hyperspectral}& NAILRMA\cite{he2015hyperspectral}& S$^3$LRR\cite{mei2018simultaneous}& Proposed\\
    \hline
    \hline
    {1} & 11    & 1998  & \textbf{100.00} & \underline{99.95} & 96.40  & 96.76  & 97.45  & 93.59  & \textbf{100.00} \\
    {2} & 19    & 3707  & \textbf{100.00} & 99.23  & 98.87  & 98.44  & 98.55  & 95.45  & \underline{99.37} \\
    {3} & 10    & 1966  & 89.93  & \underline{95.28} & 90.34  & 89.95  & 89.59  & 90.33  & \textbf{99.20} \\
    {4} & 7     & 1387  & \underline{98.39} & \textbf{99.66} & 96.65  & 98.00  & 96.37  & 84.62  & 96.61  \\
    {5} & 14    & 2664  & \textbf{99.24} & 91.59  & 92.08  & 93.90  & 94.29  & 87.81  & \underline{98.80} \\
    {6} & 20    & 3939  & \textbf{99.98} & 97.03  & 99.53  & 99.17  & 99.24  & 98.47  & \underline{99.80} \\
    {7} & 18    & 3561  & \textbf{99.96} & 99.17  & 97.93  & 99.31  & 99.35  & 95.89  & \underline{99.60} \\
    {8} & 57    & 11214 & 93.01  & 79.04  & \underline{96.33} & 87.04  & 87.26  & 92.60  & \textbf{99.06} \\
    {9} & 32    & 6171  & \textbf{100.00} & 98.08  & 98.56  & 98.18  & 98.05  & \underline{98.65} & \textbf{100.00 } \\
    {10} & 17    & 3261  & \textbf{97.30} & 95.46  & 81.12  & 85.87  & 87.38  & 89.26  & \underline{95.92} \\
    {11} & 6     & 1062  & 96.06  & \underline{96.13} & 87.64  & 87.28  & 89.73  & 77.39  & \textbf{99.04} \\
    {12} & 10    & 1917  & 70.27  & 95.89  & 97.98  & \underline{99.23} & 99.02  & 81.71  & \textbf{100.00} \\
    {13} & 5     & 911   & \textbf{99.63} & 90.61  & 90.93  & 97.56  & 97.32  & 77.30  & \underline{98.28} \\
    {14} & 6     & 1064  & 86.53  & \textbf{96.46} & 87.61  & 90.12  & 90.14  & 87.81  & \underline{92.27} \\
    {15} & 37    & 7231  & \underline{97.46} & 67.32  & 96.17  & 71.84  & 86.94  & 93.55  & \textbf{99.52} \\
    {16} & 10    & 1797  & \underline{98.64} & \textbf{99.33} & 92.22  & 91.56  & 92.70  & 88.43  & 98.45  \\
    \hline
    \hline
    OA    &  -   &   -   & \underline{96.14} & 89.23  & 95.22  & 90.61  & 92.83  & 92.39  & \textbf{98.98} \\
    AA    &  -   &   -   & \underline{95.40} & 93.77  & 93.77  & 92.76  & 93.96  & 89.55  & \textbf{98.49} \\
    $\kappa$      &  -   &   -   & \underline{95.70} & 88.00  & 94.69  & 89.53  & 92.03  & 91.53  & \textbf{98.86} \\
    \hline
    \hline
    \end{tabular}%
    \begin{tablenotes}
    \item[*]{1: Brocoli\underline{ }green\underline{ }weeds\underline{ }1 2: Brocoli\underline{ }green\underline{ }weeds\underline{ }2 3: Fallow 4: Fallow\underline{ }rough\underline{ }plow 5: Fallow\underline{ }smooth 6: Stubble 7: Celery 8: Grapes\underline{ }untrained 9: Soil\underline{ }vinyard\underline{ }develop
10: Corn\underline{ }senesced\underline{ }green\underline{ }weeds 11: Lettuce\underline{ }romaine\underline{ }4wk 12: Lettuce\underline{ }romaine\underline{ }5wk 13: Lettuce\underline{ }romaine\underline{ }6wk 14: Lettuce\underline{ }romaine\underline{ }7wk 15: Vinyard\underline{ }untrained
16: Vinyard\underline{ }vertical\underline{ }trellis}
     \end{tablenotes}
  \end{threeparttable}
\end{table*}%

\begin{table*}[htbp]
 \centering
  \caption{Classification accuracy (\%) and standard deviation (mean$\pm$std) of ten repeated experiments on \textit{Salinas Valley} obtained by different methods under different percentages of training pixels per class.  The best  and second-best results are highlighted in bold and underlined, respectively.}
  \label{tab:SVmp}
    \begin{tabular}{c|c|c|c|c|c|c|c|c}
     \hline
    \hline
    P & metric & SGR\cite{xue2017sparse}& CCJSR\cite{tu2018hyperspectral}& LLRSSTV\cite{he2018hyperspectral}& LRMR\cite{zhang2013hyperspectral}& NAILRMA\cite{he2015hyperspectral}& S$^3$LRR\cite{mei2018simultaneous}&Proposed \\
    \hline
    0.1\% & OA    & \underline{93.33$\pm$2.12} & 80.72$\pm$2.50 & 87.67$\pm$0.94 & 80.62$\pm$2.54 & 82.77$\pm$1.92 & 79.47$\pm$3.30 & \textbf{95.85$\pm$1.52}\\
          & AA    & \underline{92.80$\pm$2.70} & 82.84$\pm$4.87 & 84.31$\pm$2.43 & 82.55$\pm$2.94 & 83.55$\pm$2.80 & 74.25$\pm$3.33 & \textbf{95.22$\pm$1.39}\\
          & $\kappa$& \underline{92.58$\pm$2.35} & 78.50$\pm$2.78 & 86.25$\pm$1.03 & 78.38$\pm$2.79 & 80.78$\pm$2.13 & 77.16$\pm$3.64 & \textbf{95.37$\pm$1.70}\\
    \hline
    0.3\% & OA    & \underline{95.77$\pm$1.92} & 86.50$\pm$1.16 & 93.47$\pm$1.26 & 88.04$\pm$1.44 & 90.57$\pm$1.62 & 89.74$\pm$1.16 & \textbf{98.43$\pm$0.61}\\
          & AA    & \underline{95.49$\pm$1.30} & 91.94$\pm$0.90 & 90.51$\pm$2.51 & 90.49$\pm$1.34 & 91.76$\pm$1.51 & 86.68$\pm$1.19 & \textbf{97.92$\pm$0.56}\\
          & $\kappa$& \underline{95.29$\pm$2.13} & 84.96$\pm$1.30 & 92.72$\pm$1.41 & 86.67$\pm$1.60 & 89.50$\pm$1.81 & 88.58$\pm$1.28 & \textbf{98.25$\pm$0.68}\\
    \hline
    0.5\% & OA    & \underline{96.14$\pm$1.01} & 89.23$\pm$1.16 & 95.22$\pm$0.88 & 90.61$\pm$0.58 & 92.83$\pm$0.60 & 92.39$\pm$0.81 & \textbf{98.98$\pm$0.17}\\
          & AA    & \underline{95.40$\pm$1.93} & 93.77$\pm$0.50 & 93.77$\pm$0.83 & 92.76$\pm$0.45 & 93.96$\pm$0.49 & 89.55$\pm$1.46 & \textbf{98.49$\pm$0.32}\\
          & $\kappa$& \underline{95.70$\pm$1.12} & 88.00$\pm$1.30 & 94.69$\pm$0.98 & 89.53$\pm$0.65 & 92.03$\pm$0.67 & 91.53$\pm$0.91 & \textbf{98.86$\pm$0.18}\\
    \hline
    0.7\% & OA    & 95.76$\pm$1.50 & 90.45$\pm$0.50 & \underline{96.04$\pm$0.88} & 91.22$\pm$0.64 & 93.52$\pm$0.88 & 94.52$\pm$0.83 & \textbf{99.15$\pm$0.18}\\
          & AA    & \underline{95.12$\pm$1.30} & 94.46$\pm$0.31 & 94.53$\pm$1.42 & 93.58$\pm$0.79 & 94.70$\pm$1.03 & 92.49$\pm$0.92 & \textbf{98.78$\pm$0.46}\\
          & $\kappa$& 95.29$\pm$1.67 & 89.37$\pm$0.56 & \underline{95.59$\pm$0.98} & 90.21$\pm$0.71 & 92.79$\pm$0.97 & 93.89$\pm$0.93 & \textbf{99.05$\pm$0.20}\\
 		    \hline
            \hline
      \end{tabular}%
  \label{tab:addlabel}%
\end{table*}%

\begin{table*}[t]
  \centering
  \caption{Average accuracy(\%) of ten repeated experiments on \textit{Pavia University} obtained by different methods with 0.2$\%$ training samples per class.  The best  and second-best results are highlighted in bold and underlined, respectively.}
  \label{tab:PUsp}
    \begin{threeparttable}
    \begin{tabular}{c|c|c||cccccc||c}
    \hline
    \hline
  Class & {Training} & {Testing} & SGR\cite{xue2017sparse}&CCJSR\cite{tu2018hyperspectral}&LLRSSTV\cite{he2018hyperspectral}& LRMR\cite{zhang2013hyperspectral}& NAILRMA\cite{he2015hyperspectral}& S$^3$LRR\cite{mei2018simultaneous}& Proposed\\
    \hline
    \hline
    {1} & 14    & 6617  & 86.23 & \underline{88.35} & 76.11 & 75.50 & 74.77 & 79.53 & \textbf{92.02} \\
    {2} & 38    & 18611 & \underline{97.40} & 82.29 & 95.85 & 93.16 & 92.60 & 91.26 & \textbf{99.19} \\
    {3} & 5     & 2094  & \textbf{87.08} & 49.50 & 70.43 & 54.34 & 58.21 & 60.08 & \underline{82.77} \\
    {4} & 7     & 3057  & 76.43 & \textbf{87.76} & 70.74 & 76.17 & \underline{81.64} & 70.13 & 78.94 \\
    {5} & 3     & 1342  & \textbf{100.00} & 92.54 & 88.26 & 89.03 & 89.11 & 91.21 & \underline{93.02} \\
    {6} & 11    & 5018  & 61.52 & 59.81 & \underline{84.50} & 65.54 & 73.32 & 71.49 & \textbf{90.37} \\
    {7} & 3     & 1327  & \underline{84.74} & 63.25 & 76.32 & 64.85 & 73.84 & 53.72 & \textbf{93.70} \\
    {8} & 8     & 3674  & 76.78 & 46.52 & \underline{80.81} & 73.88 & 73.08 & 66.34 & \textbf{93.66} \\
    {9} & 2     & 945   & 75.53 & 98.47 & 99.58 & \textbf{99.75} & \underline{99.74} & 99.46 & 99.71 \\
    \hline
    \hline
    OA    &   -   &   -   & \underline{86.87} & 73.30 & 86.35 & 81.53 & 82.88 & 80.94 & \textbf{93.96} \\
    AA    &   -   &   -   & \underline{82.86} & 74.28 & 82.51 & 76.91 & 79.59 & 75.91 & \textbf{91.49} \\
    $\kappa$    &   -  &   -  & \underline{82.35} & 64.05 & 81.85 & 75.34 & 77.33 & 74.60 & \textbf{91.92} \\
    \hline
    \hline
    \end{tabular}%
    \begin{tablenotes}
    \item[*]{1: Asphalt	2: Meadows 3: Gravel 4: Trees 5: Painted metal sheets 6: Bare Soil 7: Bitumen 8: Self-Blocking Bricks 9: Shadows}	

     \end{tablenotes}
  \label{tab:addlabel}%
  \end{threeparttable}
\end{table*}

\begin{table*}[htbp]
 \centering
  \caption{Classification accuracy (\%) and standard deviation (mean$\pm$std) of ten repeated experiments on \textit{Pavia University} obtained by different methods under different percentages of training pixels per class. The best  and second-best results are highlighted in bold and underlined, respectively.}
  \label{tab:PUmp}
    \begin{tabular}{c|c|c|c|c|c|c|c|c}
     \hline
    \hline
    P & metric & SGR\cite{xue2017sparse}& CCJSR\cite{tu2018hyperspectral}& LLRSSTV\cite{he2018hyperspectral}& LRMR\cite{zhang2013hyperspectral}& NAILRMA\cite{he2015hyperspectral}& S$^3$LRR\cite{mei2018simultaneous}&Proposed \\
    \hline
   0.1\% & OA       & \underline{83.95$\pm$5.36} & 68.90$\pm$2.91 & 76.78$\pm$5.68 & 73.63$\pm$4.02 & 75.58$\pm$5.12 & 72.28$\pm$3.89 & \textbf{87.87$\pm$3.76} \\
         & AA       & \underline{80.86$\pm$5.97} & 70.24$\pm$2.71 & 72.55$\pm$5.82 & 71.45$\pm$4.17 & 73.19$\pm$4.68 & 68.85$\pm$4.63 & \textbf{84.66$\pm$4.39} \\
         &$\kappa$  & \underline{78.77$\pm$7.10} & 58.51$\pm$3.59 & 69.35$\pm$7.45 & 65.17$\pm$4.99 & 67.79$\pm$6.42 & 63.38$\pm$5.08 & \textbf{83.85$\pm$4.96} \\
    \hline
   0.2\% & OA       & \underline{86.87$\pm$2.97} & 73.30$\pm$2.16 & 86.35$\pm$2.83 & 81.53$\pm$1.58 & 82.88$\pm$2.57 & 80.94$\pm$1.61 & \textbf{93.96$\pm$1.34} \\
         & AA       & \underline{82.86$\pm$4.10} & 74.28$\pm$2.66 & 82.51$\pm$4.18 & 76.91$\pm$2.61 & 79.59$\pm$2.23 & 75.91$\pm$2.76 & \textbf{91.49$\pm$2.15} \\
         &$\kappa$  & \underline{82.35$\pm$3.98} & 64.05$\pm$2.75 & 81.85$\pm$3.76 & 75.34$\pm$2.06 & 77.33$\pm$3.35 & 74.60$\pm$2.18 & \textbf{91.92$\pm$1.79} \\
    \hline
   0.3\% & OA       & {89.27$\pm$2.01} & 74.43$\pm$1.62 & \underline{89.79$\pm$1.76} & 84.86$\pm$1.41 & 87.06$\pm$1.67 & 83.50$\pm$1.59 & \textbf{95.80$\pm$0.98} \\
         & AA       & {85.31$\pm$2.51} & 75.53$\pm$1.92 & \underline{86.80$\pm$2.30} & 81.80$\pm$1.59 & 84.12$\pm$1.88 & 78.82$\pm$3.05 & \textbf{94.56$\pm$1.78} \\
         & $\kappa$ & {85.69$\pm$2.80} & 65.50$\pm$2.21 & \underline{86.43$\pm$2.34} & 79.86$\pm$1.74 & 82.81$\pm$2.21 & 78.02$\pm$2.07 & \textbf{94.41$\pm$1.32} \\
   \hline
   0.5\% & OA       & 91.18$\pm$2.01 & 77.20$\pm$2.02 & \underline{93.06$\pm$1.44} & 88.17$\pm$1.48 & {91.60$\pm$1.35} & 87.05$\pm$0.88 & \textbf{96.90$\pm$0.94} \\
         & AA       & 88.00$\pm$2.01 & 78.19$\pm$2.30 & \underline{91.16$\pm$2.08} & 84.85$\pm$2.29 & {89.80$\pm$2.07} & 83.14$\pm$1.87 & \textbf{96.11$\pm$1.74} \\
         & $\kappa$ & 88.31$\pm$2.69 & 69.21$\pm$2.63 & \underline{90.79$\pm$1.92} & 84.21$\pm$2.01 & {88.86$\pm$1.79} & 82.75$\pm$1.14 & \textbf{95.88$\pm$1.27} \\
          \hline
          \hline
    \end{tabular}
  \label{tab:addlabel}
\end{table*}
\subsection{Comparison with State-of-the-Art Methods}
We demonstrated the advantage of SP-DLRR by comparing with state-of-the-art HSI classification methods, including 1) a semi-supervised
method, i.e., sparse graph regularization method (SGR) \cite{xue2017sparse}, 2) a supervised method, i.e. combining correlation coefficient and joint sparse representation (CCJSR) \cite{tu2018hyperspectral}, and  3) four low-rank prior-based methods for HSIs, i.e., spatial-spectral total variation regularized local low-rank matrix recovery (LLRSSTV) \cite{he2018hyperspectral}, low-rank matrix recovery (LRMR) \cite{zhang2013hyperspectral}, noise-adjusted iterative low-rank matrix approximation (NAILRMA) \cite{he2015hyperspectral}, and simultaneous spatial and spectral low-rank
representation (S$^3$LRR) \cite{mei2018simultaneous}.
We set all the compared methods with the recommended parameters in their papers. For each method, the experiments were repeated 10 times with randomly selected training samples each time, and the average performance was reported.

Table \ref{tab:IPsp} lists the classification accuracy of each class by different methods on \textit{Indian Pines} when 5 $\%$ labeled samples of each class were used for training and the remaining ones for testing. It could be seen that SP-DLRR achieves the highest values of OA, AA and $\kappa$ and outperforms the compared methods to a large extent. Specifically, our method achieves the highest classification accuracy over 7 classes out of 16 classes, and the second best performance over 6 classes, which are very close to the best one. SP-DLRR shows remarkable superiority over the other four low-rank-based methods on the classes with only a few samples, i.e., classes 1, 7, and 9.
The reason is that our method not only captures the local spatial structure but also promotes the inter-class discriminability such that these classes are not absorbed during the process.
In addition, the effectiveness of our method was further validated under the experiments with different percentages of training pixels. For \textit{Indian Pines}, the percentage of the randomly selected training pixels on each class was set as $P=1\%$, $3\%$, $5\%$, and $7\%$, while the rest pixels were used for testing.
 Table \ref{tab:IPmp} shows the classification results. It can be observed that SP-DLRR consistently outperform the other methods. Particularly, when the training pixels on each class are extremely few, e.g., $P=1\%$ and $3\%$, the superiority of our method is more obvious, which is credited to the comprehensive and fine modeling of HSIs.

 \begin{figure*}[htbp]
\centering
\subfigure[]{
\begin{minipage}[t]{\length\linewidth}
\centering
\includegraphics[width=1.0\textwidth]{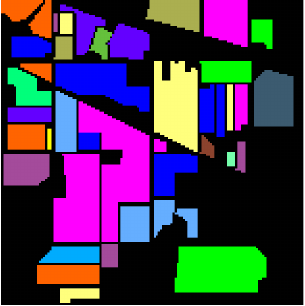}
\label{Fig:7}
\end{minipage}
}
\centering
\subfigure[]{
\begin{minipage}[t]{\length\linewidth}
\centering
\includegraphics[width=1.0\textwidth]{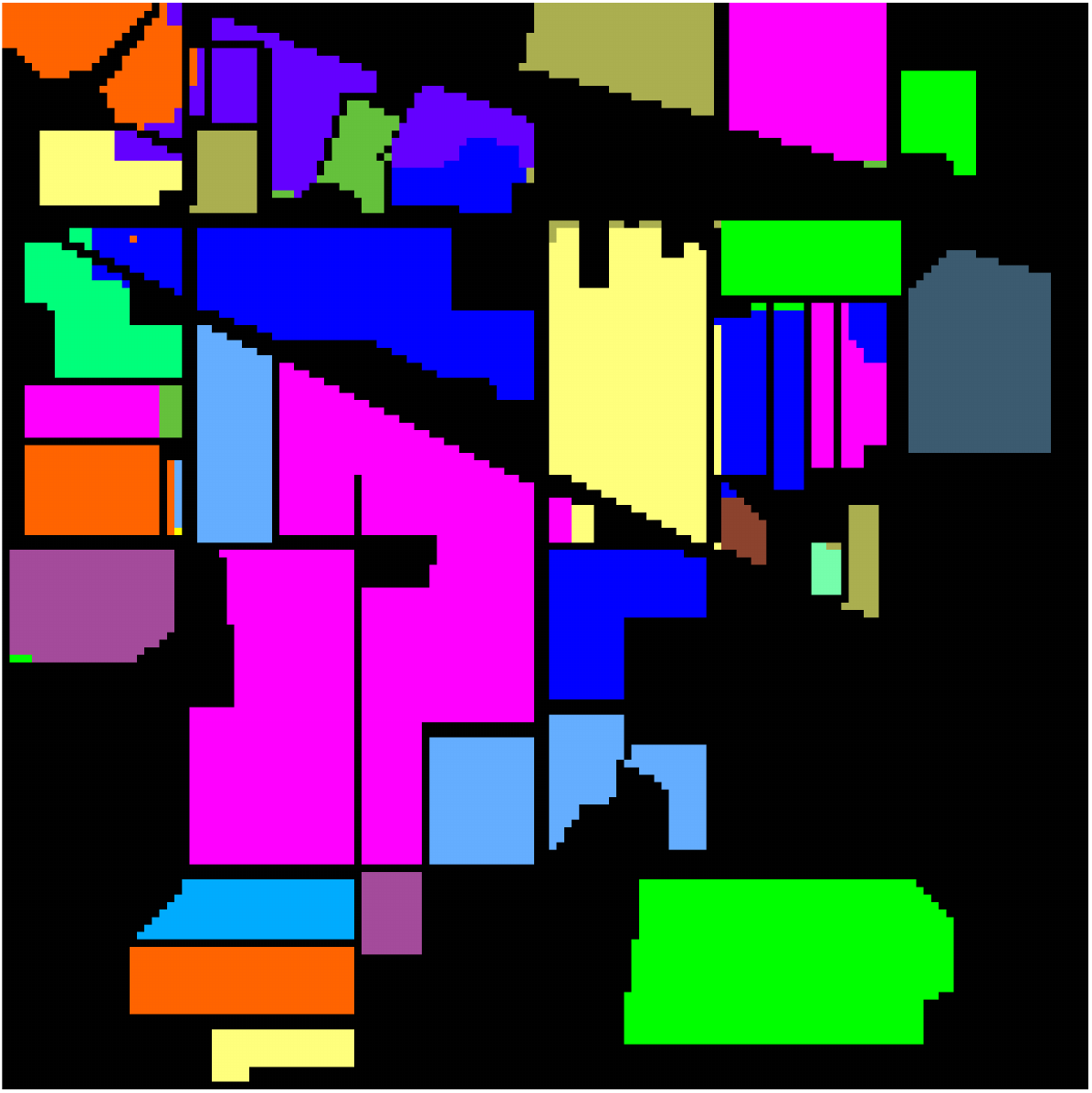}
\label{Fig:7}
\end{minipage}
}
\centering
\subfigure[]{
\begin{minipage}[t]{\length\linewidth}
\centering
\includegraphics[width=1.0\textwidth]{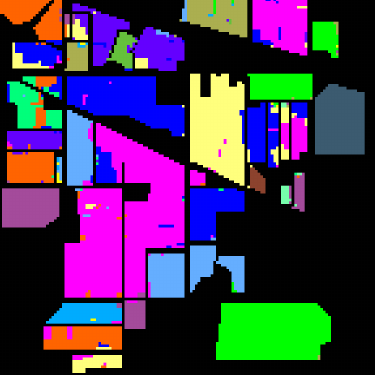}
\label{Fig:7}
\end{minipage}
}
\centering
\subfigure[]{
\begin{minipage}[t]{\length\linewidth}
\centering
\includegraphics[width=1.0\textwidth]{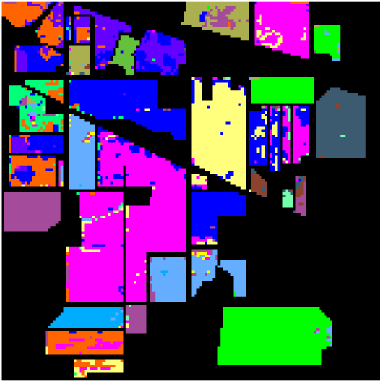}
\label{Fig:7}
\end{minipage}
}
\centering
\subfigure[]{
\begin{minipage}[t]{\length\linewidth}
\centering
\includegraphics[width=1.0\textwidth]{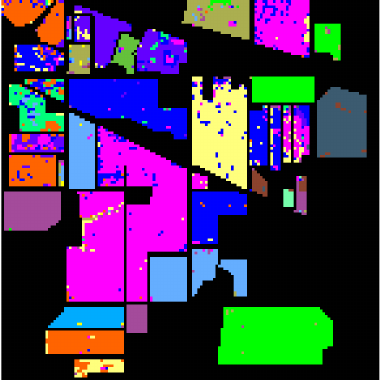}
\label{Fig:7}
\end{minipage}
}
\centering
\subfigure[]{
\begin{minipage}[t]{\length\linewidth}
\centering
\includegraphics[width=1.0\textwidth]{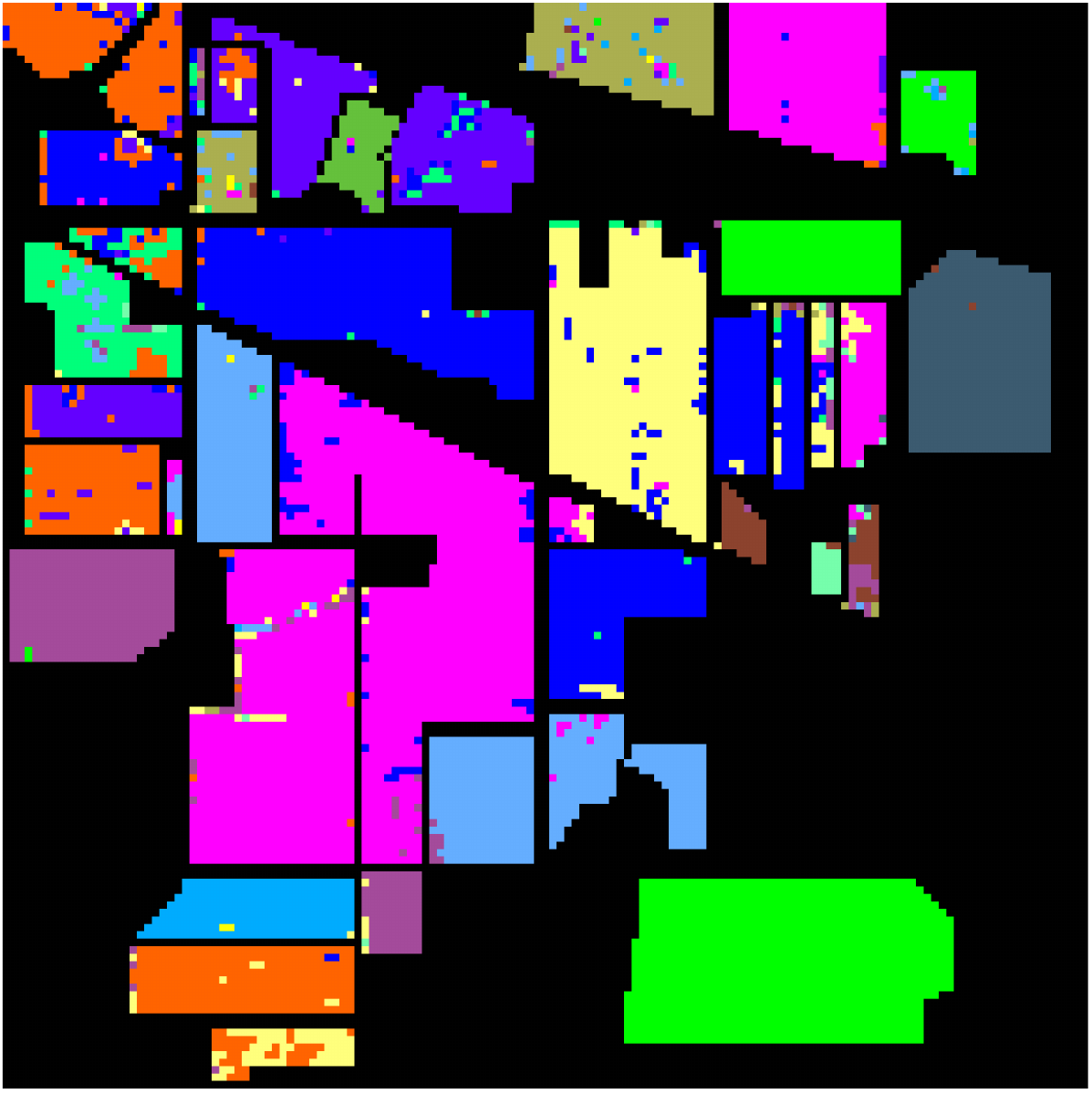}
\label{Fig:7}
\end{minipage}
}
\centering
\subfigure[]{
\begin{minipage}[t]{\length\linewidth}
\centering
\includegraphics[width=1.0\textwidth]{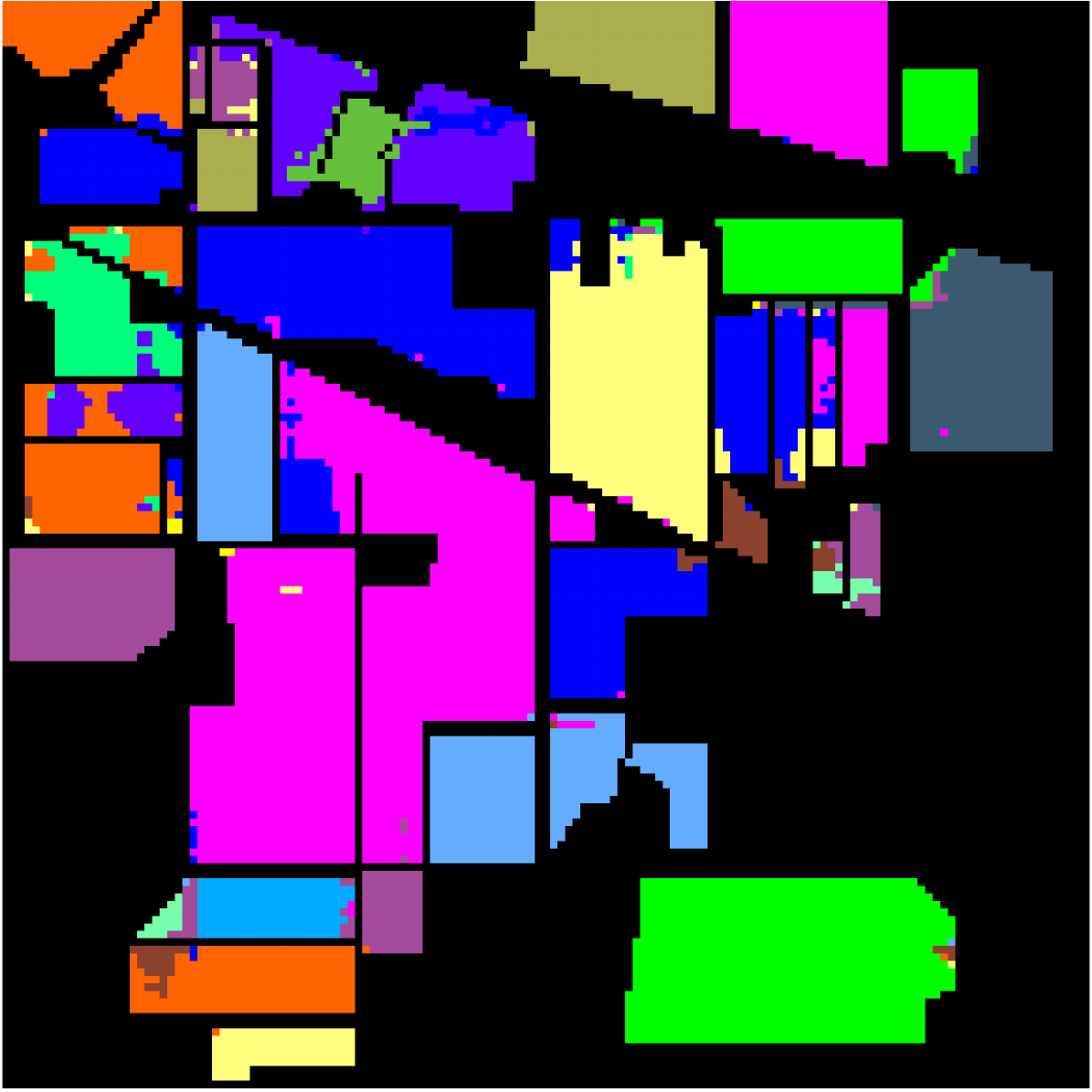}
\label{Fig:7}
\end{minipage}
}
\centering
\subfigure[]{
\begin{minipage}[t]{\length\linewidth}
\centering
\includegraphics[width=1.0\textwidth]{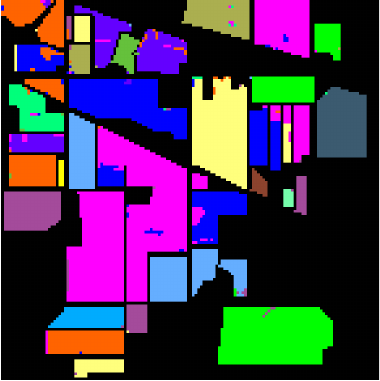}
\label{Fig:7}
\end{minipage}
}
\caption{Classification maps obtained by different methods for \textit{Indian Pines} when 5\% samples per class were used for training. The subfigures (a) to (h) represent the corresponding classification map of Groundtruth, SGR (90.51\%), CCJSR (92.34\%), LLRSSTV (83.51\%), LRMR (88.23\%), NAILRMA (90.56\%), S$^3$LRR (93.38\%), Proposed (96.82\%), respectively. (the OAs are reported in the parentheses).}
\label{fig:cmapIP}
\end{figure*}

\begin{figure*}[htbp]
\centering
\subfigure[]{
\begin{minipage}[t]{\lengthw\linewidth}
\centering
\includegraphics[width=1.0\textwidth]{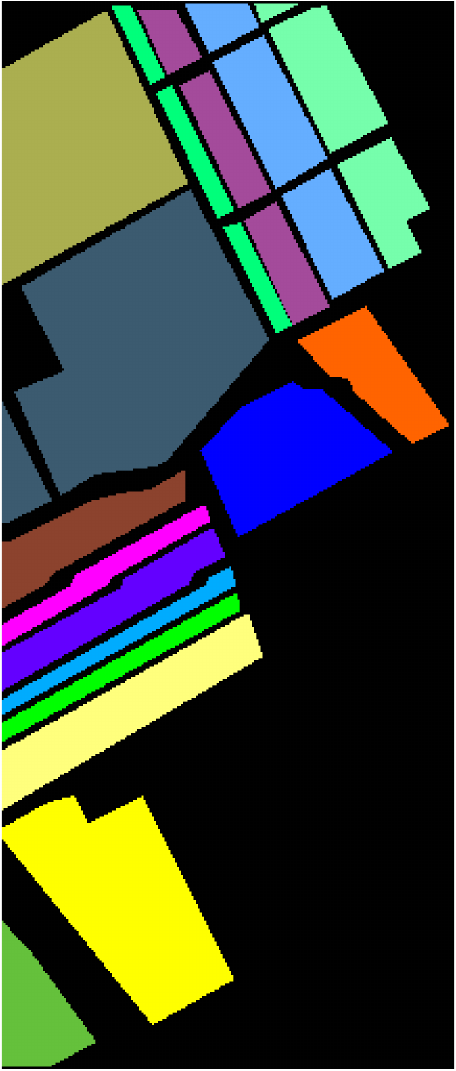}
\label{Fig:7}
\end{minipage}
}
\centering
\subfigure[]{
\begin{minipage}[t]{\lengthw\linewidth}
\centering
\includegraphics[width=1.0\textwidth]{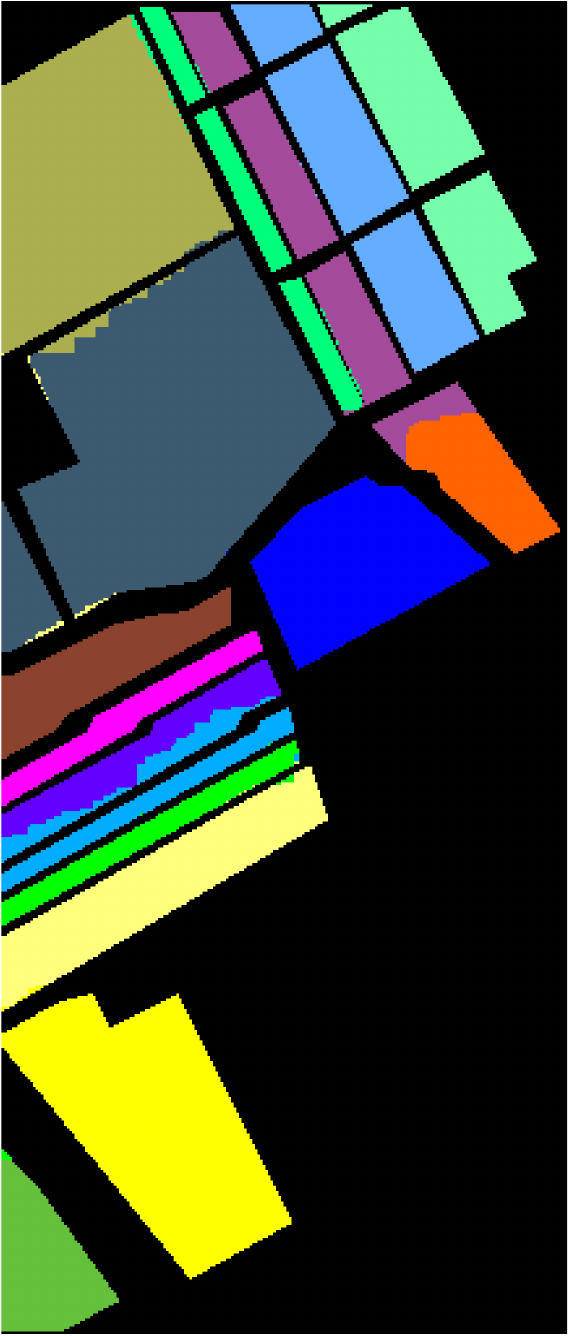}
\label{Fig:7}
\end{minipage}
}
\centering
\subfigure[]{
\begin{minipage}[t]{\lengthw\linewidth}
\centering
\includegraphics[width=1.0\textwidth]{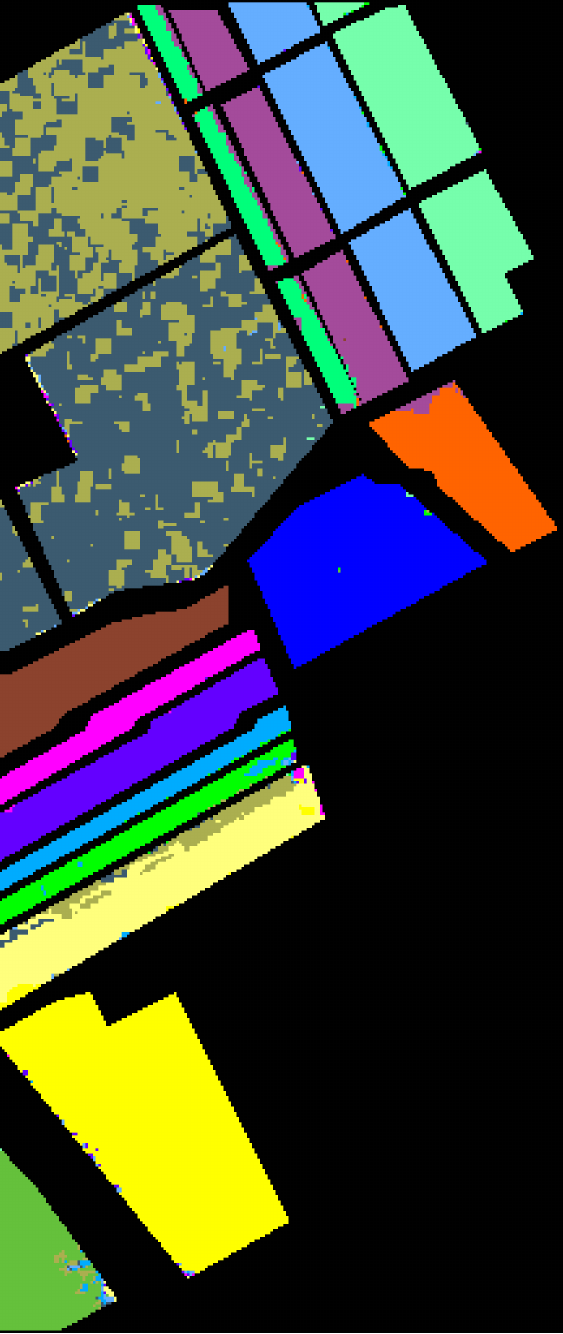}
\label{Fig:7}
\end{minipage}
}
\centering
\subfigure[]{
\begin{minipage}[t]{\lengthw\linewidth}
\centering
\includegraphics[width=1.0\textwidth]{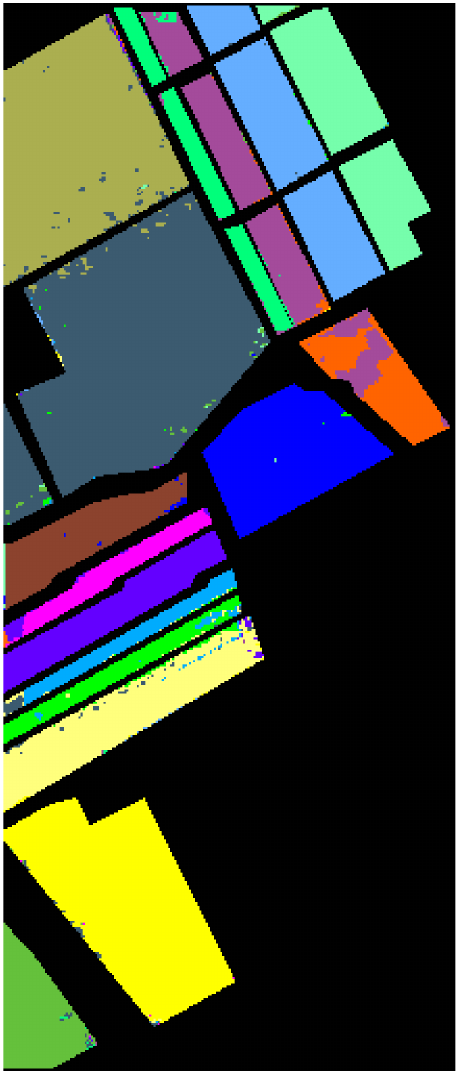}
\label{Fig:7}
\end{minipage}
}
\centering
\subfigure[]{
\begin{minipage}[t]{\lengthw\linewidth}
\centering
\includegraphics[width=1.0\textwidth]{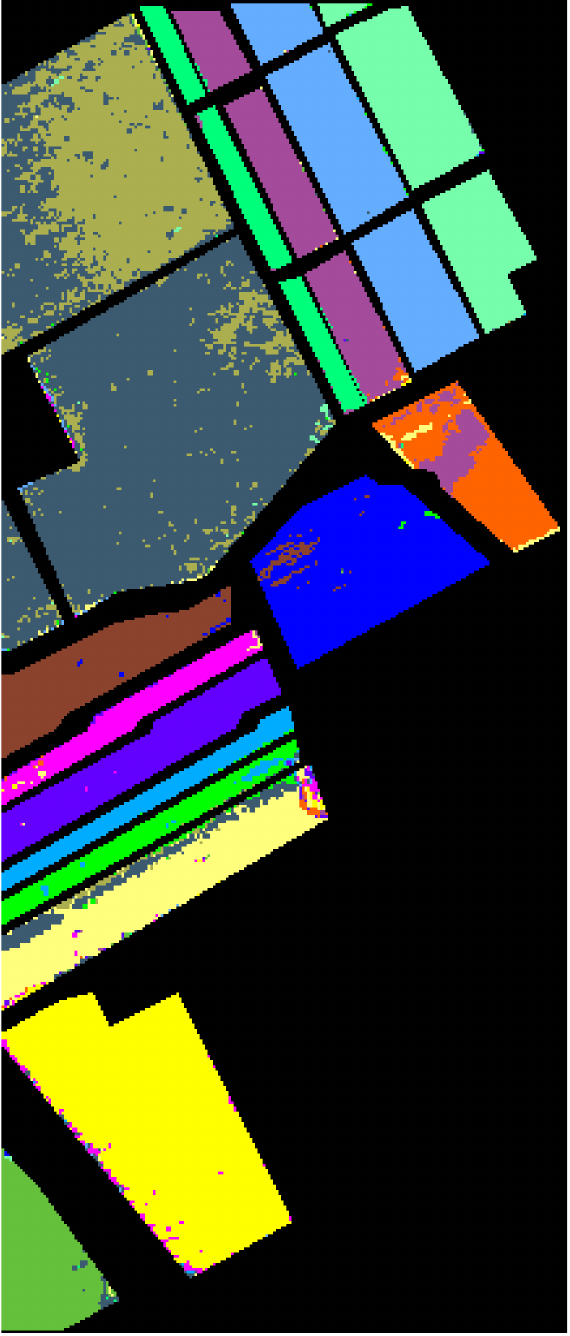}
\label{Fig:7}
\end{minipage}
}
\centering
\subfigure[]{
\begin{minipage}[t]{\lengthw\linewidth}
\centering
\includegraphics[width=1.0\textwidth]{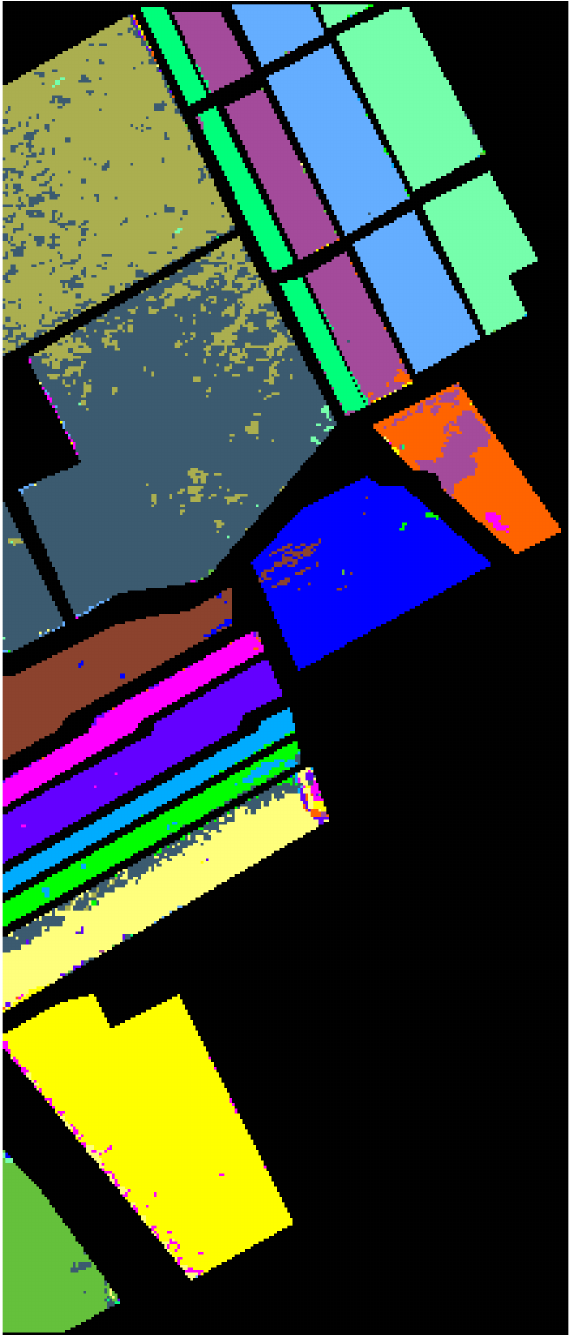}
\label{Fig:7}
\end{minipage}
}
\centering
\subfigure[]{
\begin{minipage}[t]{\lengthw\linewidth}
\centering
\includegraphics[width=1.0\textwidth]{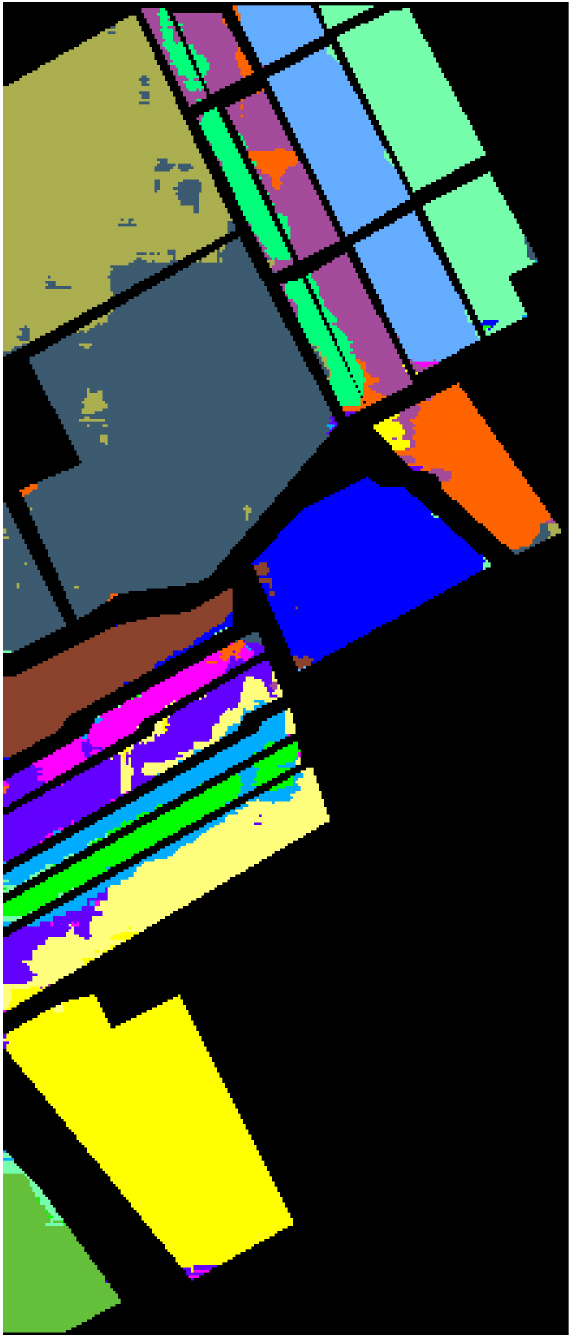}
\label{Fig:7}
\end{minipage}
}
\centering
\subfigure[]{
\begin{minipage}[t]{\lengthw\linewidth}
\centering
\includegraphics[width=1.0\textwidth]{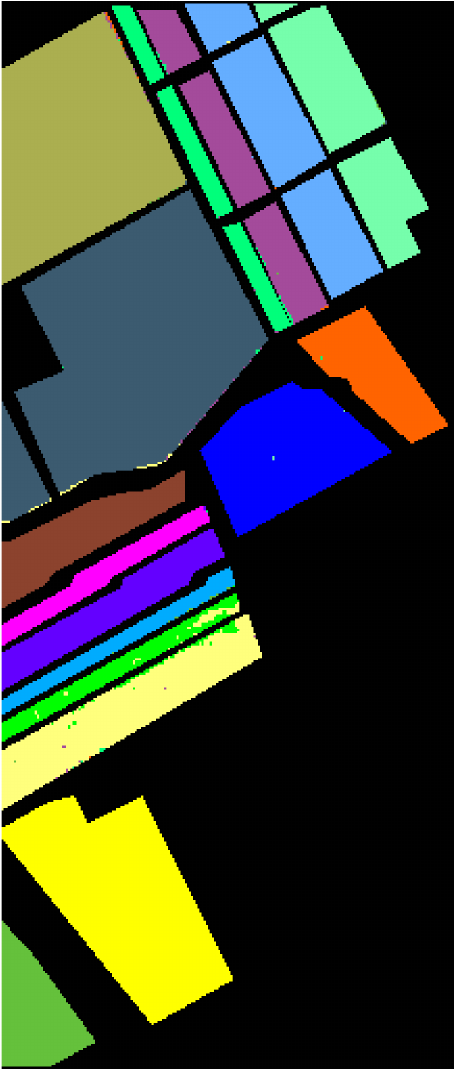}
\label{Fig:7}
\end{minipage}
}
\caption{Classification maps obtained by different methods for \textit{Salinas Valley} when 0.5\% samples per class were used for training. The subfigures (a) to (h) represent the corresponding classification map of Groundtruth, SGR (97.45\%), CCJSR (88.95\%), LLRSSTV (96.52\%), LRMR (89.85\%), NAILRMA (92.37\%), S$^3$LRR (92.44\%), Proposed (99.23\%), respectively. (the OAs are reported in the parentheses).}
\label{fig:cmapSA}
\end{figure*}

\begin{figure*}[htbp]
\centering
\subfigure[]{
\begin{minipage}[t]{\lengthw\linewidth}
\centering
\includegraphics[width=1.0\textwidth]{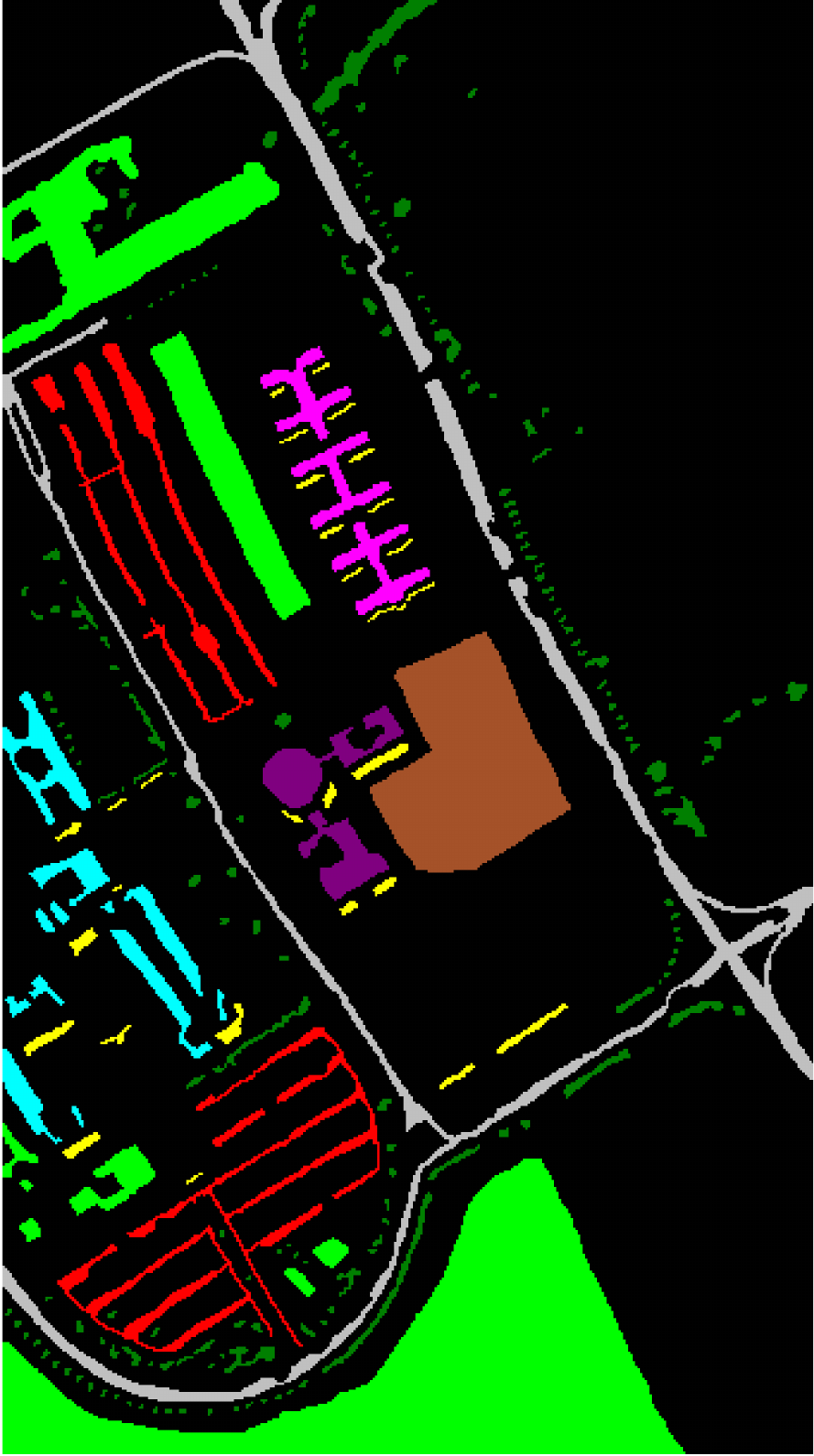}
\label{Fig:7}
\end{minipage}
}
\centering
\subfigure[]{
\begin{minipage}[t]{\lengthw\linewidth}
\centering
\includegraphics[width=1.0\textwidth]{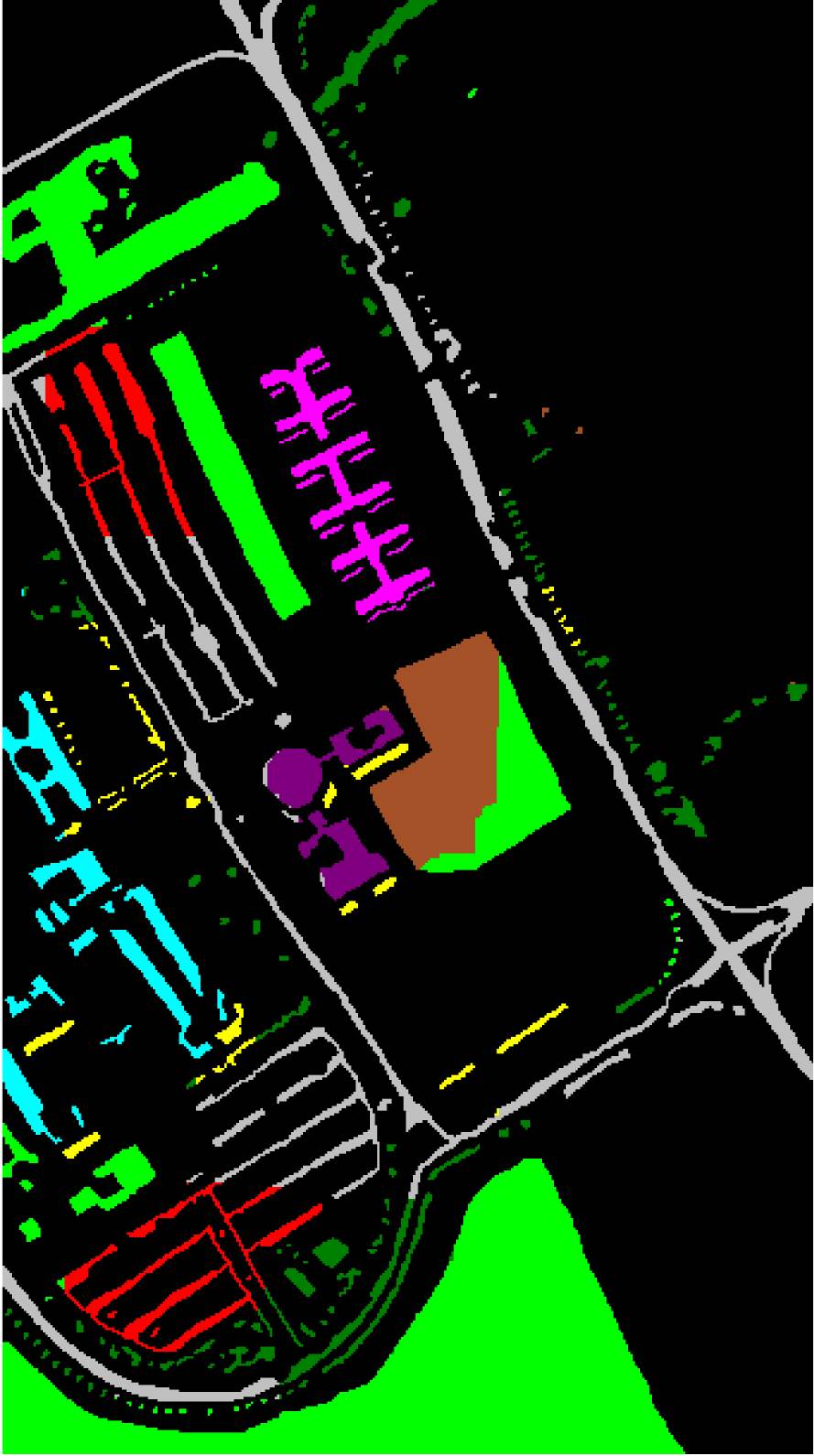}
\label{Fig:7}
\end{minipage}
}
\centering
\subfigure[]{
\begin{minipage}[t]{\lengthw\linewidth}
\centering
\includegraphics[width=1.0\textwidth]{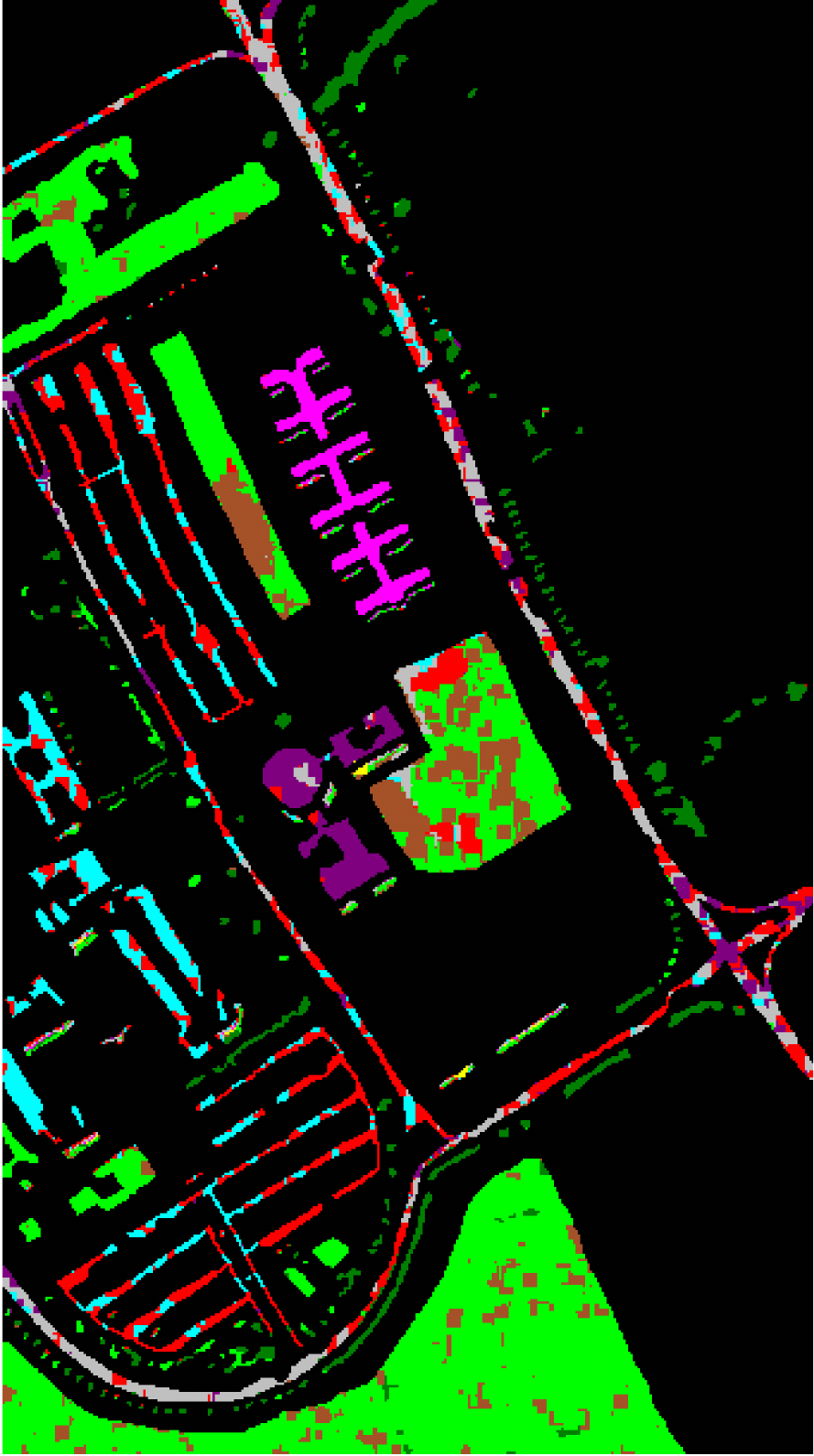}
\label{Fig:7}
\end{minipage}
}
\centering
\subfigure[]{
\begin{minipage}[t]{\lengthw\linewidth}
\centering
\includegraphics[width=1.0\textwidth]{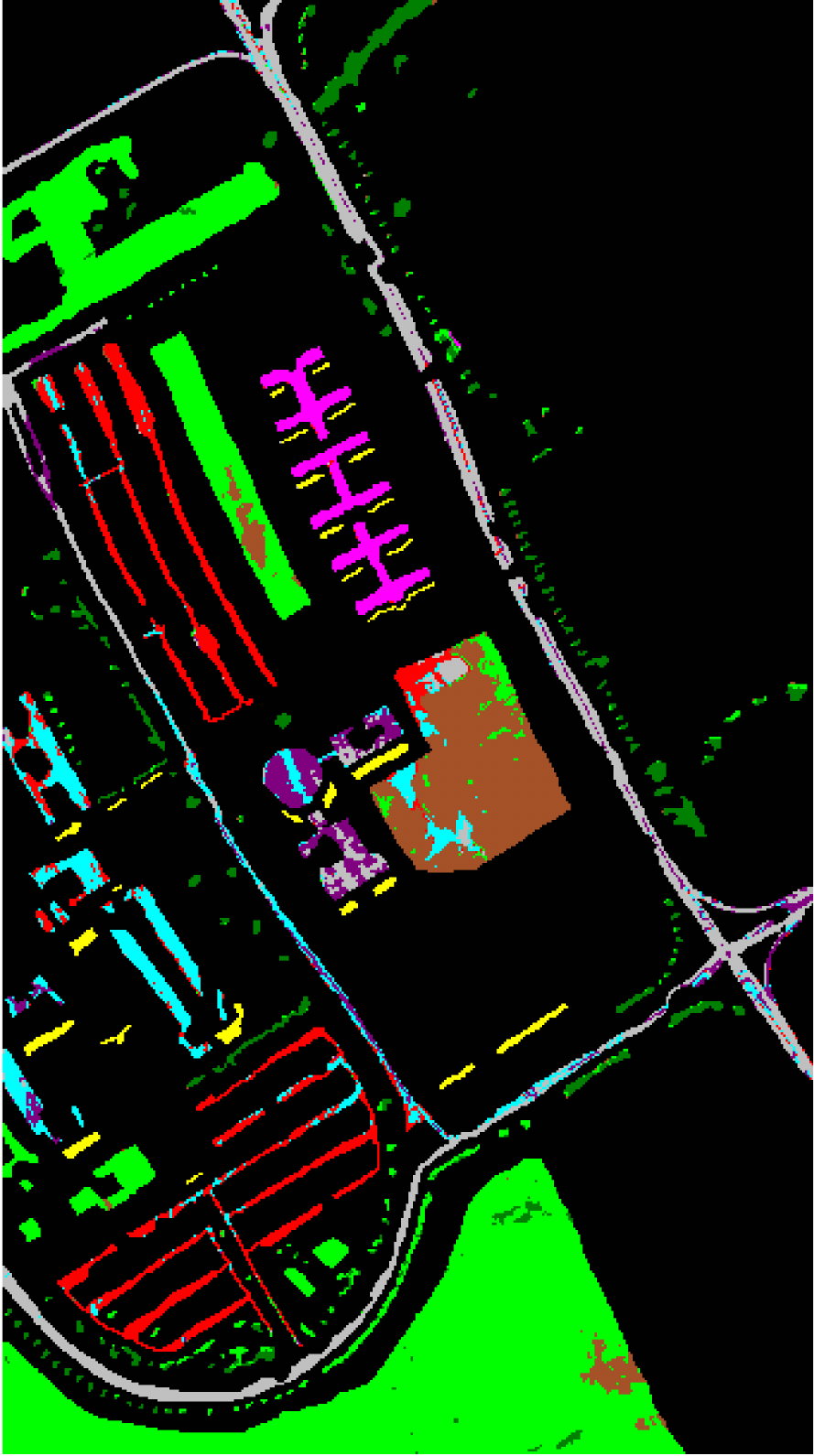}
\label{Fig:7}
\end{minipage}
}
\centering
\subfigure[]{
\begin{minipage}[t]{\lengthw\linewidth}
\centering
\includegraphics[width=1.0\textwidth]{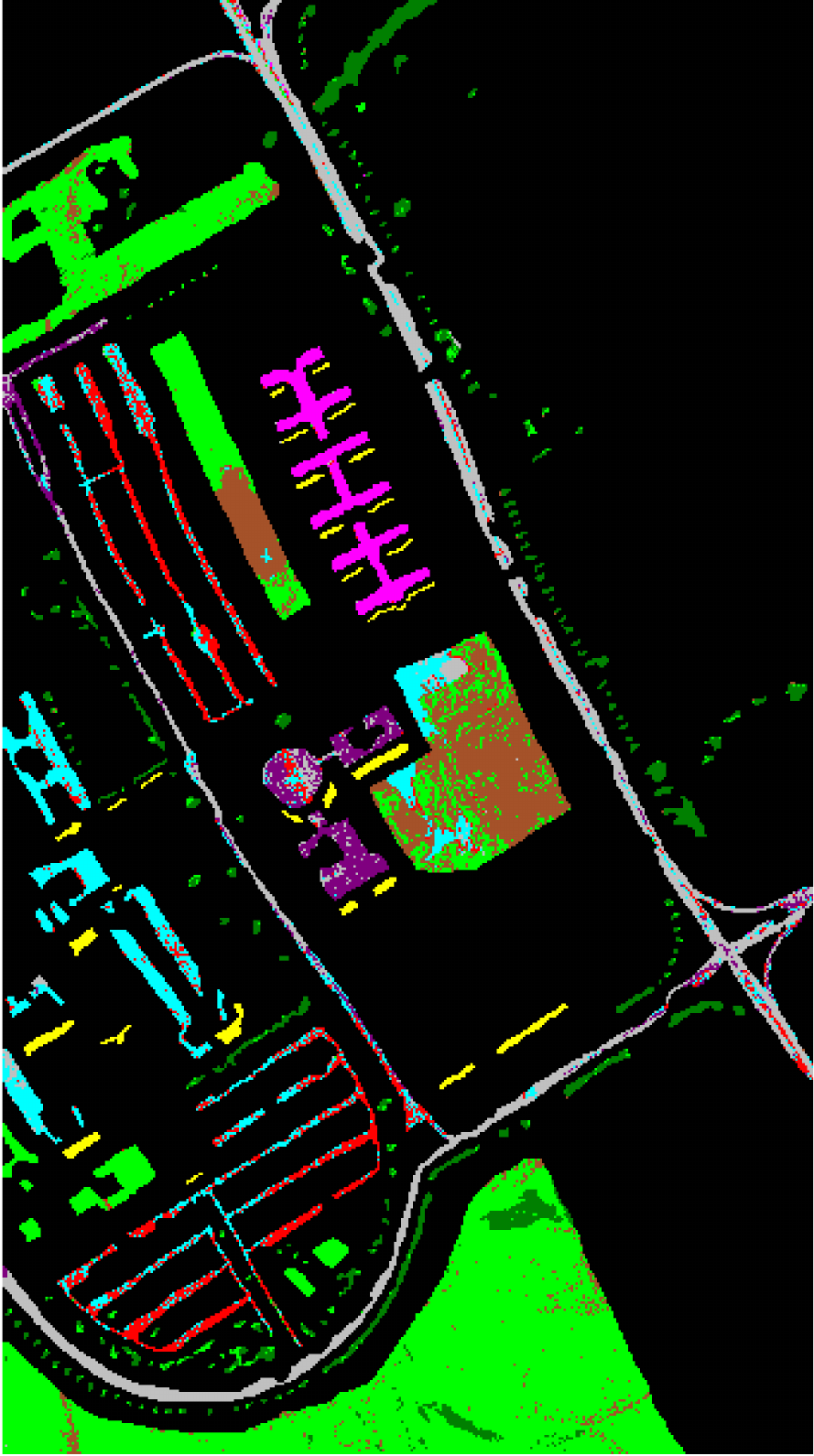}
\label{Fig:7}
\end{minipage}
}
\centering
\subfigure[]{
\begin{minipage}[t]{\lengthw\linewidth}
\centering
\includegraphics[width=1.0\textwidth]{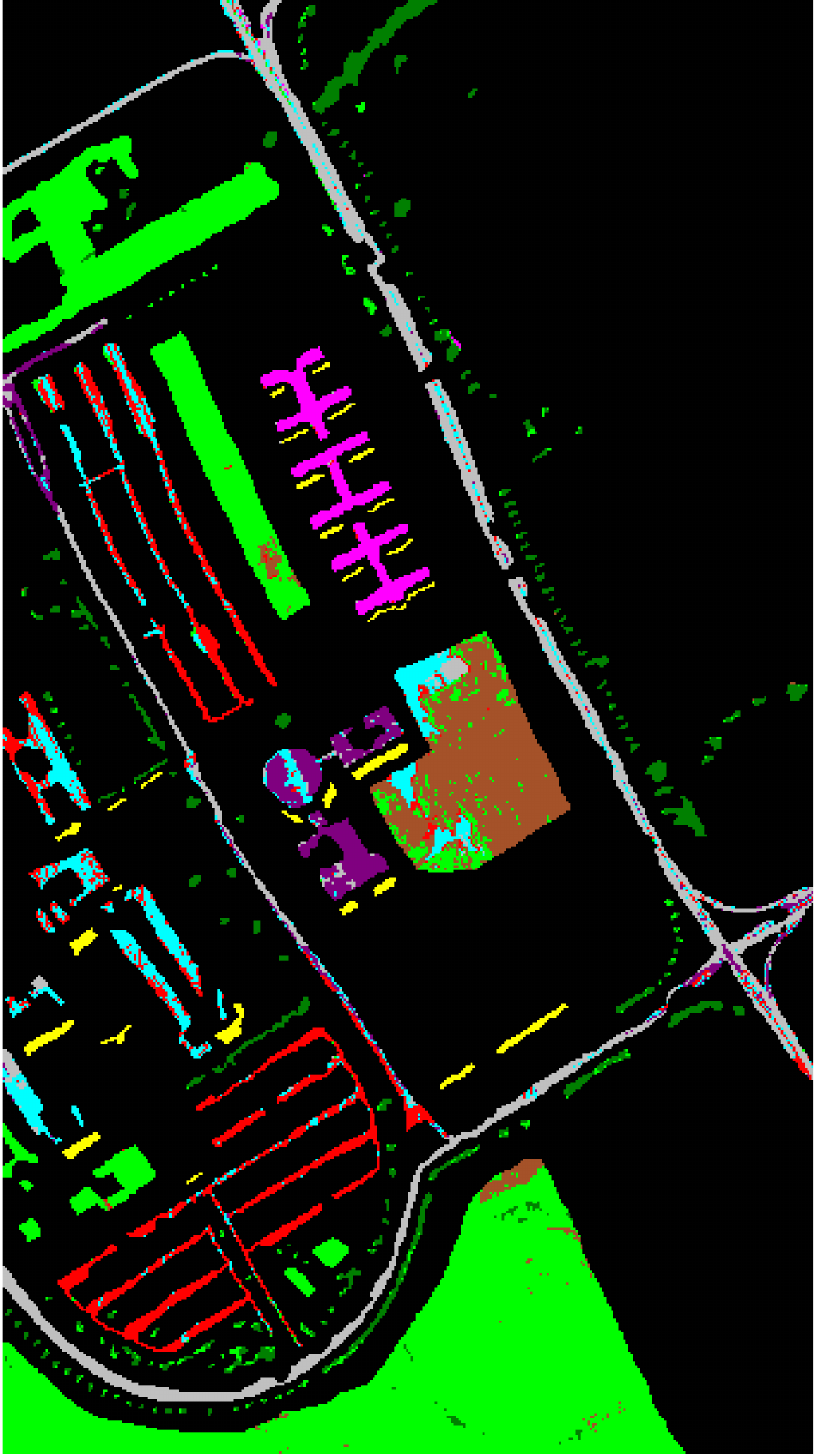}
\label{Fig:7}
\end{minipage}
}
\centering
\subfigure[]{
\begin{minipage}[t]{\lengthw\linewidth}
\centering
\includegraphics[width=1.0\textwidth]{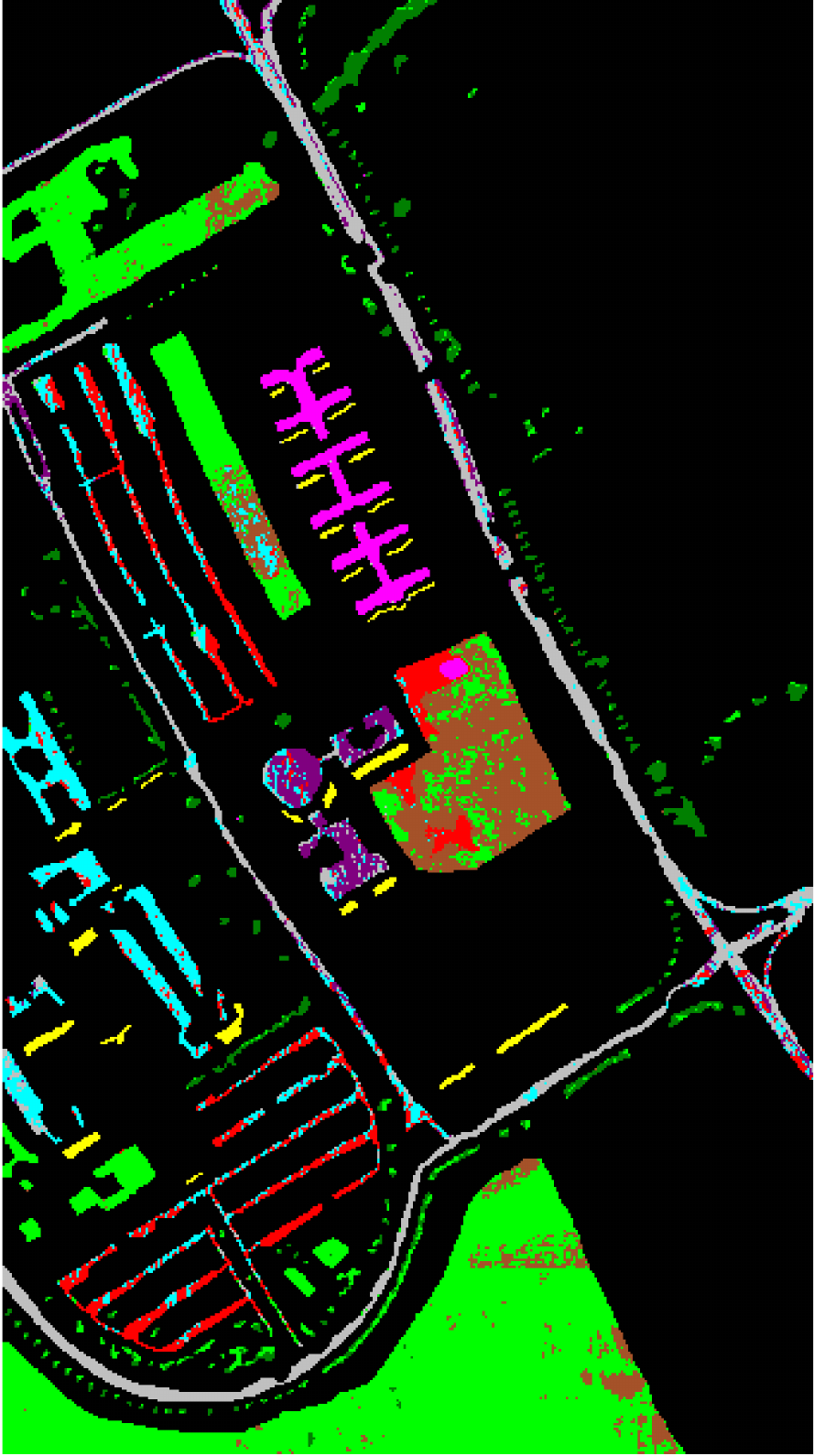}
\label{Fig:7}
\end{minipage}
}
\centering
\subfigure[]{
\begin{minipage}[t]{\lengthw\linewidth}
\centering
\includegraphics[width=1.0\textwidth]{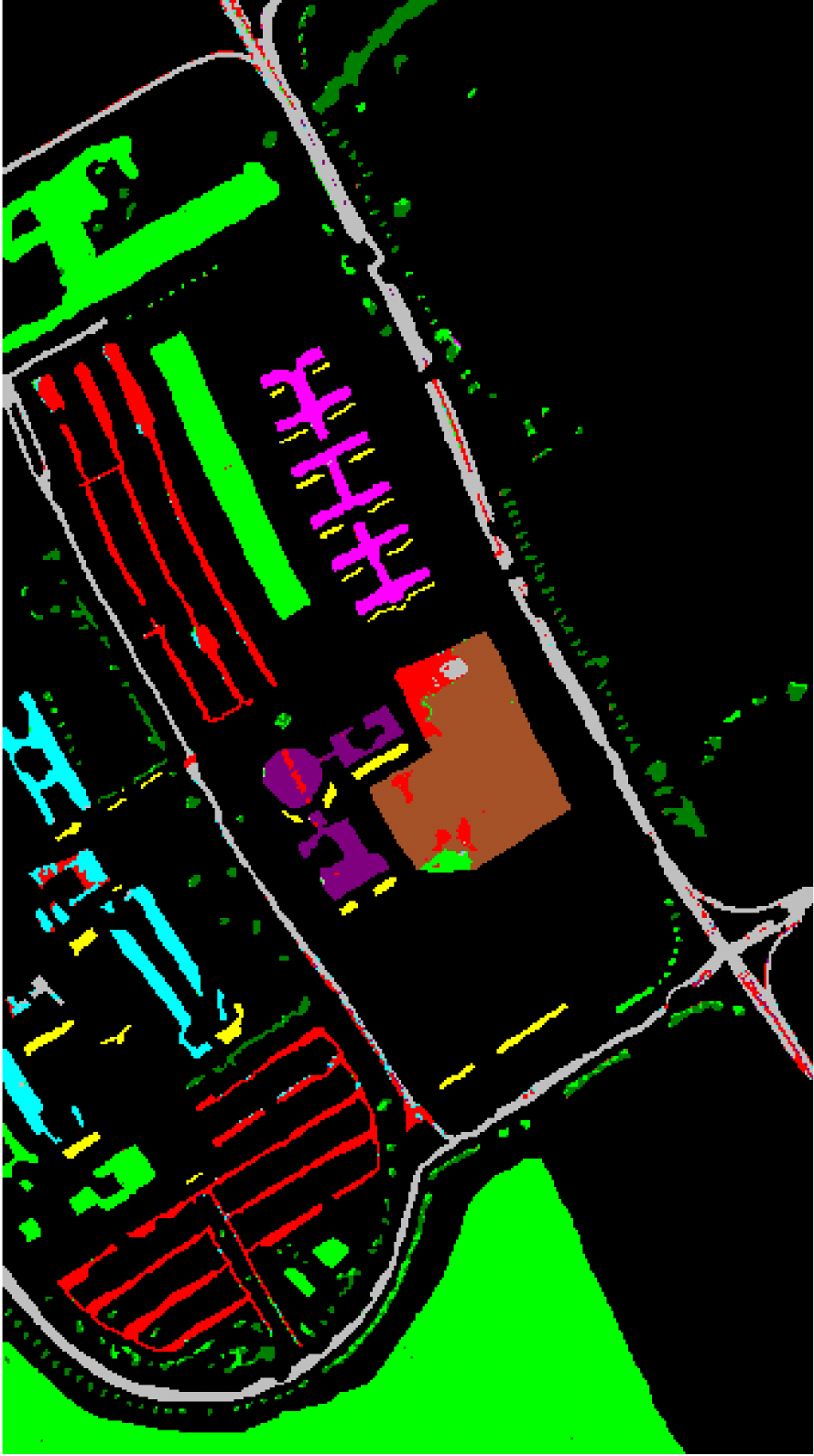}
\label{Fig:7}
\end{minipage}
}
\caption{Classification maps obtained by different methods for \textit{Pavia University} when 0.2\% samples per class were used for training. The subfigures (a) to (h) represent the corresponding classification map of Groundtruth, SGR (87.93\%), CCJSR (70.47\%), LLRSSTV (86.22\%), LRMR (80.89\%), NAILRMA (86.30\%), S$^3$LRR (80.49\%), Proposed (93.68\%), respectively. (the OAs are reported in the parentheses).}
\label{fig:cmapPU}
\end{figure*}

For \textit{Salinas Valley}, Table \ref{tab:SVsp} compares the classification accuracy from different methods, in which  0.5 $\%$ labeled samples of each class were used for training and the remaining ones for testing. It can be seen that SP-DLRR achieves the best performance in terms of all the three metrics OA, AA and $\kappa$. More specifically, for the classes 3, 8, 11, and 15, SP-DLRR shows the significant superiority.
In addition, Table \ref{tab:SVmp} shows the classification results for all the compared methods under different ratios of training samples for each class. 
We can see that SP-DLRR outperforms other methods on all the cases, where the significant superiority still can be seen when the training samples per class are extremely few, e.g., $P=0.1\%$.

For \textit{Pavia University}, Table \ref{tab:PUsp} shows the classification accuracy by different methods when 0.2 \% labeled samples of each class were used for training and the remaining ones for testing, where SP-DLRR shows remarkable superiority on the classes 1, 2, 6, 7, and 8. In addition, as shown in Table \ref{tab:PUmp}, SP-DLRR always achieves the best performance under different ratios of training samples for each class. 

Finally, the superiority of SP-DLRR is further demonstrated in Figs. \ref{fig:cmapIP}, \ref{fig:cmapSA}, and \ref{fig:cmapPU}, which show the classification maps of different methods over the three datasets.
It can be seen that the classification maps by SP-DLRR are closer to the ground-truth ones, indicating more pixels are correctly classified.

\subsection{Comparison of Computational Complexity}
Table \ref{tab:runningtime} compares the running times of different methods, 
where it can be seen that SP-DLRR consumes more time than the other methods, which is caused by the repeated computation of the DLRR of the input. Note that all these experiments were implemented with MATLAB 2017 on a PC with E5-2640 CPU @ 2.40 GHz.

\begin{table*}[htbp]
  \centering
  \caption{Comparison of running time (in seconds) of different methods. The best  and second-best results are highlighted in bold and underlined, respectively.}
  \label{tab:runningtime}
    \begin{threeparttable}
    \begin{tabular}{c|c|c|c|c|c|c|c|c}
    \hline
    \hline
  & P & SGR\cite{xue2017sparse}&CCJSR\cite{tu2018hyperspectral}&LLRSSTV\cite{he2018hyperspectral}& LRMR\cite{zhang2013hyperspectral}& NAILRMA\cite{he2015hyperspectral}& S$^3$LRR\cite{mei2018simultaneous}& Proposed\\
    \hline
   \textit{Indian Pines} & {5\%} & 116.90  & 226.20  & 174.46  & \underline{111.92}  & \textbf{24.40}  & 506.12  & 1321.60  \\
   \hline
    \textit{Salinas Valley}& {0.5\%}  & \underline{232.89}  & 580.92  & 809.70  & 600.08  & \textbf{79.76}  & 2128.70  & 5087.90  \\
    \hline
     \textit{Pavia University} & {0.2\%}  & 380.79 & \underline{132.49} & 987.95 & 735.28 & \textbf{85.21} & 2465.35 & 4830.00  \\
   \hline
    \hline
    \end{tabular}
    \end{threeparttable}
  \label{tab:addlabel}
\end{table*}

\section{Conclusion}
In this paper, we have proposed a novel superpixel guided discriminative low-rank representation method for HSI classification. Owing to the exploration of the local spatial structure in pixel-wise and the low-rank property in a local and global joint manner, our method not only increases the intra-class similarity but also promotes the inter-class discriminability. The experiments validate that our method produces much better performance than state-of-the-art methods. In addition, our method can handle the imbalance behavior of the data better. In future, we will consider improving the efficiency of SP-DLRR by using Anderson acceleration \cite{zhang2019accelerating}.
\section*{Acknowledgment}
The authors would like to thank Dr. Z. Xue for sharing the code of SGR \cite{xue2017sparse} to us.
\balance
\bibliographystyle{ieeetr}
\bibliography{LRGR}
\end{document}